\documentclass{article}

\usepackage{arxiv}

\usepackage[utf8]{inputenc} 
\usepackage[T1]{fontenc}    
\usepackage{hyperref}       
\usepackage{url}            
\usepackage{booktabs}       
\usepackage{amsfonts}       
\usepackage{nicefrac}       
\usepackage{microtype}      
\usepackage{lipsum}
\usepackage{graphicx}
\graphicspath{ {./images/} }
\usepackage{amsmath,amssymb,color,pgfplots,fancyhdr,hyperref,datetime,caption,wrapfig,parskip,booktabs,tikz,nicefrac,array,pdfpages,nomencl,soul,multirow,booktabs,graphicx, rotating,ragged2e,caption, subcaption, amsfonts, amssymb, verbatim, titlesec, float, multirow, xcolor, subcaption, pgfplots, nameref, hyperref}

\usepackage{fancyhdr}
\title{Machine learning pipeline for battery state of health estimation}

\author{
 Darius Roman \\
  Smart Systems Group\\
  School of Engineering \& Physical Sciences\\
  Heriot-Watt University\\
  Edinburgh, EH14 4AS, UK \\
  \texttt{dvr1@hw.ac.uk} \\
   \And
 Saurabh Saxena \\
  Center for Advanced Life Cycle Engineering \\
  University of Maryland \\ 
  College Park, MD 20742, USA \\
  \texttt{saxenas@umd.edu} \\
  \And
 Valentin Robu \\
  CWI, National Research Institute for \\ 
  Mathematics and Computer Science\\
  Amsterdam, The Netherlands\\
  \texttt{v.robul@cwi.nl} \\
  \AND
  David Flynn \\
   Smart Systems Group\\
   School of Engineering \& Physical Sciences\\
   Heriot-Watt University\\
   Edinburgh, EH14 4AS, UK \\
   \texttt{dflynn@hw.ac.uk} \\
   \And
   Michael Pecht \\
    Center for Advanced Life Cycle Engineering \\
    University of Maryland \\ 
    College Park, MD 20742, USA \\
    \texttt{pecht@umd.edu} \\
    \AND
    Note:\\
    \textit{Pre-print version of the article in Nature Machine Intelligence}
}

\begin{document}
\maketitle
\begin{abstract}

Lithium-ion batteries are ubiquitous in modern day applications ranging from portable electronics to electric vehicles. Irrespective of the application, reliable real-time estimation of battery state of health (SOH) by on-board computers is crucial to the safe operation of the battery, ultimately safeguarding asset integrity. In this paper, we design and evaluate a machine learning pipeline for estimation of battery capacity fade --- a metric of battery health --- on 179 cells cycled under various conditions. The pipeline estimates battery SOH with an associated confidence interval by using two parametric and two non-parametric algorithms. Using segments of charge voltage and current curves, the pipeline engineers 30 features, performs automatic feature selection and calibrates the algorithms.  When deployed on cells operated under the fast-charging protocol, the best model achieves a root mean squared percent error of 0.45\%. This work provides insights into the design of scalable data-driven models for battery SOH estimation, emphasising the value of confidence bounds around the prediction. The pipeline methodology combines experimental data with machine learning modelling and can be generalized to other critical components that require real-time estimation of SOH. 
 
\end{abstract}


\section{Introduction}

Rechargeable Li-ion batteries play a crucial role in many modern-day applications ranging from portable electronics and medical devices, to renewable energy integration in power grids and electric vehicles. The steep decrease in the price of lithium-ion based battery storage by 73\% from 2010 to 2016, to an all-time low of \$273$/$kWh in 2017 \cite{bloomberg_report} opened up a significant energy storage market evaluated at \$65 billion in 2017 \cite{bernhart2019challenges}. Irrespective of the application, Li-ion batteries degrade with time. With ageing, cells exhibit a loss of capacity and an increase in impedance. The rate of degradation is influenced by the dynamic operating conditions, including varying charge/discharge rates, different voltage operation limits and temperature fluctuations. The ability to estimate degradation in real-time irrespective of the various failure mechanisms and their degradation paths is crucial for safe and effective battery management systems \cite{ev_diagnosis}. Battery state of health can be used to predict battery's expected lifetime, however, the feasibility of online state of health estimation via direct measurement of chemical reaction parameters inside batteries remains limited \cite{review_intro}. 

State of health (SOH) is a parameter that quantifies the general condition of a battery and its ability to deliver the specified performance, measured as capacity or impedance, when compared to its unused state. This work focuses on the battery capacity as the health indicator due to its correlation to the energy storage capability of batteries and its direct impact on the remaining run time and life of the batteries. Capacity fade estimation has received considerable research interest from industry and academia \cite{review_intro}, \cite{battery_review_1}, \cite{battery_review_2}, \cite{battery_review_3} and a number of methods have been proposed. The current approaches to capacity fade estimation involves parameter estimation using either of the following modelling types, equivalent circuit models (ECMs) \cite{electrical_equiv_model_review}, \cite{online_ECM}, \cite{EIS_3}, electrochemical models \cite{electrochemistry_model_nasa}, \cite{electro_model_bole}, \cite{parama_chemical_model}, or data-driven models \cite{severson_satandford}, \cite{RUl_prognostics_NASA}, \cite{prognostics_NASA}, \cite{batery_health_bayesin_SVM}, \cite{SVM_battery_health}, \cite{attia2020closed}. Electrochemical models approximate the chemical processes that take place inside a battery cell during operation. This type of modeling requires detailed cell specifications, such as electrode materials and electrolyte chemistry. The method typically deploys complex partial differential equations, leading to significant requirements of both memory and computational power. ECMs, on the other hand, employ circuit components with empirical nonlinear parameters \cite{online_ECM}. Compared to electrochemical models, ECMs use fewer inputs, considerably reducing the number of parameters required to be learnt over time, however, they have limited accuracy and robustness due to assumptions in battery behavior \cite{electrical_equiv_model_review}. Furthermore, in order to determine ECM model parameters, such as the ohmic resistance and the parallel resistor-capacitor impedance, at different state of charge values, pulse discharging \cite{pulse_1} and electrochemical impedance spectroscopy is typically necessary \cite{EIS_3}, \cite{EIS_1}, \cite{EIS_2}, however such measurements are not a viable solution for online applications.

Conversely, the data-driven  approach displays a series of advantages such as a chemistry-agnostic modelling capability and an ability to analyse a wide range of degradation mechanisms and operating conditions, including rare loading events often overlooked by simplified models or physics-based simulations.  To date, numerous studies have employed machine learning tools for the analysis of battery SOH estimation. Several studies \cite{birkl_degradation}, \cite{IC_full_curve_ensemble}, \cite{IC_curves_1}, \cite{IC_curves_2}, \cite{online_SOH_partical_IC} showed that incremental capacity (IC) and differential voltage (DV) curves, a method developed for use in cell aging mechanism analysis \cite{birkl_degradation}, can also be used for offline and online capacity fade estimation. However, the approach has several drawbacks linked to obtaining the IC and DV curves that substantially reduce its practicality. The differentiation of the capacity-voltage curve to obtain the IC curve amplifies noise and propagates it into the algorithm. Additionally, both curves must cover a sufficient voltage range for the method to work and, for obtaining a high curve fidelity, it is restricted to low charge current rates(1/5 to 1/25 C-rate) \cite{quick_online}, \cite{dubarry2006incremental}, \cite{weng_onboard_IC}. Unless a low current value is used during charging protocol and the key part of the capacity-voltage curve is recorded, such that specific peak points in the IC curve are captured, the method is impractical for online deployment.

An alternative is to train an algorithm on the raw voltage-time data curve, eliminating the need for differentiation  \cite{charge_curve_gpr}, \cite{Howey_GPR}. Notably, Richardson et al. \cite{Howey_GPR} operated on sections of the voltage-time data itself by first smoothing the curve using a Savitsky-Golay filter and then used equispaced voltage values as the input to a Gaussian process regression (GPR) algorithm. However, GPR is considerably slow to train due to its computational cost of learning governed by the kernel function \cite{slow_GPR}, making it unsuitable for online deployment. The high computational complexity, also severely limits its scalability to incorporate bigger datasets. Additionally, the algorithm is sensitive to the section of the voltage-time curve used as input to the GPR. Other Bayesian-based methods, such as the relevance vector machine (RVM) \cite{bayesian_batterya}, have also been used to estimate battery capacity fade.  Unfortunately, RVM also suffers from slow training, particularly when compared to frequentist-based algorithms \cite{accelerating_RVM}. Shen et al. \cite{slow_GPR} presented options for accelerating GPR, however, they compromise accuracy. In contrast to \cite{Howey_GPR} where the constant current part of the charging profile was used, Wang et al. \cite{CCCT_CVCT_Zhang} used the constant voltage section to estimate capacity fade using support vector regression (SVR). Although SVR is faster than GPR, it lacks the ability to estimate uncertainty.  This inability to estimate uncertainty stemmed from various sources is a major limiting factor when discussing complex dynamic systems, such as Li-ion batteries. SOH assessment without corresponding measures of uncertainty associated with the estimation does not provide sufficient information to form a decision or corrective action plan \cite{prognostics_nasa_uncertainty}. 

Previous work \cite{electrical_equiv_model_review}-\cite{CCCT_CVCT_Zhang} includes limited assessments of SOH uncertainty or none at all. The proposed machine learning pipeline is capable of real-time estimation of battery SOH and associated algorithm uncertainty referred to as battery health and uncertainty management pipeline (BHUMP). BHUMP operates by passing incoming data streams through a hierarchical sequence of processing steps by first engineering features based on segments of raw charge curves.  It then performs offline automatic feature selection, augments the dataset with adversarial examples, and estimates battery health and associated uncertainty with the aid of four machine learning algorithms. Uncertainty is quantified based on calibration error and an adapted accuracy measure, the $\alpha$-$\beta$ accuracy zone. There are numerous battery designs \cite{review_batteries_2} and chemistries available \cite{review_batteries}, therefore the pipeline is deployed on a total of 179 cells, three designs (prismatic, pouch, and cylindrical), two chemistries (LiFePO$_{4}$, and LiCoO$_{2}$), three charge protocols (constant current, constant current - constant voltage, and 2-step fast-charge), and various discharge rates. 

This paper refines and extensively tests new and improved machine learning algorithms for the capacity fade estimation problem, but also defines metrics for estimating and accurately quantifying uncertainty in ML models used in battery research. BHUMP provides battery researchers with a scalable SOH estimation solution that is adaptable to any cell chemistry and operating condition. BHUMP is more accurate than conventional methods as the battery is ageing, uses a set of engineered features capable of capturing battery intrinsic degradation, and is capable of estimating cell SOH in under 15 minutes at any point in its life-cycle. An accurate SOH method combined with a quantifiable metric for uncertainty propagation that feeds into SOC and run time calculations improves battery performance and ultimately extends cell lifetime.

 \section{Machine learning pipeline approach} 

\subsection{Pipeline overview}
From a machine learning perspective, determining battery capacity fade can be considered a multivariate supervised regression problem. We use a pipeline-based approach, where features are engineered from charge/discharge curves, on which a Bayesian or frequentist model is trained. Additionally, uncertainty is quantified by predicting a distribution mean and an associated standard deviation. Our learning method is divided into two stages, namely, Stage 1: Offline pipeline creation and training and Stage 2: Online SOH estimation. The offline stage ensures feature engineering, training data augmentation, automatic feature selection, algorithm training, and uncertainty calibration. The online stage diagnoses the cell using the trained pipeline under the assumptions that it is given a battery cell of unknown capacity. Supplementary Figure 1 provides a summary of the two stages via a flowchart of the method.

\begin{figure*}[h!]
\centering
\resizebox{\textwidth}{!}{%
\begin{tabular}[t]{ccc}

\begin{tabular}{c}
    \smallskip
        \begin{subfigure}[b]{.48\textwidth}
            \centering
            \caption{}
            \includegraphics[width=\textwidth]{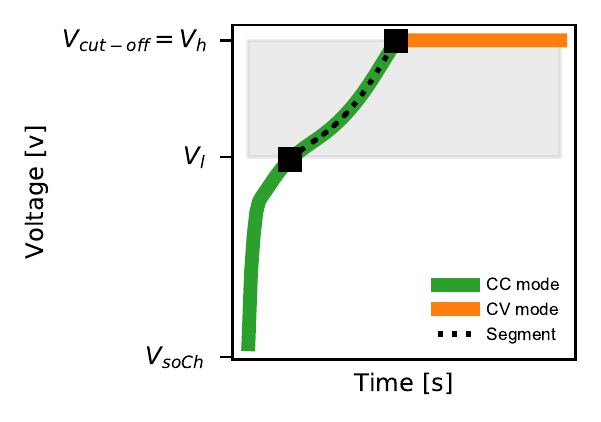}
          \label{thresholds_v_gr1}

        \end{subfigure}\\
        \begin{subfigure}[b]{.48\textwidth}
            \centering
            \caption{}
            \includegraphics[width=\textwidth]{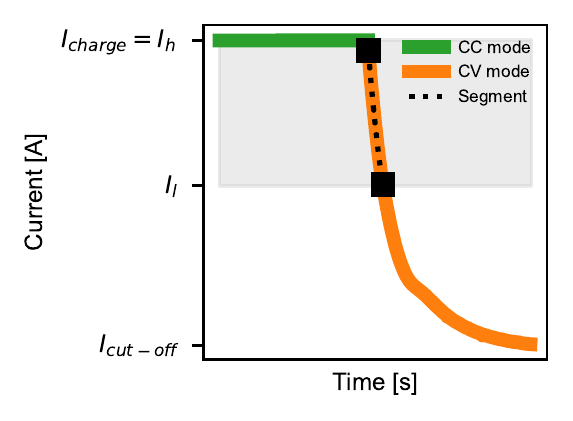}
            \label{thresholds_c_gr1}

        \end{subfigure}
\end{tabular}
&
\begin{tabular}{c}
        \smallskip
            \begin{subfigure}[b]{.45\textwidth}
                \caption{}
                \centering
                \includegraphics[width=.9\textwidth]{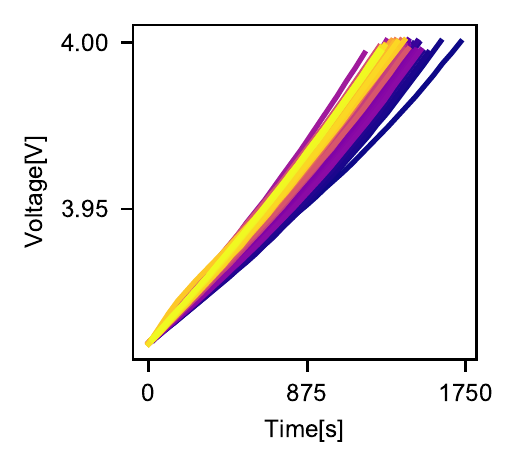}
                \label{b1}
            \end{subfigure}\\
            \begin{subfigure}[b]{.45\textwidth}
                \centering
                \caption{}
                \includegraphics[width=.9\textwidth]{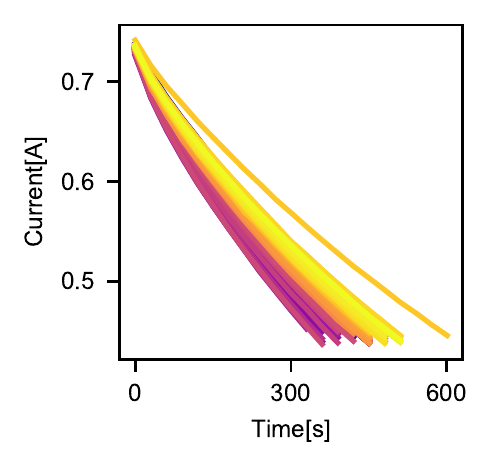}
                \label{d1}
            \end{subfigure}
\end{tabular}
&
\begin{subfigure}{0.5\textwidth}
    \centering
    \caption{}
    \includegraphics[width=.47\textwidth]{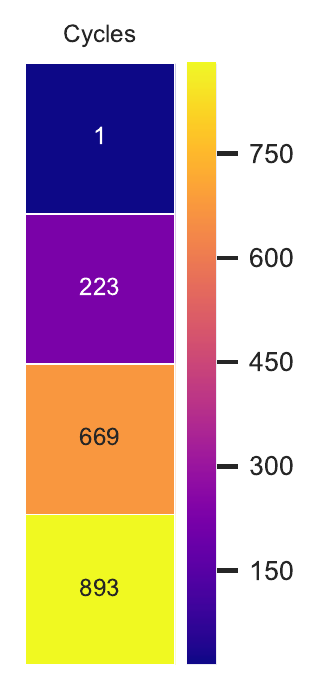}
    \label{e1}
\end{subfigure}

\end{tabular}%
}

    \caption{\textbf{The CC-CV charge protocol and extracted ageing segment of the curves for a Li-ion pouch cell.} \textbf{\protect\subref{thresholds_v_gr1}} Voltage during charge protocol. \textbf{\protect\subref{thresholds_c_gr1}} Current during charge protocol, \textbf{\protect\subref{b1}} Extracted ageing voltage curve segments corresponding to marked grey area, \textbf{\protect\subref{d1}} Extracted ageing current curve segments corresponding to marked grey area, \textbf{\protect\subref{e1}} Heatmap of ageing with cycle number. Note: Refer to Methods section for abbreviations.}
    \label{features_calce}
\end{figure*}

Feature engineering is split into automatic feature generation or extraction through techniques such as neural network auto-encoders \cite{auto_encoder_bearing}, \cite{deep_feature_sunthesis}, and manual feature construction based on domain knowledge \cite{feature_Nick}, \cite{online_F1_F2_Zhang}. We adopt a domain knowledge-based approach, where we show the algorithm feature choice based on importance to target variable. We also, provide a hypothesis for the underlying physical degradation quantified by the selected segments of the charge curves in Supplementary Note 1. Supplementary Table 1 summarize the attributes recorded during life cycle testing. The pipeline focuses on segments of the charge curves to capture degradation in the electrodes during cycling (Figure \ref{features_calce} illustrates typical extracted segments). The extracted charge-curve segments are further used in the feature engineering process (see Methods for details).  

The pipeline creates a total of 30 features, and selects the most relevant features using a random forest based recursive feature elimination with cross-validation (RF-RFE-CV) similar to the one introduced in \cite{guyon_RFE_CV}. Recursive feature elimination generally outperforms other conventional methods \cite{RF_RFE}, \cite{RF_RFE_2}, hence the adoption here (refer to Methods section for further details). Before training the algorithms, we perform data augmentation by introducing adversarial examples as proposed by Goodfellow et al.~\cite{goodfellow_fgsm} in combination with the weight decay algorithm (see Methods). The use of adversarial examples in our datasets was motivated by the need to ameliorate the differences in battery design/chemistry. In addition, training on adversarial data makes the algorithm robust to outliers, prevents overfitting and reduces distribution variance around the estimated mean. Synthetic data generation generated from electrochemical models like the pseudo-two-dimensional model proposed by Doyle et. al. \cite{doyle1993modeling} can also be regarded as a data augmentation policy. Such an approach harnesses the potential of both electrochemical and data based models and we believe future work must incorporate synthetic data as well.

The augmented dataset  then serves as the training input to four algorithms: random forest (RF) and deep neural network ensemble (dNNe), Bayesian ridge regression (BRR) and Gaussian process regression (GPR). Unlike Bayesian based algorithms, BRR and GPR, frequentist algorithms are unable to quantify uncertainty naturally due to their formulation. To overcome such limitations, we consider two modified ensemble based algorithms: RF with Infinitesimal Jackknife (IJ) based confidence intervals \cite{RF_CI} and the ensemble of neural networks as described in \cite{deep_ensamble}. For training of the algorithms  a random search approach is used for hyper-parameter tuning \cite{random_hyper_tuinning}, with the exception of the deep ensemble where the Adam optimiser is used. We have found that drawing random samples from a uniform distribution works best for BRR and GPR parameters, whereas for RF and dNNe parameters random initialisation gives satisfactory results. In addition, a batch cross-validation method is used during the hyper-parameter tuning, where each batch is represented by one cell. This prevents the over-fitting of the models and mimics online deployment. Machine learning models in engineering require a stringent performance evaluation both from an error and uncertainty quantification perspective. The models are initially re-calibrated followed by an evaluation based on mean absolute percent error, root mean squared error and uncertainty estimation metrics (refer to  Methods section for further details). 

\subsection{Methods}

This study developed a pipeline approach for battery SOH estimation, called BHUMP and it incorporates a series of hierarchical steps, feature engineering, feature selection and data augmentation prior to model fitting and tuning as follows.

\textbf{Feature engineering} performs mathematical manipulations of extracted parts of the voltage curve during the constant current charge protocol based on a lower voltage threshold, $V_l$ and an upper voltage threshold, $V_h$ (refer to the grey area of Figure \ref{thresholds_v_gr1}) for all datasets except for cells charged with a 2-step fast-charge protocol. A characteristic to the 2-step fast charge protocol is that the cells can be charged from 0\% SOC to 80\% SOC with high currents ranging from 3.6 C-rate to 6 C-rate. In this work, due to the nature of the charging method in the 2-step fast charge, we only use the constant-current constant-voltage (CC-CV) charge part of the charging protocol as per the black dotted segments in the grey area observed in Supplementary Figures 2a, 2b. The values of $V_h$ and $V_l$ can be selected based on the intended application and the depth of discharge of the cell. In this work we select $V_h$ to be equal to cut-off voltage, $V_{cut-off}$. Refer to Supplementary Note 2 on how we select $V_l$. Additional features are developed on extracted segments of the current curve during the constant voltage charge protocol based on two current threshold values, $I_h$ and $I_l$ respectively (see Figure \ref{thresholds_c_gr1}) for all cells except for the 2-step fast charge protocol. We select $I_h$ equal to charge C-rate, while the lower threshold value, $I_l$, equal to a current drop of 40\% from $I_h$. This allows for sufficient data to be recorded while keeping the diagnostics time to a minimum. For cells cycled with the 2-step fast charge we select the current curve in Supplementary Figure 2b. 
The obtained segments of voltage and current charge curves are further processed to obtain a plethora of features as described in Supplementary Note 3. Supplementary Table 2 summarises all features generated from processing the curves.

\textbf{Feature selection} with recursive feature elimination and cross-validation (RFE-CV) performs selection and subset reduction automatically without requirements of user-based thresholds, such as a maximum number of features to be selected. To suit battery data, we modify the original formulation by replacing the decision function algorithm with a random forest (RF) as opposed to the support vector machine (SVM) used in \cite{guyon_RFE_CV}. The replacement is motivated by RF's ability to deal with unscaled data. We call the resultant modified algorithm RF-RFE-CV. We use 700 decision tree estimators for the random forest algorithm and we set the number of cross-validations equal to the number of batteries in the feature selection dataset (see Supplementary Note 5 for data partition). We perform feature selection for each battery dataset based on a subset of the training data to avoid introducing optimistically biased performance estimates. 

Battery SOH is quantified as capacity fade with reference to the first cycle as per equation \ref{SOH}, where $C_i$ represents capacity value at $i^{th}$ cycle and $C_1$ is the capacity at the first cycle measured by a complete charge-discharge operation.  
\begin{equation}
    SOH = \frac{C_i}{C_1}
    \label{SOH}
\end{equation}

The role of the algorithm is to map from inputs $\mathbf{x}$ to target variable $y$ by means of a function $f(\mathbf{x}, \mathbf{\theta})$:
\begin{equation}
    y = f(\mathbf{x}, \mathbf{\theta} ) + \epsilon
    \label{general_model}
\end{equation}
 where $\mathbf{\theta}$ are the model weights vector and $\epsilon \sim \mathcal{N}(0, \Sigma)$ is a normally distributed noise parameter. Based on the selected algorithm, the function $f(\mathbf{x}, \mathbf{\theta})$ may take different forms based on underlying assumptions of each algorithm. The learned model can then be used to make predictions of capacity given a test vector $\mathbf{x}^*$. 
 
\textbf{Data augmentation} is carried out using the \emph{fast gradient sign method (FGSM)} in combination with the weight decay algorithm (ridge regression). We have found that a Ridge regularised model in combination with the FGSM was able to reduced the confidence interval (CI) around the estimated mean, despite being a simpler model than the original formulation in \cite{goodfellow_fgsm} which was based on a neural network. Given an input $\mathbf{x}$ with a target $y$ and loss $l(\mathbf{\theta}, \mathbf{x}, y)$, FGSM generates an adversarial example using:
\begin{equation}
    \mathbf{x}_{adv}=\mathbf{x} + \gamma \cdot sign(\nabla_x l(\theta, \mathbf{x}, y))
\end{equation}

where $\gamma$ is a small value such that the max value of the perturbation is bounded and $\nabla_x$ is the gradient with respect to $\mathbf{x}$. Because each feature in the dataset has a different range, we set $\gamma$ to 0.01 or 1\% times the range of each feature vector. The adversarial examples are concatenated with the original training data to create a comprehensive training dataset. Note, other methods for data augmentations can also be used such as the ones proposed in \cite{deep_ensamble}, \cite{adversarial_regression}, \cite{ead_attack}, \cite{ead_attack_with_L1}, however the effect of data augmentation on model performance is beyond the scope of the present work. 

The study solves eq. \ref{general_model} by making use of four algorithms as follows:

\textbf{Bayesian Ridge regression (BRR)}  considers a probabilistic model of the regression problem. The algorithm estimates a spherical Gaussian prior over the model weights given by $p(\mathbf{\theta}|\lambda) = \mathcal{N}(\mathbf{\theta}|0, \lambda^{-1} \textbf{I}_p)$, where $\lambda^{-1}$ is the precision. The priors over $\alpha$ (the regulariser) and $\lambda$ are chosen to be gamma distributions. All parameters, $\mathbf{\theta}$, $\lambda$ and $\alpha$, are jointly estimated during training as per the implementation in \cite{scikit-learn}. Posterior inference can be performed in a closed-form because the prior is conjugate. For a complete explanation of the algorithm refer to \cite{bishop2006pattern}. 

\textbf{Gaussian process regression (GPR)} is a nonparametric, Bayesian approach to regression defining a probability distribution over functions rather than random variables, thus eq. \ref{general_model} is solved by:

\begin{equation}
    f(\mathbf{x}) \sim \mathcal{GP}(m(\mathbf{x}), k(\mathbf{x}, \mathbf{x}'))
    \label{GPR}
\end{equation}

where $m(\mathbf{x})$ is the mean and $k(\mathbf{x}, \mathbf{x}')$ is the covariance function. Note, as defined above, GPR does not require learning the parameters of the regression function $f(\mathbf{x}, \mathbf{\theta} )$,  in  a  traditional sense. The mean and covariance are defined by:

\begin{equation}
    m(\mathbf{x}) = \mathbb{E}[f(\mathbf{x})]
\end{equation}
\begin{equation}
    k(\mathbf{x}, \mathbf{x}') = \mathbb{E}[(f(\mathbf{x}-m(\mathbf{x}))(f(\mathbf{x}'-m(\mathbf{x}'))]
\end{equation}

GPR assigns a prior probability to every possible function, where higher probabilities are given to functions that the algorithm considers to be more likely, for example, because they are smoother than other functions. For our implementation, we make use of the standard radial basis kernel (RBF) as detailed in \cite{gpr_mit_rasmussen}, where a mathematical explanation of the algorithm is also given. Other kernel options exist, however, we do not explore the effect of kernel choice on algorithm performance. 

\textbf{Random Forest (RF)} is a collection of constructed decision trees who sequentially conduct binary splits of the data to produce a homogeneous subset. For a comprehensive explanation of the algorithm refer to \cite{RF}. We adopt a bagging approach where the ensemble members are trained on different bootstrap samples of the training set and we set the number of decision trees in the forest to 1500. The variability of the predictions estimated by the random forest has been investigated based on the study from \cite{RF_CI}, where the confidence interval's variance has been obtained using the bootstrap replicates used to train the random forest itself.

\textbf{Deep ensemble of neural networks (dNNe)}. Ensemble methods combine different regressors into a meta-regressor and we consider an ensemble of deep neural networks as proposed in \cite{deep_ensamble}. Each network in the ensemble incorporates 2 hidden layers with an output of two layers one for the mean, $\mu(x)$ and the other for variance, $\sigma^2$ with $\sigma^2 > 0$. We use the negative log-likelihood as a function of the predicted mean and variance for scoring purposes. We also use a feed-forward architecture of 2 densely connected hidden layers. Each layer decreases in size by 50\% neurons based on the number of input features. When the input number features is less than 10, we force the network's hidden layers to 4 neurons in the first layer and 3 in the second layer. For example, when 18 input features are considered, the first hidden layer consists of 9 neurons, followed by 4 neurons in the second hidden layer. Each network used in this work has the following parameters: first hidden layer implies a ReLU activation function, followed by a Leaky ReLU for the second hidden layer and a Sigmoid function for the output layer. Additionally, we make use of Adam optimiser with a learning rate of 0.001 and a batch size equal to the number of cycles for each cell in the training set.

All models are evaluated based on mean absolute percent error (MAPE) and root mean squared error (RMSPE).

\begin{equation}
    MAPE(y_i^*, y_i) = \frac{1}{N}\sum^{N}_{i=1}\frac{|y_i^*-y_i|}{y_i}
    \label{MAPE}
\end{equation}

\begin{equation}
    RMSPE(y_i^*, y_i) = \sqrt{\frac{1}{N}\sum^{N}_{i=1}\left(\frac{y_i^*-y_i}{y_i}\right)^2}
    \label{RMSPE}
\end{equation}

where $y_i$ is the measured capacity value, $y_i^*$ is the estimated capacity value, and $n$ is the total number of samples. 

In a regression setting, we obtain probabilistic forecasts using one of the algorithms described above through the estimation of a Gaussian distribution $\mathcal{N}(\mu, \sigma^2)$, where $\mu$ is the mean estimated capacity and $\sigma^2$ is the associated uncertainty quantified as variance. To evaluate the usefulness of predictive uncertainty for decision making, we create reliability diagnostics curves analogous to the work in \cite{calibration_DL_kuleshov}.  To plot calibration curves, we divide each predicted confidence interval in $m$ confidence levels that are monotonically increasing on the interval $[0, 1]$ i.e. $0<p_1<p_2<...<p_m<1$. We then compute the empirical probability for each threshold by counting the frequency of true labels in each confidence level $p_m$. Mathematically this can be summarised as:
\begin{equation}
    \hat{p}_m = \frac{|{y_n}|F_n(y_n)\leq p_m, n=1,...,N|}{N}
\end{equation}

Based on the reliability curve assessment, we then perform re-calibration using isotonic regression \cite{isotonic_regression}. A well-calibrated regressor should lie very close to the ideal diagonal curve, e.g. results Figure \ref{calibration_group1_dNNE}. We use the calibration score($C_{score}$) as a numerical score to describe the quality of the calibration when referenced to a 90\% confidence interval and sharpness ($Sh$) to describe average standard deviation. 

\begin{equation}
    C_{score} = \frac{1}{N}\sum_{i=1}^{N} \hat{p}_{m=90\%}
    \label{c_error}
\end{equation}

Sharpness is calculated as an average of model output variance for each prediction and is given by:
\begin{equation}
    Sh = \frac{1}{n} \sum_{i=1}^{n} \sigma_i
    \label{sh}
\end{equation}

where $i$ is the sample number and $n$ is the the total number of sampels.

\begin{figure}[h!]
    \centering
    \includegraphics[width = .4\textwidth]{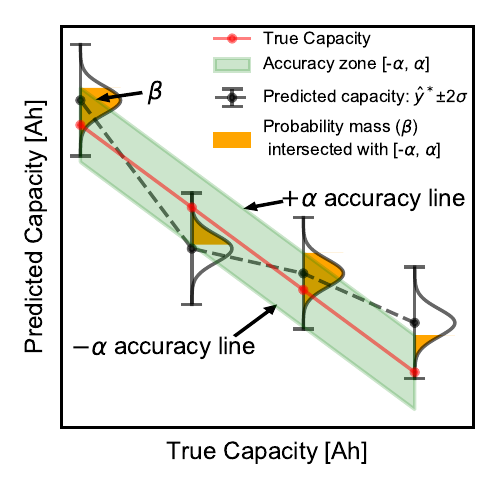}
    \captionof{figure}{$\alpha$-accuracy zone and $\beta$ probability mass illustration.}
    \label{accuracy_zone}
\end{figure}

We further propose an assessment of uncertainty prediction via prognostics performance metrics from an engineering point of view, adopted from \cite{prog_metrics_nasa}. First, we introduce the accuracy zone defined by a threshold, $\alpha$ (see Figure \ref{accuracy_zone}), which is calculated as a percentage error from the true capacity value, i.e. $ y \pm \alpha $. We select an $\alpha$ of $\pm$1.5\% (alpha can be adjusted based on intended application). Based on the frequency of predicted values residing in the accuracy zone, we calculate the $\alpha$-accuracy. Finally, we calculate the average probability mass of the prediction PDF within the $\alpha$ bounds called $\beta$, refer to Figure \ref{accuracy_zone}. Ideally, $\beta$ should be one, suggesting that the predicted confidence interval is small and encapsulates the entire $\alpha$-accuracy zone. Since $\alpha$ summarises the notion of \emph{desired accuracy}, $\alpha^+$ is the upper bound for estimates above the accuracy zone, and $\alpha^-$ represents low estimates or value residing under the desired accuracy zone. Depending on the application, both or any one of the low or high estimates may be undesirable. We chose to calculate the percentage of early predictions (estimates residing below the true label, the red line in Figure \ref{accuracy_zone}), denoted here by PEP, as a measure of algorithm uncertainty in a critical application scenario. 

\section{Dataset}

\begin{table*}[h!]
\centering
\resizebox{\textwidth}{!}{%
\begin{tabular}{@{}cccccccc@{}}
\toprule
Group{*}                       & I     & I     & I     & I               & I               & II               & III                  \\ \hline
Dataset  & CALCE CS2 & CALCE CX2 & CALCE PL & NASA 5 & NASA11 & TRI & Oxford \\ \midrule
Manufacturer & Unknown  & Unknown  & Unknown  & LG Chem              & LG Chem              & A123 Systems             & Kokam                 \\ \midrule
Cathode \textsuperscript{***}  & LiCoO$_{2}$ & LiCoO$_{2}$ & LiCoO$_{2}$ & LiCoO$_{2}$             & LiCoO$_{2}$             & LiFePO$_{4}$             & \begin{tabular}[c]{@{}c@{}}LiCoO$_{2}$ /\\ LiNiMnCoO$_{2}$\end{tabular} \\ \midrule
Form factor & Prismatic & Prismatic & Pouch  & \begin{tabular}[c]{@{}c@{}}18650\\ Cylindrical\end{tabular} & \begin{tabular}[c]{@{}c@{}}18650\\ Cylindrical\end{tabular} & \begin{tabular}[c]{@{}c@{}}18650\\ Cylindrical\end{tabular} & Pouch                 \\ \midrule
\# cells  & 6   & 6   & 2   & 8               & 25               & 124               & 8                  \\ \midrule
Charge  & CC-CV  & CC-CV  & CC-CV  & CC-CV              & CC-CV              & Fast-charge             & CC                  \\ \midrule
Discharge & 2 regimes & 2 regimes & 1 regime & 2 regimes             & 7 regimes             & 1 regime              & 1 regime                 \\ \midrule

\multicolumn{8}{l}{ \begin{tabular}[c]{@{}l@{}}\textsuperscript{*}\footnotesize{Groups based on charge protocol}, \textsuperscript{**}\footnotesize{Toyota Research Institute}, \textsuperscript{***}\footnotesize{Information from manufacturer, not verified}\end{tabular}}                                        
\end{tabular}%
}
\caption{Datasets overview. Note: refer to Supplementary Note 4 for data sources.}
\label{data_rep}
\end{table*}
 
We investigate the performance of BHUMP on a total of 179 Li-ion cells as referenced in Table \ref{data_rep}. The cells have been grouped into three categories based on the charging protocol used: constant current - constant voltage (CC-CV) protocol in Group I (47 cells),  2-step fast charge protocol in Group III (8 cells), and constant current (CC) protocol in Group II (124 cells). The separation is important for separate model training and feature selection, as well as model performance assessment on different charge protocols. A detailed explanation of each dataset used can be found in Supplementary Note 4.

\section{Algorithm performance}

\subsubsection{Group I data}

Subject to the previously described pipeline steps the feature selection algorithm, RF-RFE-CV chose 18 of the 30 engineered features as the optimum number of attributes for the cells in Group I (refer to Supplementary Figure 8a and Supplementary Table 3). From a threshold point of view, we select a $V_h$ of 4.2V for all batteries in this Group with an associated $V_l$ of 3.9V. Refer to Supplementary Note 5 for train/test partitions. 

\begin{table}[h!]
\centering
\resizebox{.47\textwidth}{!}{%
\begin{tabular}{l|c|c|c|c|c|c|c|}
\cline{2-8}
       & \multicolumn{1}{l|}{MAPE} & \multicolumn{1}{l|}{RMSPE} & \multicolumn{1}{l|}{$C_{score}$} & \multicolumn{1}{l|}{$Sh$} & \multicolumn{1}{l|}{$\alpha$-accuracy} & \multicolumn{1}{l|}{$\beta$} & \multicolumn{1}{l|}{$PEP$} \\ \hline
\multicolumn{1}{|l|}{BRR} & 1.52 & 2.49 & 84.49 & 0.021 & 70.00 & 0.57 & 68.92 \\ \hline
\multicolumn{1}{|l|}{GPR} & 1.49 & 2.24 & 92.23 & 0.025 & 65.00 & 0.48 & 71.76 \\ \hline
\multicolumn{1}{|l|}{RF} & 0.72 & 0.91 & 100 & 0.046 & 92.00 & 0.29 & 95.29 \\ \hline
\multicolumn{1}{|l|}{dNNe} & 0.65 & 0.92 & 88.01 & 0.0082 & 93.00 & 0.93 & 97.71\\ \hline
\end{tabular}%
}
\caption{Results for Group I cell no. 38.}
\label{results_group1_cell_38}
\end{table}

We illustrate results for BHUMP when dNNe is considered as base algorithm in Figure \ref{predictions_group1_dNNE} (results for all other algorithms are shown in Supplementary Figures 11, 12, 13) for a randomly chosen pouch cell battery, cell no. 38 and summarise algorithm performance on this cell in Table \ref{results_group1_cell_38}. The cell was cycled in full depth of discharge between 4.2V to 2.7V at a discharge C-rate of 0.5C (or 0.55 A) with a CC-CV charging protocol at a current value of 0.5 C-rate. Table \ref{results_group1} summarises each algorithms' performance on cell no. 38. Comparing dNNe in Figure~\ref{pred_group1_dNNe} to the other algorithms BRR, GPR, and RF, we show that the resultant confidence interval is considerably smaller (all figures display a confidence level equivalent to a 95\% quantile i.e. $ \mu \pm 2\cdot \sigma$). This indicates that the model is sharper, resulting in a high $\beta$ score (refer to Table \ref{results_group1_cell_38} for results). Where the predictions are less accurate, such as is the prediction in the first few cycles (see Figure~\ref{pred_group1_dNNe}), the error bars capture this variability well. On this battery, dNNe also achieves the best RMSPE and MAPE together with a high calibration score. As per Table \ref{results_group1_cell_38}, the estimates for this cell vary between RMSPE 0.65\% to 1.52\%, showing that all 4 algorithms can achieve high performance. The same conclusion is not valid for calibration, however. Reliability plots indicate that RF exhibits high variance even after calibration, refer to Supplementary Figure 13.

\begin{figure*}[h!]
  \centering
\begin{subfigure}[b]{.55\textwidth}
     \centering
     \caption{}
                    \includegraphics[width=\linewidth]{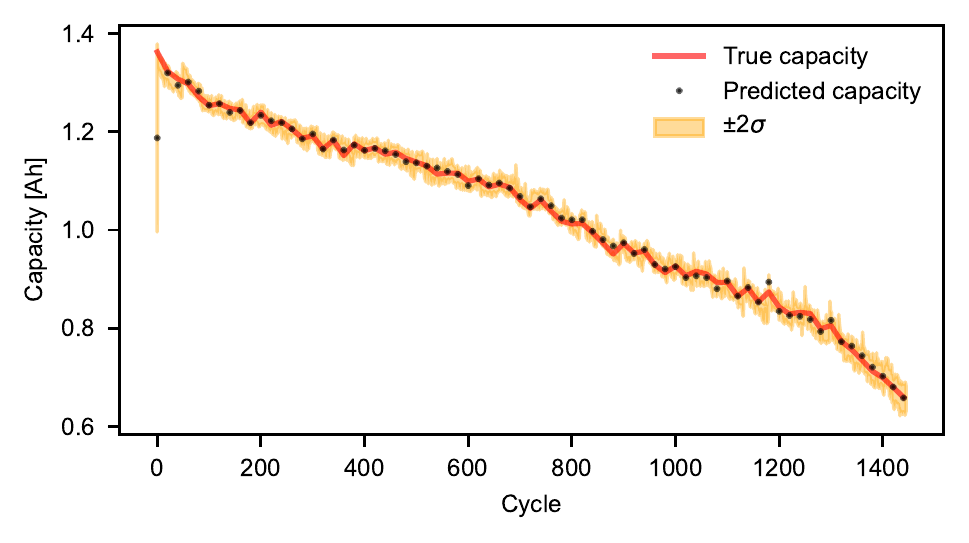}
     \label{pred_group1_dNNe}
\end{subfigure}%
\begin{subfigure}[b]{.3\textwidth}
     \centering
     \caption{}
                    \includegraphics[width=\linewidth]{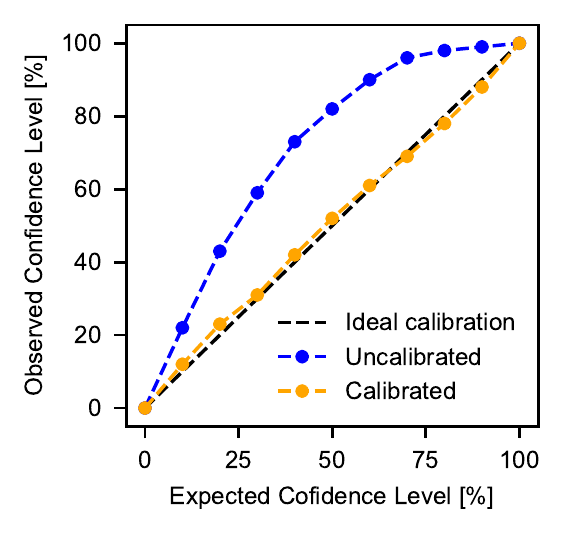}
     \label{calibration_group1_dNNE}
  \end{subfigure}%
  \hfill 
  \begin{subfigure}[b]{.55\textwidth}
     \centering
     \caption{}
                    \includegraphics[width=\linewidth]{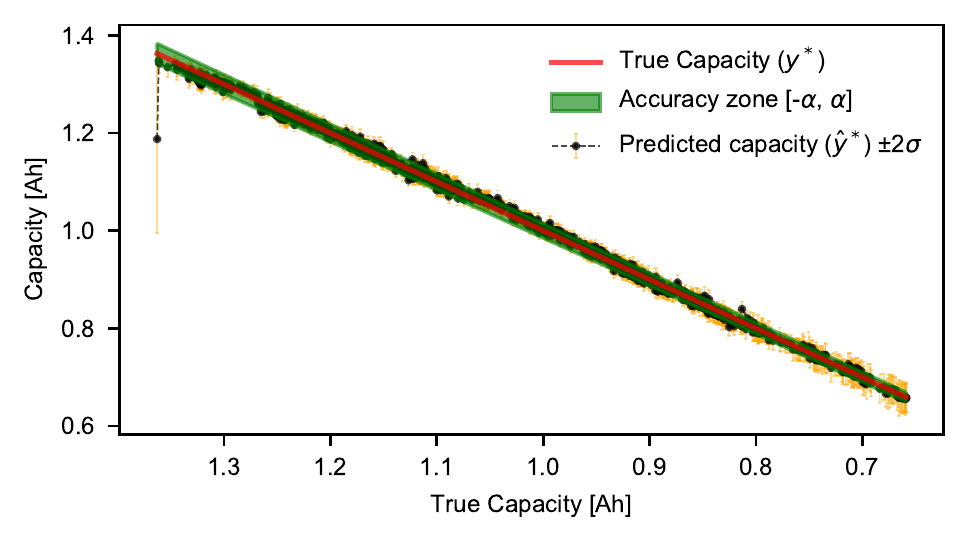}
     \label{prediction_vs_true_dNNe_group1}
\end{subfigure}%
\begin{subfigure}[b]{.3\textwidth}
     \centering
     \caption{}
                    \includegraphics[width=\linewidth]{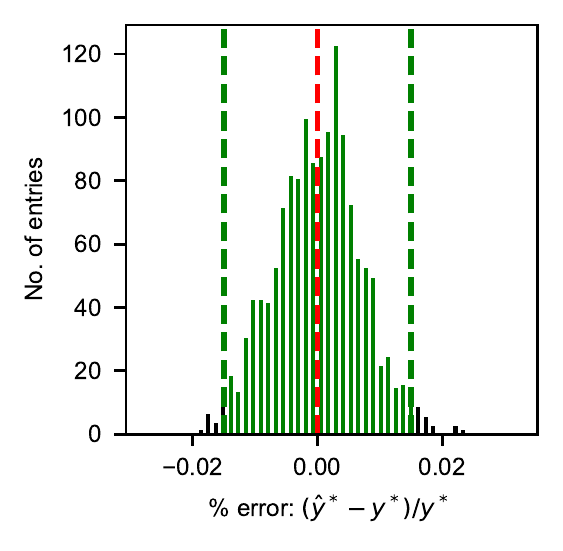}
     \label{hist_group1_dNNE}
  \end{subfigure}%
    \hfill
  \caption{\textbf{Prediction results with dNNe Group I cell no. 38.} \textbf{\protect\subref{pred_group1_dNNe}} dNNe prediction as a function of cycle, \textbf{\protect\subref{calibration_group1_dNNE}} dNNE calibration results, \textbf{\protect\subref{prediction_vs_true_dNNe_group1}} dNNe actual vs. predicted capacity, \textbf{\protect\subref{hist_group1_dNNE}} Histogram of \% error. Note: $y^*$ - true capacity, $\hat{y}^*$ - predicted capacity} 
  \label{predictions_group1_dNNE}
 \end{figure*}

When discussing average results across all cells in Group I (Table \ref{results_group1}), RF achieves on average a low calibration error of 54.70\% possibly due to the method used for estimating the variance, Infinitesimal Jackknife. In practice we prefer a more conservative system, particularly in safety-critical applications. This implies that the number of capacity estimates lower than the true label residing in the $\alpha$-accuracy zone (Figure \ref{accuracy_zone}) should exceed the number of capacity values estimated above it i.e. PEP should be close to 100\%. At the same time, too low of a capacity estimate would result in a far too conservative algorithm. However, such behaviour is captured by an increase in RMSPE and thus mitigated for naturally. With reference to Figure \ref{hist_group1_dNNE} together with Table \ref{results_group1_cell_38} one can conclude that dNNe is conservative, achieving the highest PEP.

\begin{table}[h!]
\centering
\resizebox{.47\textwidth}{!}{%
\begin{tabular}{l|c|c|c|c|c|c|c|}
\cline{2-8}
       & \multicolumn{1}{l|}{MAPE} & \multicolumn{1}{l|}{RMSPE} & \multicolumn{1}{l|}{$C_{score}$} & \multicolumn{1}{l|}{$Sh$} & \multicolumn{1}{l|}{$\alpha$-accuracy} & \multicolumn{1}{l|}{$\beta$} & \multicolumn{1}{l|}{$PEP$} \\ \hline
\multicolumn{1}{|l|}{BRR} & 4.65 & 5.54 & 89.16 & 0.104 & 25.76 & 0.25 & 36.57 \\ \hline
\multicolumn{1}{|l|}{GPR} & 3.70 & 4.51 & 83.62 & 0.089 & 32.04 & 0.29 & 60.07 \\ \hline
\multicolumn{1}{|l|}{RF} & 2.17 & 2.70 & 54.70 & 0.093 & 35.94 & 0.36 & 65.47 \\ \hline
\multicolumn{1}{|l|}{dNNe} & 3.30 & 4.26 & 86.28 & 0.043& 32.14 & 0.58 & 63.26 \\ \hline
\end{tabular}%
}
\caption{Average results over Group I dataset. }
\label{results_group1}
\end{table}

Overall, despite RF achieving the lowest average RMSPE and MAPE (Table \ref{results_group1}) it does not output well-calibrated predictions, nor it displays a high sharpness value. At the expense of 1.13\% in MAPE and 1.56\% in RMSPE, the dNNe outputs a well-calibrated model, on average being less than 4\% under the ideal calibration score.

\subsubsection{Group II data}

 Group II dataset is the largest dataset incorporating a total of 124 cells. While the dataset exhibits a high variance in charge profiles, it does not have any variation in discharge conditions (all cells in the dataset are discharged at 4 C-rate). This, in turn, showcases the effect of the charge profile on the estimation accuracy of the 4 algorithms. Training is performed on features engineered based on the CC-CV curve obtained after the cell reaches 80\% SOC (refer to Supplementary Figures 2a and 2b). Refer to Supplementary Note 5 for train/test partitions. RF-RFE-CV selects a total of 5 features (Supplementary Figure 8b and Supplementary Table 4) out of a total of 30 engineered features. We believe this is caused by the fact that the dataset only incorporates one discharge profile as well as just a single battery type.

\begin{table}[ht!]
\centering
\resizebox{.47\textwidth}{!}{%
\begin{tabular}{l|c|c|c|c|c|c|c|}
\cline{2-8}
       & \multicolumn{1}{l|}{MAPE} & \multicolumn{1}{l|}{RMSPE} & \multicolumn{1}{l|}{$C_{score}$} & \multicolumn{1}{l|}{$Sh$} & \multicolumn{1}{l|}{$\alpha$-accuracy} & \multicolumn{1}{l|}{$\beta$} & \multicolumn{1}{l|}{$PEP$} \\ \hline
\multicolumn{1}{|l|}{BRR} & 0.72 & 0.90 & 65.49 & 0.005 & 89.00 & 98.00 & 20.70 \\ \hline
\multicolumn{1}{|l|}{GPR} & 1.23 & 1.63 & 69.94 & 0.011 & 65.00 & 85.00 & 22.16 \\ \hline
\multicolumn{1}{|l|}{RF} & 0.23 & 0.43 & 87.42 & 0.002 & 98.00 & 100 & 42.81 \\ \hline
\multicolumn{1}{|l|}{dNNe} & 0.34 & 0.48 & 71.31 & 0.002 & 98.00 & 100 & 31.50 \\ \hline
\end{tabular}%
}
\caption{Results for Group II cell no. 1.}
\label{results_group2_cell_1}
\end{table}

\begin{figure*}[h!]
  \centering
\begin{subfigure}[b]{.55\textwidth}
     \centering
     \caption{}
                    \includegraphics[width=\linewidth]{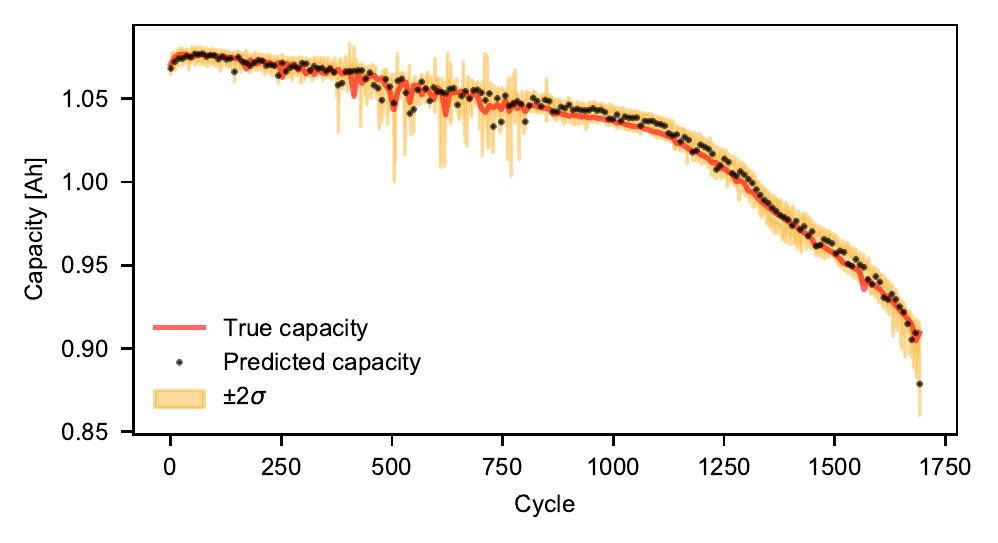}
     \label{pred_group2_dNNe}
\end{subfigure}%
\begin{subfigure}[b]{.3\textwidth}
     \centering
     \caption{}
                    \includegraphics[width=\linewidth]{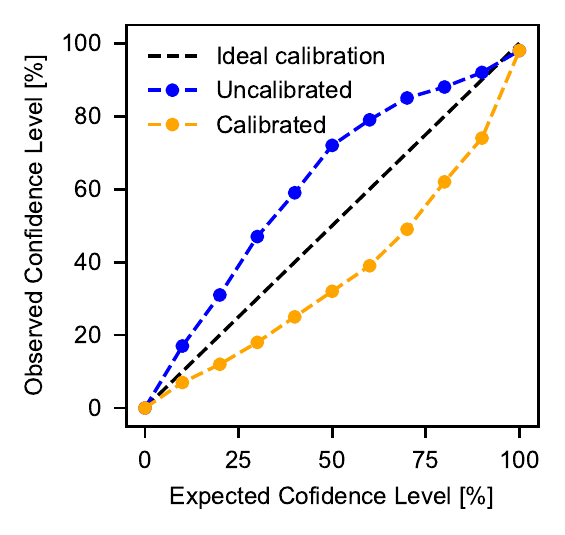}
     \label{calibration_group2_dNNE}
  \end{subfigure}%
  \hfill 
  \begin{subfigure}[b]{.55\textwidth}
     \centering
     \caption{}
                    \includegraphics[width=\linewidth]{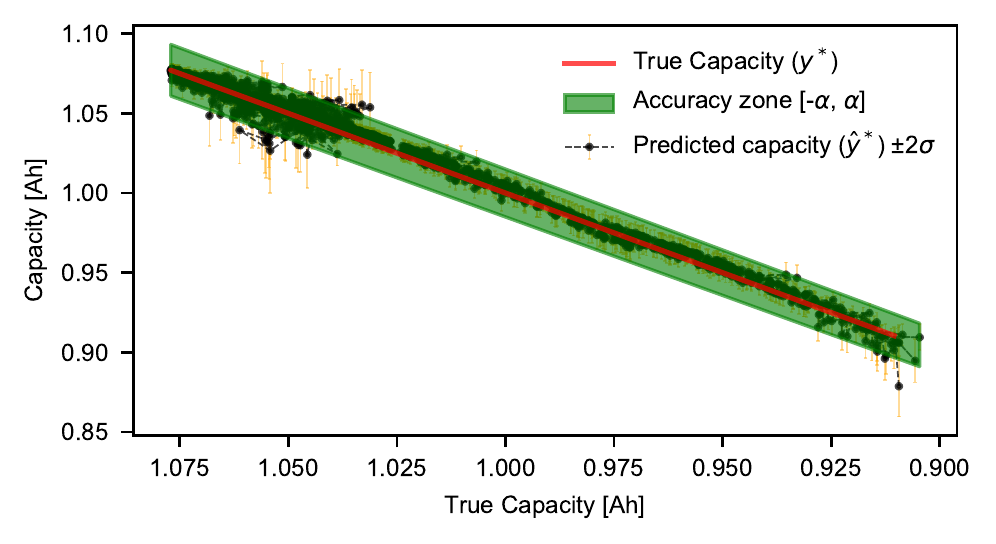}
     \label{prediction_vs_true_dNNe_group2}
\end{subfigure}%
\begin{subfigure}[b]{.3\textwidth}
     \centering
     \caption{}
                    \includegraphics[width=\linewidth]{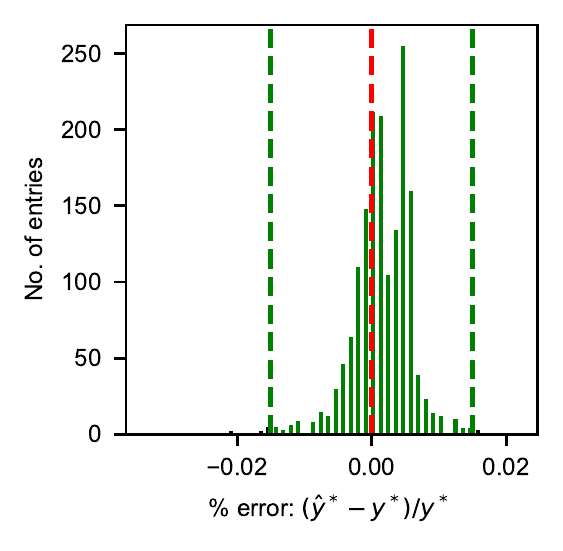}
     \label{hist_group2_dNNE}
  \end{subfigure}%
    \hfill
  \caption{\textbf{Prediction results with dNNe Group II cell no. 1.} \textbf{\protect\subref{pred_group2_dNNe}} dNNe prediction as a function of cycle, \textbf{\protect\subref{calibration_group2_dNNE}} dNNE calibration results, \textbf{\protect\subref{prediction_vs_true_dNNe_group2}} dNNe actual vs. predicted capacity, \textbf{\protect\subref{hist_group2_dNNE}} Histogram of \% error. Note: $y^*$ - true capacity, $\hat{y}^*$ - predicted capacity} 
  \label{predictions_group2_dNNE}
 \end{figure*}

Figure \ref{predictions_group2_dNNE} illustrate BHUMP performance with a dNNe as base algorithm for cell no. 1, whilst Suplementary Figures 14, 15, 16 summarise results for all other algorithms. The cell has undergone fast charge profile of 3.6 C-rate to 80\% SOC, beyond which the cell is charged with CC of 1C followed by the CV charging. The reason cell 1 was selected in this case was to illustrate the performance of the algorithms when there is a high number of outliers in capacity data (Figure \ref{pred_group2_dNNe}). With reference to Table \ref{results_group2_cell_1}, RF achieves lowest error and highest scores as well as a good calibration compared to all other algorithms. On this particular cell, dNNe achieves the second best performance, however it does not output a well calibrated model, despite showing a good average calibration score as per Table \ref{results_group2}.

Average results of the 4 algorithms are concisely summarised in Table \ref{results_group2}. All models are able to estimate the SOH with less than 2\% RMPSE; this underlines the fact that the models are not affected by the fast-charge section of the charging protocol. RF achieves the highest accuracy with a low sharpness value and high percentages for all other metrics except for calibration where it exhibits over-confidence. In terms of calibration error, dNNe achieves the closest score to a 90\% confidence interval with 91.02\%. dNNe is also the second-best performing algorithm achieving good scores across all metrics as summarised in Table \ref{results_group2_cell_1}. In comparison, the two Bayesian-based algorithms exhibit a higher percentage error as well as higher sharpness values. However, they tend to be more conservative, averaging a PEP over 60\%.

\begin{table}[h!]
\centering
\resizebox{.47\textwidth}{!}{%
\begin{tabular}{l|c|c|c|c|c|c|c|}
\cline{2-8}
       & \multicolumn{1}{l|}{MAPE} & \multicolumn{1}{l|}{RMSPE} & \multicolumn{1}{l|}{$C_{score}$} & \multicolumn{1}{l|}{$Sh$} & \multicolumn{1}{l|}{$\alpha$-accuracy} & \multicolumn{1}{l|}{$\beta$} & \multicolumn{1}{l|}{$PEP$} \\ \hline
\multicolumn{1}{|l|}{BRR} & 0.45 & 0.76 & 91.72 & 0.005 & 97.31 & 99.19 & 62.86 \\ \hline
\multicolumn{1}{|l|}{GPR} & 1.00 & 1.91 & 93.14 & 0.012 & 90.43 & 83.74 & 63.21 \\ \hline
\multicolumn{1}{|l|}{RF} & 0.11 & 0.14 & 79.72 & 0.001 & 99.84 & 99.96 & 58.77 \\ \hline
\multicolumn{1}{|l|}{dNNe} & 0.23 & 0.45 & 91.02 & 0.002 & 99.53 & 99.50 & 53.41 \\ \hline
\end{tabular}%
}
\caption{Average results over Group II dataset}
\label{results_group2}
\end{table}

In conclusion, from an accuracy and sharpness perspective, the best performing algorithm on dataset Group II is RF, whilst the poorest performance is achieved by GPR. When it comes to uncertainty metrics, and in particular calibration, RF exhibits over-confidence with a $C_{score}$ of 79.72\%. Such behaviour is also identified in Group I dataset where RF was, in fact, difficult to calibrate despite the rich dataset. A more reliable calibration score is achieved by dNNe at the expense of a loss of 0.12\% in MAPE and 0.31\% in RMSPE (refer to Table \ref{results_group2}).

\subsubsection{Group III data}

On Group III  we emphasise on the suitability of BHUMP to battery state of health estimation for automotive applications. Group III includes 8 Kokham 740 mAh batteries that have been dynamically cycled under the ARTEMIS \cite{andre2004artemis} dynamic driving
profile, followed by characterisation cycles. Each characterisation cycle consists of low rate CC charge and discharge cycles, repeated every 100 cycles. We use the characterisation cycles for diagnostics purposes to derive features and estimate battery health. This dataset incorporates the lowest variability both in terms of input feature values and capacity degradation values due to the identical charge-discharge conditions. This, in turn, affects feature selection as BHUMP only selects 5 out of the 18 engineered features (note charge protocol does not include CV part of the charge, hence 12 features are missing) as shown in Supplementary Figure 8c and Supplementary Table 5. We keep the same threshold values as in Group I cells for the CC part of the curves, namely a $V_h$ of 4.2V and a $V_l$ of 3.9V on which feature are engineered. Refer to Supplementary Note 5 for train/test partitions.

\begin{figure*}[h!]
  \centering
\begin{subfigure}[b]{.55\textwidth}
     \centering
     \caption{}
                    \includegraphics[width=\linewidth]{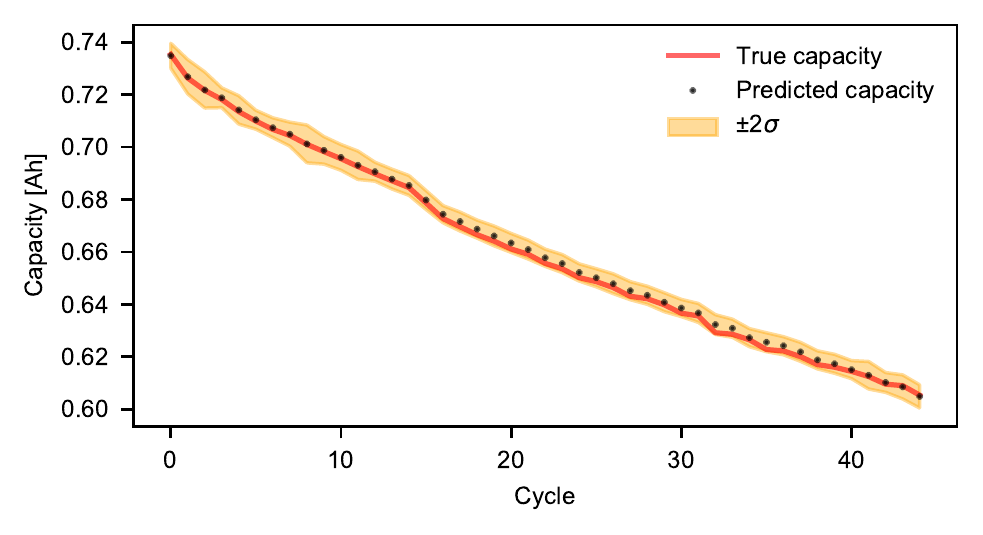}
     \label{pred_group3_dNNe}
\end{subfigure}%
\begin{subfigure}[b]{.3\textwidth}
     \centering
     \caption{}
                    \includegraphics[width=\linewidth]{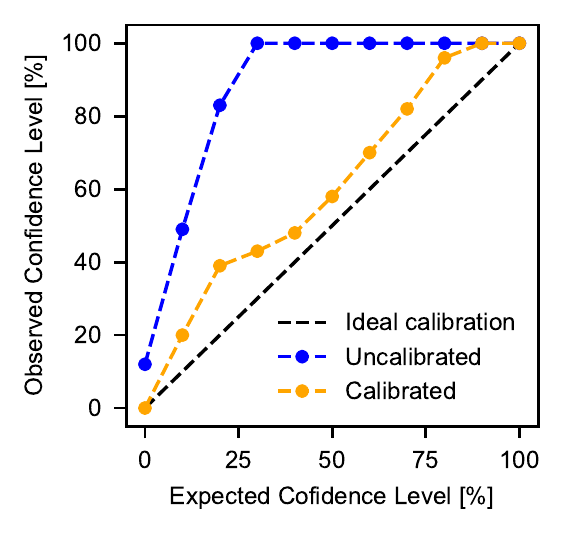}
     \label{calibration_group3_dNNE}
  \end{subfigure}%
  \hfill 
  \begin{subfigure}[b]{.57\textwidth}
     \centering
     \caption{}
                    \includegraphics[width=\linewidth]{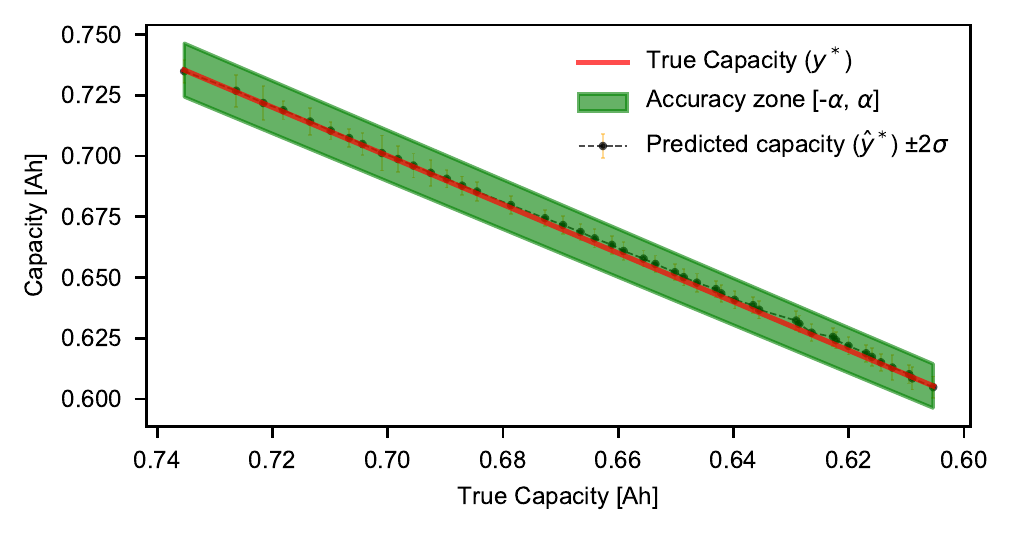}
     \label{prediction_vs_true_dNNe_group3}
\end{subfigure}%
\begin{subfigure}[b]{.3\textwidth}
     \centering
     \caption{}
                    \includegraphics[width=\linewidth]{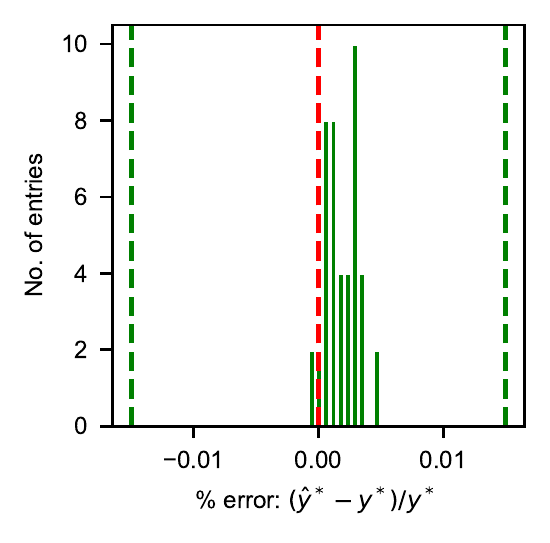}
     \label{hist_group3_dNNE}
  \end{subfigure}%
    \hfill
  \caption{\textbf{Prediction results with dNNe Group III cell no. 5.} \textbf{\protect\subref{pred_group3_dNNe}} dNNe prediction as a function of cycle, \textbf{\protect\subref{calibration_group3_dNNE}} dNNE calibration results, \textbf{\protect\subref{prediction_vs_true_dNNe_group3}} dNNe actual vs. predicted capacity, \textbf{\protect\subref{hist_group3_dNNE}} Histogram of \% error. Note: $y^*$ - true capacity, $\hat{y}^*$ - predicted capacity} 
  \label{predictions_group3_dNNE}
 \end{figure*}

\begin{table}[ht!]
\centering
\resizebox{.47\textwidth}{!}{%
\begin{tabular}{l|c|c|c|c|c|c|c|}
\cline{2-8}
       & \multicolumn{1}{l|}{MAPE} & \multicolumn{1}{l|}{RMSPE} & \multicolumn{1}{l|}{$C_{score}$} & \multicolumn{1}{l|}{$Sh$} & \multicolumn{1}{l|}{$\alpha$-accuracy} & \multicolumn{1}{l|}{$\beta$} & \multicolumn{1}{l|}{$PEP$} \\ \hline
\multicolumn{1}{|l|}{BRR} & 0.11 & 0.15 & 95.55 & 0.89 & 100 & 100 & 31.11 \\ \hline
\multicolumn{1}{|l|}{GPR} & 0.16 & 0.19 & 71.11 & 1.21 & 100 & 100 & 15.55 \\ \hline
\multicolumn{1}{|l|}{RF} & 0.17 & 0.21 & 97.77 & 2.01 & 100 & 100 & 24.44 \\ \hline
\multicolumn{1}{|l|}{dNNe} & 0.20 & 0.25 & 100.00 & 2.93 & 100 & 100 & 6.67 \\ \hline
\end{tabular}%
}
\caption{Results for Group III cell no. 5.}
\label{results_group3_cell_5}
\end{table}

For visualisation purposes, we illustrate results for the randomly selected cell no. 5 for dNNe in Figure \ref{predictions_group3_dNNE} and Supplementary Figures 17, 18, 19 for all other algorithms. It is clear, from Table \ref{results_group3_cell_5} that performance on cell 5 is dominated by BRR based on all measures of accuracy and uncertainty quantification. However, all algorithms deployed on cell no. 5 (Table \ref{results_group3_cell_5}) achieve a MAPE and RMSPE smaller than the proposed accuracy zone threshold $\alpha$ of $\pm$1.5\%.

Average results are summarised in Table \ref{results_group3}. In terms of accuracy measures, on average, BRR outperforms all other methods, including the dNNe. As argued in \cite{ann_sample_size} linear regression outperforms considerably more complex algorithms, including NNs when dealing with small sample size that exhibits little variance. Despite the low error, BRR does not achieve a good calibration score as opposed to dNNE. dNNe is the second-best performing algorithm in terms of accuracy (MAPE and RMSPE). It also exhibits adequate results for all other metrics, including $PEP$ where it scores the highest.

\begin{table}[h!]
\centering
\resizebox{.47\textwidth}{!}{%
\begin{tabular}{l|c|c|c|c|c|c|c|}
\cline{2-8}
       & \multicolumn{1}{l|}{MAPE} & \multicolumn{1}{l|}{RMSPE} & \multicolumn{1}{l|}{$C_{score}$} & \multicolumn{1}{l|}{$Sh$} & \multicolumn{1}{l|}{$\alpha$-accuracy} & \multicolumn{1}{l|}{$\beta$} & \multicolumn{1}{l|}{$PEP$} \\ \hline
\multicolumn{1}{|l|}{BRR} & 0.26 & 0.32 & 68.11 & 1.20 & 100 & 100 & 23.54 \\ \hline
\multicolumn{1}{|l|}{GPR} & 0.52 & 0.65 & 42.42 & 2.37 & 90.50 & 97.25 & 23.22 \\ \hline
\multicolumn{1}{|l|}{RF} & 0.36 & 0.44 & 72.62 & 2.16 & 88.5 & 100 & 25.44 \\ \hline
\multicolumn{1}{|l|}{dNNe} & 0.30 & 0.39 & 91.17 & 2.01 & 98.25 & 99.75 & 27.95 \\ \hline
\end{tabular}%
}
\caption{Average results over Group III dataset}
\label{results_group3}
\end{table}

In conclusion, when considering average results over all 4 test cells as referenced in Table \ref{results_group3}, dNNe achieves second-best accuracy while attaining the best calibration score of 91.17\%.

\section{Discussion on practical applicability of BHUMP}

BHUMP can complement battery management systems (BMS), for both SOC and SOH estimation, and replace the traditional ECMs altogether. While conventional approaches rely on measuring the capacity in static conditions such as full charge-discharge, BHUMP can estimate capacity fade from sections of the charge profile, accommodating for partial discharge scenarios or various operating conditions such as random or high discharge rates. We succinctly summarised in the results section, BHUMP can estimate capacity fade under fast charging protocol (Group II data) as well as random discharge (Group III data cycled under ARTEMIS driving protocol) typical to the operation of an EV battery pack. Future work could further extend to other charge-discharge protocols and open-source datasets such as the one in \cite{attia2020closed}.

Temperature variations during charging could further introduce uncertainty into the measurement of charge curves and propagate it into the estimation algorithm. Possible mitigation includes the use of temperature as an input when training BHUMP or considering additionally in-situ or operando sensory information such as optical and digital images or X-ray \cite{handoko2018understanding} such that the algorithm learns the correlation between temperature, generated features and SOH indicator. Due to such variations, SOH assessment without corresponding measures of algorithm uncertainty does not provide sufficient information to form a decision or corrective action. In addition to inherent algorithm bias, dataset variability also seems to affect the prediction error. To accommodate for such variations in the data BHUMP introduces 30 engineered features and makes use of an unsupervised feature selection algorithm (RF-RFE-CV). Given a training dataset RF-RFE-CV selects a subset of input features, indicating that features must be selected based on intended application, battery design and charge protocol. Despite such dataset variations, we think that deep learning has the potential to exceed it in the future as it requires little tuning from the user and can take advantage of parallelisation and an increasing amount of computational capabilities by deployment on graphics processing units (GPU) and modern data storage solutions. In addition, when training data consists of limited samples or training data is not relevant to the intended application, transfer learning can be used to reduce prediction errors. New hardware, architectures and learning algorithms that are currently being developed for neural network implementation will only accelerate this process, allowing for active learning techniques to be used when deployed onboard a vehicle. More concretely, BHMUP with dNNe as the base algorithm can incorporate transfer learning when trained on a particular cell design and re-trained on a reduced sample set for a different cell design. Additionally, BHUMP can also incorporate active learning as data becomes available when deployed online on different cell design, chemistry or operating temperature.

\section{Conclusion} 

The two widely adopted modelling techniques for online battery state of health (SOH) estimation are equivalent circuit models and electrochemical models. However, when deployed online, the trade-off between accuracy and computational efficiency is difficult to achieve. This paper introduced an alternative, machine learning-based solution called battery health and uncertainty management pipeline (BHUMP).  The pipeline provides a set of benefits over conventional methods including adaptability to the charging protocols and the discharge current rates, and prediction without knowledge of battery design, chemistry, and operating temperature.

The paper explores four algorithms: Bayesian ridge regression (BRR), Gaussian process regression (GPR), random forest (RF), and a deep ensemble of neural networks (dNNe), as the base algorithm for BHUMP. All algorithms are assessed on error values and the ability to quantify uncertainty. Results indicate that the lowest error achieved depends on the charging protocol adopted. The lowest error was achieved by RF for constant current - constant voltage protocol and fast charge protocol,  and BRR for the constant-current protocol. When considering uncertainty assessment metrics, however, RF is hard to calibrate and is overly optimistic in its predictions. At the expense of an average increase in MAPE of 0.43\% and RMSPE of 0.97\%, dNNe, generally achieves a better calibration score, consistently achieving the second-lowest error irrespective of charge protocol. On the fast-charging protocol, the best dNNe model achieved a RMSPE of 0.45\% with a calibration score of 91.02\% when referenced to a 90\% confidence interval.

Overall, our work highlights the value of coupling machine learning tools with charge curve segments in capturing battery degradation in under 15 minutes. Moreover, we argue that despite achieving low errors, any algorithm must undergo uncertainty quantification checks before deployment in the field. Finally, we show how the use of machine learning pipelines can achieve a computationally efficient and accurate solution for cell SOH estimation. We envision machine learning pipelines to be a standard technique used in designing and implementing battery management systems of the future.

\section*{Data availability}
The datasets used in this study are available at: 
\begin{itemize}
    \item Group 1: 
    
    \url{https://web.calce.umd.edu/batteries/data.htm}
    
    \url{https://ti.arc.nasa.gov/tech/dash/groups/pcoe/prognostic-data-repository/} 
    \item Group 2: 
    
    \url{https://data.matr.io/1/projects/5c48dd2bc625d700019f3204}
    \item Group 3: 
    
    \url{https://ora.ox.ac.uk/objects/uuid:03ba4b01-cfed-46d3-9b1a-7d4a7bdf6fac}
\end{itemize}

\section*{Code availability}
\gappto{\UrlBreaks}{\UrlOrds}
Code for the data processing is available from the corresponding authors upon request. Code for the modelling work is available at: \url{http://doi.org/10.5281/zenodo.4390152}

\bibliographystyle{unsrt}  

\bibliographystyle{unsrt}  



\clearpage
\section*{Supplementary material}

\begin{figure}[h!]
 		\centering
 		\includegraphics[width= .95\linewidth]{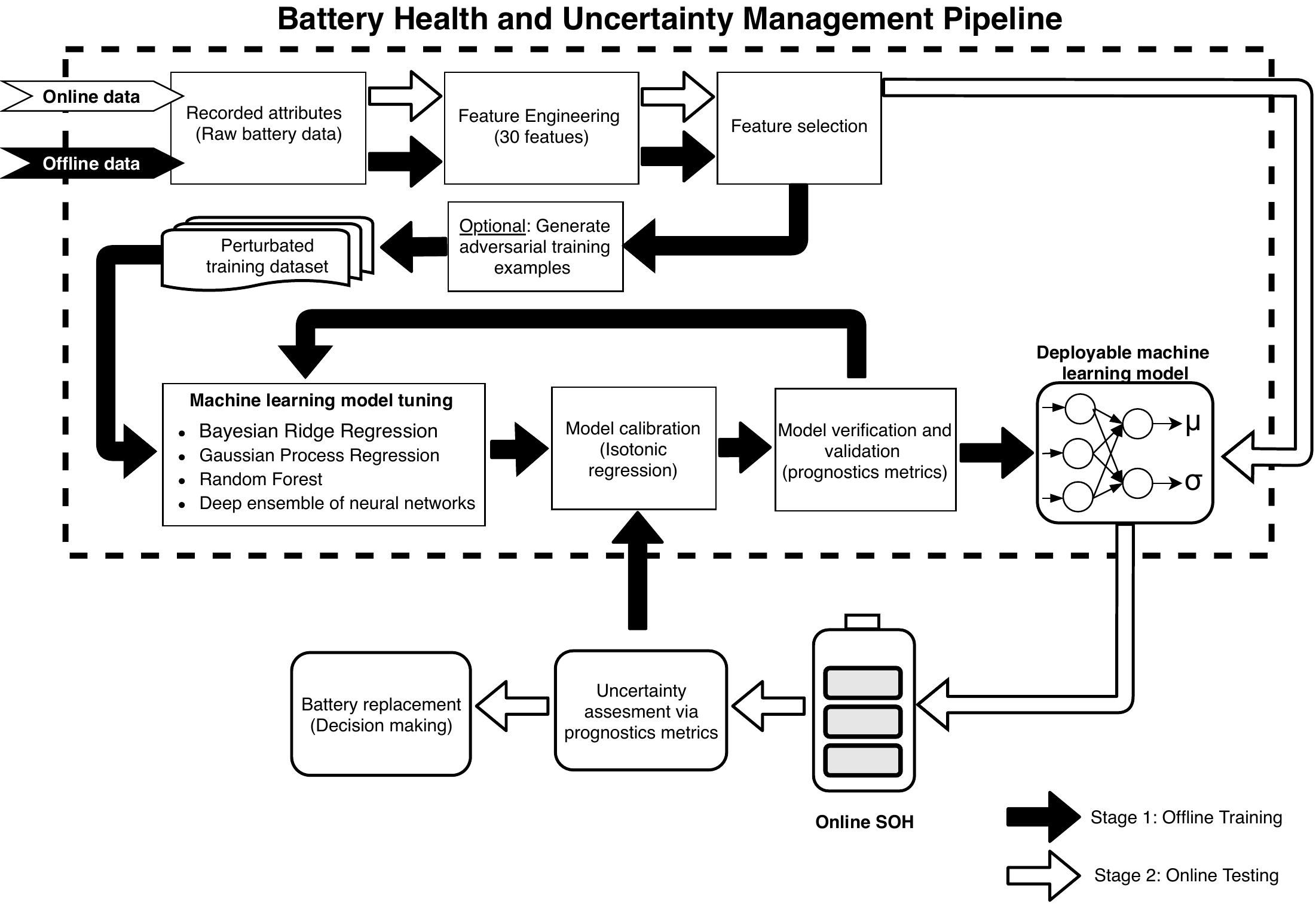}
 		\captionof{figure}{BHUMP flowchart}
 		\label{piepline_flowchart}
\end{figure}


\subsection*{Supplementary Note 1. Domain explanation of features}

Features are generated by mathematical manipulation, involving pattern recognition and information theory principles, of voltage and current charge curves. Any charging protocol finishes with both electrodes materials at their most extreme potential (and most reactive states) \cite{chem_society_journal}, namely the highest for the positive electrode and the lowest for the negative electrode. The diffusion of lithium ions inside an electrode is a complex process involving both microscopic and macroscopic processes that can potentially be partially captured by charge curves.  During charging two crucial processes occur at the anode side (graphite-based batteries considered here), namely the intercalation of lithium ions into the active material and lithium plating \cite{understanding_plating}, \cite{understanding_cathode}, \cite{factors_limit_cycle_life}. Due to intercalation kinetics at the anode, cathode deintercalates faster than the anode can intercalate, and thus during charging, the current is the main limiting factor in a graphite-based lithium-ion battery. \cite{understanding_plating}, \cite{charging_prot_on_cyclelife}, \cite{factors_limit_cycle_life} Consequently, any charging protocol suffers from such limitations. The charging protocols typically go through a constant-current (CC) mode, followed by a constant-voltage (CV) mode, see Supplementary Figure \ref{CC_CV} for a typical CC-CV charge protocol.

Zhang et al.~\cite{study_charging} investigated the effects of charging protocols in LiCoO$_2$ based batteries by creating a bespoke three-electrode cell. The authors emphasise that lithium-ion plating coexists with the intercalation process in the anode and it occurs in the late period of the CC despite the graphite not being fully lithiated. Similarly, Zhou et al. \cite{zhou2008studies_polarisation} also mention that kinetically, under high-current charging conditions, the negative electrode can be polarised to such an extent that it's potential drops below 0 V, facilitating lithium metal deposition onto the surface of the electrode particles. It is known that the duration of the CC captures the polarisation phenomenon.\cite{zhou2008studies_polarisation} Therefore, as the battery ages, the constant current charge time (CCCT) decreases. Upon the start of the CV charging, as the current decreases, the negative electrode slowly recovers to a nominal potential value. The CV mode duration is thus crucial to eliminate the polarisation effect caused during the CC mode allowing for the anode to recover and thus fully charge the battery. With aging, the constant voltage charge time (CVCT) increases as demonstrated in \cite{study_charging} and \cite{feature_Nick}.

A feasibility study of CCCT and constant voltage charge time (CVCT) as proxies for battery state of health was carried out in \cite{feature_Nick}. CVCT has already been considered as input to SOH methods in the additional studies  \cite{CVCT_Akram}, \cite{SOH_CVCT_2018}. To reduce diagnostics time, we only use sections of the charge curves as input to the algorithm. The availability of the entire charge curve in real-life applications is limited. Hence it is advantageous to design features that could be extracted from segments of such curves.  The benefits of the approach are a lower diagnostics time (as little as 15 min) and the possibility of battery SOH estimation even in partial discharge conditions.

During discharge, the process of lithium extraction/insertion happens in reverse from anode to cathode. Since discharge currents vary with usage, we only extract one feature from the discharge curve, namely the pseudo linear resistance as introduced by Saxena et al. \cite{pseudo_linear_resistance}. This is due to the instant drop in voltage associated with internal battery impedance on the application of load current. We estimate this resistance as the ratio of the observed voltage drop and the applied load current. It is understood that as the battery degrades the internal resistance of the battery increases, and hence an estimate of this internal resistance can be used as a proxy for battery SOH.\cite{pseudo_linear_resistance} We used a lagged version of this feature, i.e. pseudo linear resistance from the previous cycle to estimate the SOH at the end of a charge cycle. For a mathematical explanation of all engineered features in Supplementary Material Table \ref{attributes_extended} refer to Supplementary Note 3 Feature engineering.

\subsection*{Supplementary Note 2. Voltage threshold values}

\begin{figure*}[!ht]
\centering%
    \begin{minipage}{0.35\textwidth}
      \subcaption{}
      \includegraphics[width=\linewidth]{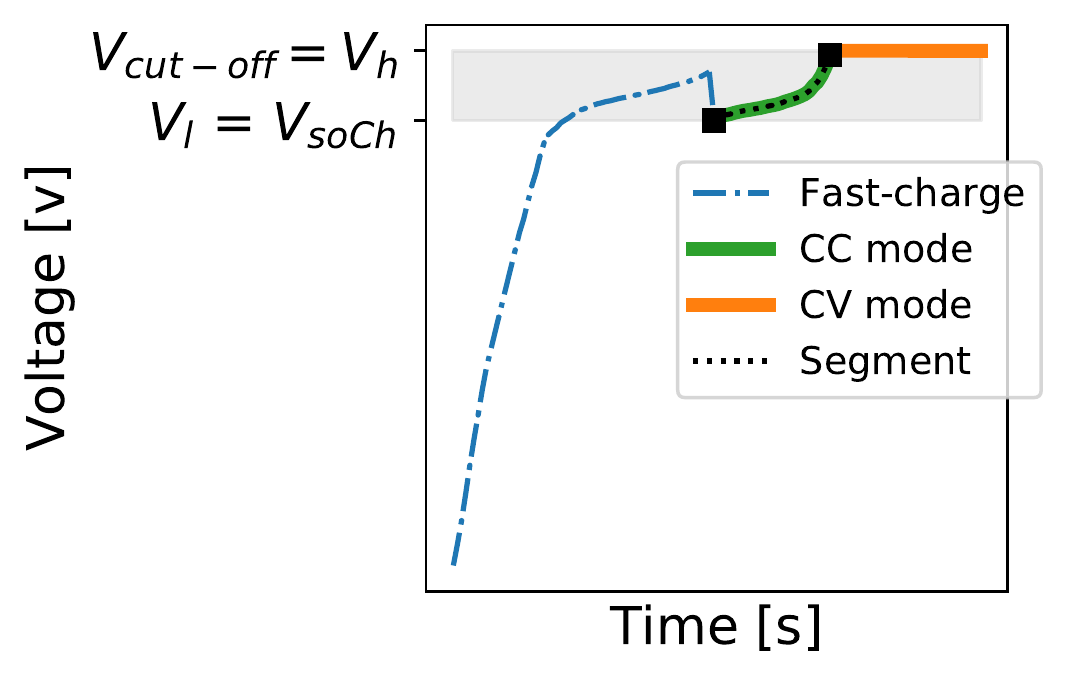}
      \label{thresholds_v_gr2}
    \end{minipage}%
    \begin{minipage}{0.35\textwidth}\centering
      \subcaption{}
      \includegraphics[width=\linewidth]{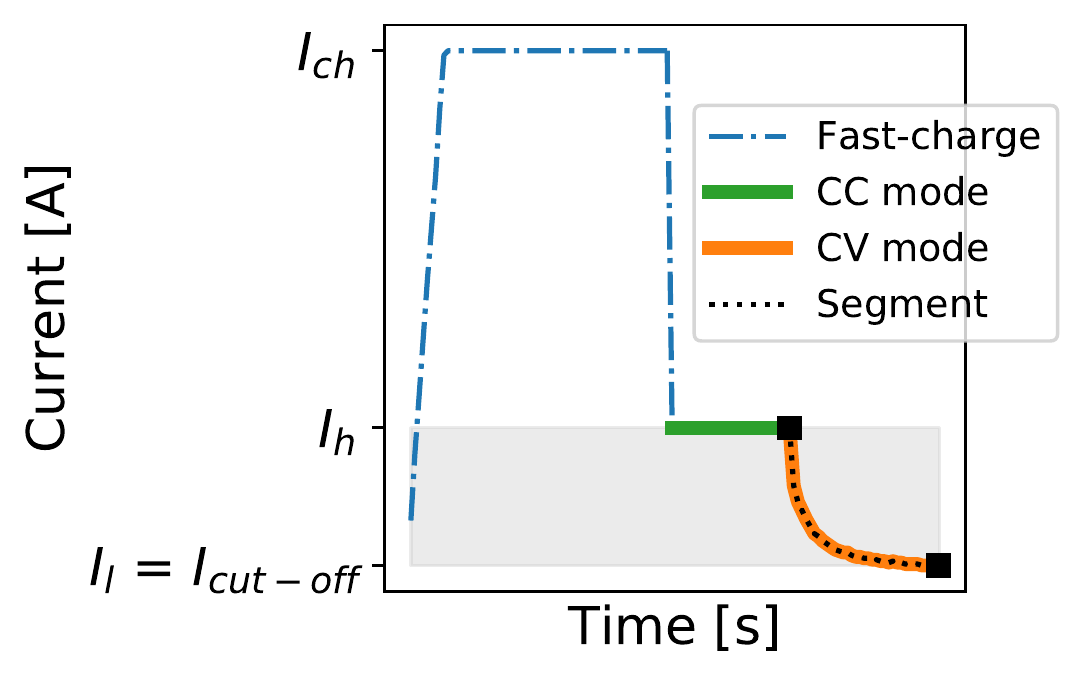}
      \label{thresholds_c_gr2}
    \end{minipage}%

  \begin{minipage}{0.25\textwidth}\centering
      \subcaption{}
      \includegraphics[width=\linewidth]{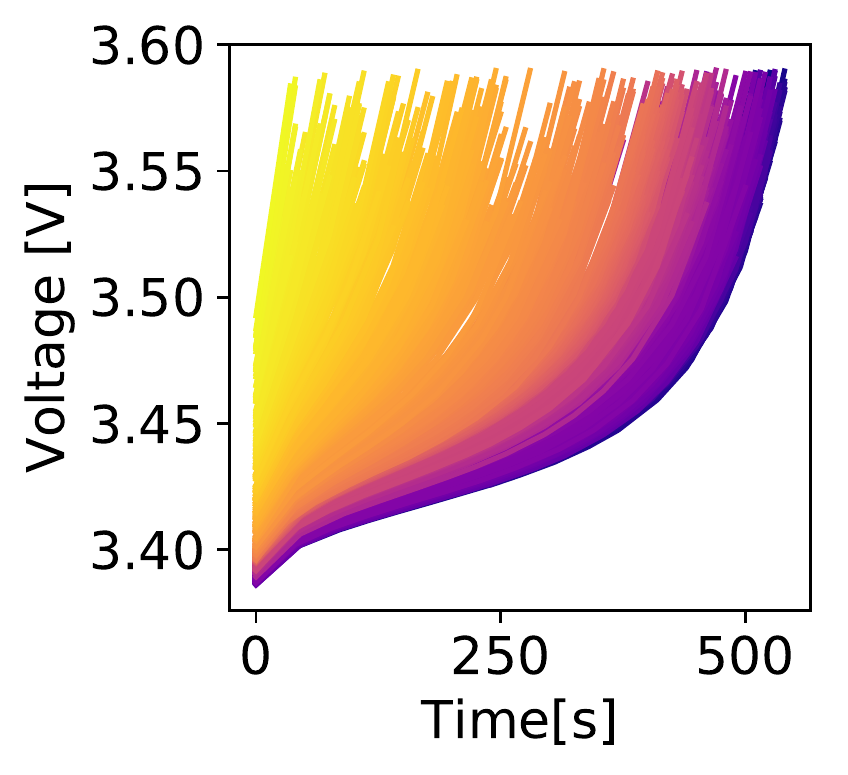}
      \label{b2}
    \end{minipage}%
  \begin{minipage}{0.25\textwidth}\centering
      \subcaption{}
      \includegraphics[width=\linewidth]{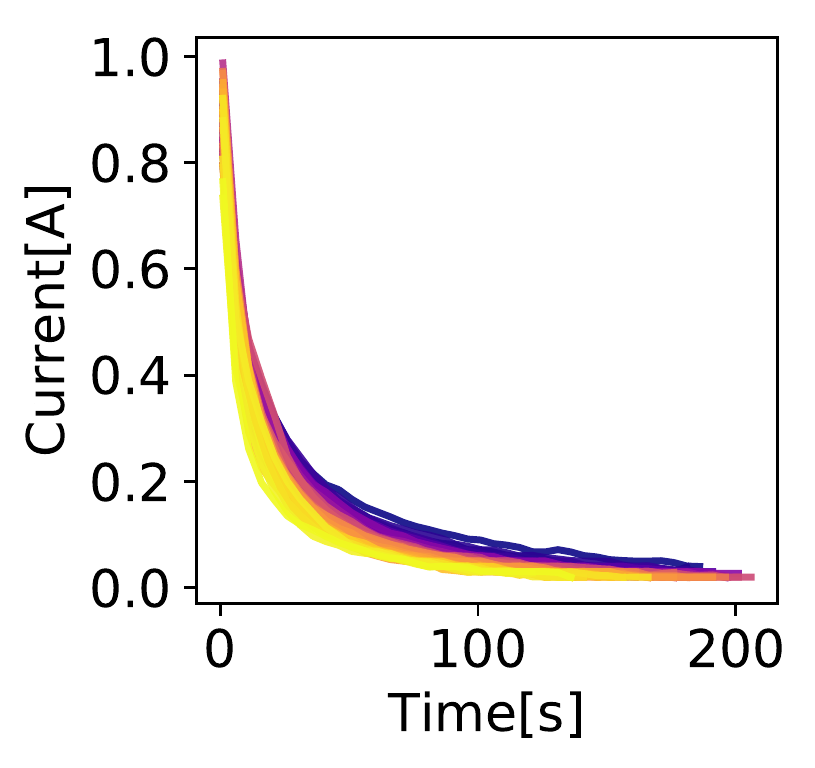}
      \label{d2}
    \end{minipage}%
  \begin{minipage}{0.15\textwidth}\centering
      \subcaption{}
      \includegraphics[width=\linewidth]{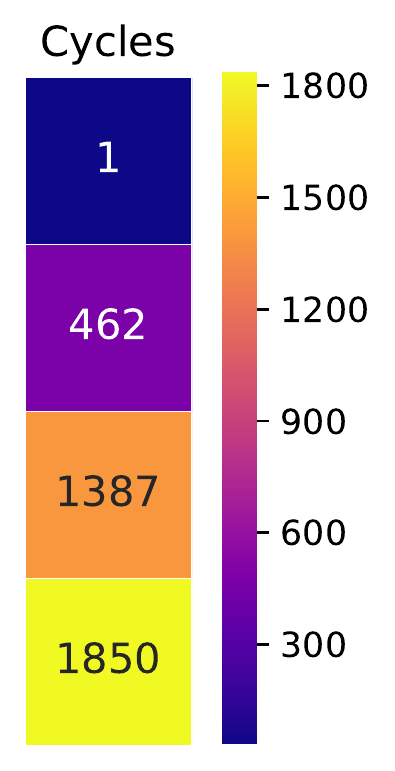}
      \label{e2}
    \end{minipage}%
    
\caption{\textbf{The 2 step fast-charge protocol and extracted ageing segment of the curves for a Li-ion pouch cell.} \textbf{\protect\subref{thresholds_v_gr2}} Voltage during charge protcol. \textbf{\protect\subref{thresholds_c_gr2}} Current during charge protocol, \textbf{\protect\subref{b2}} Extracted ageing voltage curve segments corresponding to marked grey area, \textbf{\protect\subref{d2}} Extracted ageing current curve segments corresponding to marked grey area, \textbf{\protect\subref{e2}} Heatmap of ageing with cycle number.}
   \label{features_tri}
\end{figure*}

We first define $V_h$ to be equal to charge cut-off voltage, $V_{cut-off}$, while $V_l$ is defined using the below formula:
\begin{equation}
 V_l = V_h - \Delta V
\end{equation}
where $\Delta$V is a predefined voltage range. The recorded curve between $V_l$ and $V_h$ with each charge as illustrated in Figures \ref{thresholds_v_gr2}, \ref{thresholds_v_gr2} is then normalised on the interval $[0, 1]$ by subtracting the minimum value and dividing by the resulted maximum value. Following the normalisation procedure, we proceed on mathematically deriving the features. This allows for training different batteries types and designs on the same training dataset provided they underwent the same charging protocol. To overcome issues resulting from battery terminal voltage increase after previous discharge cycle and to capture the late period of the CC charging phase (when lithium plating occurs) we make use of a $\Delta V$ equal to 0.3V. A high $V_l$ value accommodates for the increase in battery terminal voltage upon removal of load current after each discharge cycle. A behaviour commonly observed with battery ageing as referenced in Supplementary Figure~\ref{voltage_recovery}. Furthermore, a high $V_l$ threshold reduces the time necessary to record the CC charge curve while accommodating for partial discharge of the battery. Note, $\Delta V$ value and corresponding $V_l$ and $V_h$ threshold values could be adjusted based on battery type, application and user behaviour, end of life threshold, data storage capacity and processing power.

\subsection*{Supplementary Note 3. Feature engineering.}
Capacity (Q) is calculated based the charge/discharge current (I) and it is given by:
\begin{equation}
 Q = \int_{t_0}^{t_{end}} I dt
\end{equation}

Energy (E) is calculated based on capacity (Q) and voltage (V) given by:
\begin{equation}
 E = \int_{t_0}^{t_{end}} V(t) \cdot I dt
\end{equation}

\begin{figure}[h!]
    \begin{center}
        \centering
        \includegraphics[width = 0.6\textwidth]{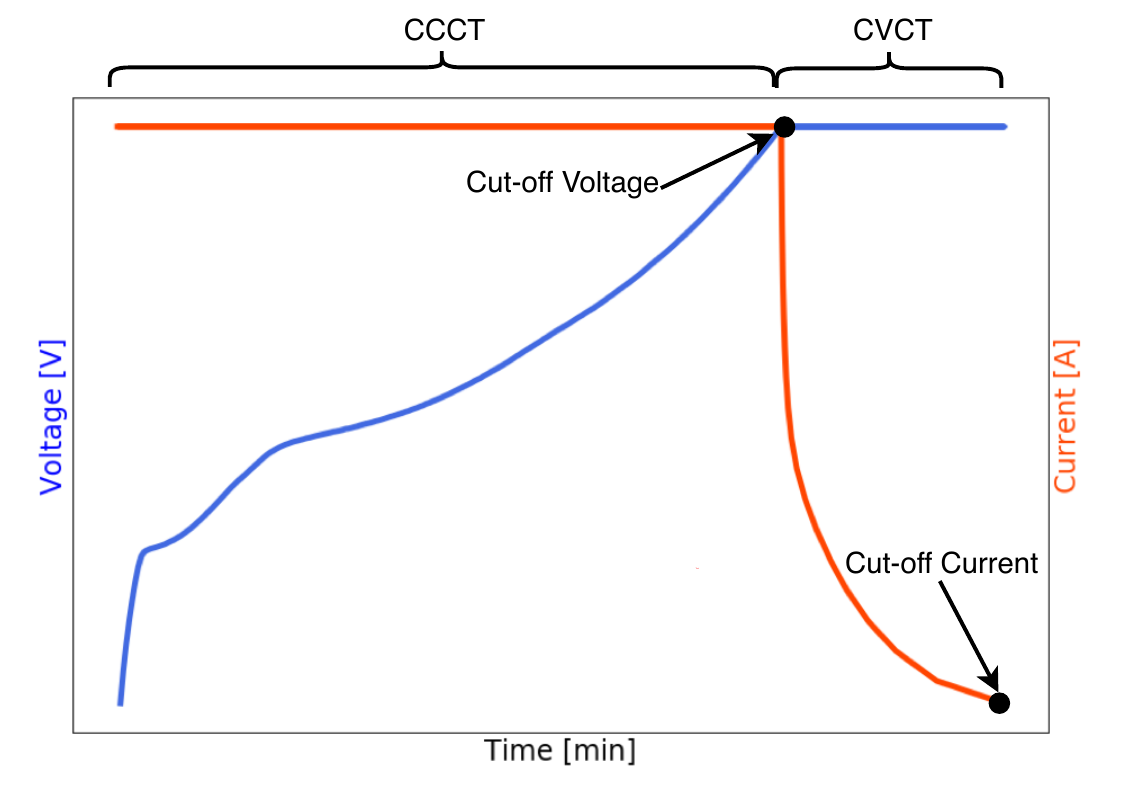}
        \caption{Typical constant current - constant voltage charge protocol. Note: CCCT=constant voltage charge time, CVCT=constant voltage charge time}
        \label{CC_CV}
    \end{center}
\end{figure}

\begin{table}[h!]
\centering
\resizebox{0.35\textwidth}{!}{%
\begin{tabular}{|c|c|}
\hline
\textbf{Attribute} & \textbf{Target variable}   \\ \hline
Cycle time  & \multirow{6}{*}{Discharge Capacity} \\ \cline{1-1}
Discharge C-rate &          \\ \cline{1-1}
Charge C-rate  &          \\ \cline{1-1}
Operational time   &          \\ \cline{1-1}
Voltage vs. Time   &          \\ \cline{1-1}
Current vs. Time   &          \\ \cline{1-1}
Charge times  &          \\ \hline
\end{tabular}%
}
\caption{Parameters recorded during cycling tests.}
\label{attributes_table}
\end{table}

\begin{table}[h!]
\centering
\resizebox{\textwidth}{!}{%
\begin{tabular}{c|l|c|}
\cline{2-3}
                 & \textbf{Feature}            & \textbf{Target variable}      \\ \hline
\multicolumn{1}{|c|}{\multirow{3}{*}{Battery specific data}}  & Nominal Capacity {[}Ah{]}         & \multirow{30}{*}{Discharge Capacity {[}Ah{]}} \\ \cline{2-2}
\multicolumn{1}{|c|}{}            & Charge Current {[}A{]}          &            \\ \cline{2-2}
\multicolumn{1}{|c|}{}            & Discharge Current {[}A{]}         &            \\ \cline{1-2}
\multicolumn{1}{|c|}{\multirow{2}{*}{Cumulative (historical) data}} & Cumulated Discharge Capacity {[}Ah{]}      &            \\ \cline{2-2}
\multicolumn{1}{|c|}{}            & Cumulated Discharge Energy {[}Wh{]}       &         \\ \cline{1-2}
\multicolumn{1}{|c|}{\multirow{2}{*}{1 Cycle Lagged Data}}   & Lagged Cycle Time {[}s{]}         &            \\ \cline{2-2}
\multicolumn{1}{|c|}{}            & Lagged Pseudo Resistance {[}$\Omega${]}      &       \\ \cline{1-2}
\multicolumn{1}{|c|}{\multirow{23}{*}{Instantaneous charge data*}} & Terminal Voltage @ Start of charge {[}V{]}    &            \\ \cline{2-2}
\multicolumn{1}{|c|}{}            & Charge time of CC segment of charge curve {[}s{]}    &  \\ \cline{2-2}
\multicolumn{1}{|c|}{}            & Charge time of CV segment of charge curve {[}s{]}    &  \\ \cline{2-2}
\multicolumn{1}{|c|}{}            & Mean current during CC segment of the curve {[}A{]}  &  \\ \cline{2-2}
\multicolumn{1}{|c|}{}            & Mean voltage during CV segment of the curve {[}A{]}  &   \\ \cline{2-2}
\multicolumn{1}{|c|}{}            & Slope of CCCV-CCCT segment of the curve              &   \\ \cline{2-2}
\multicolumn{1}{|c|}{}            & Slope of CVCC-CVCT segment of the curve              &   \\ \cline{2-2}
\multicolumn{1}{|c|}{}            & Energy during CCCV-CCCT segment of the curve {[}Wh{]} &   \\ \cline{2-2}
\multicolumn{1}{|c|}{}            & Energy during CVCC-CVCT segment of the curve {[}Wh{]} &   \\ \cline{2-2}
\multicolumn{1}{|c|}{}            & Energy ratio CCCV-CCCT / CVCC-CVCT segment of the curve &  \\ \cline{2-2}
\multicolumn{1}{|c|}{}            & Energy Difference between the curve segments (CCCV-CCCT) - (CVCC-CVCT)     &            \\ \cline{2-2}
\multicolumn{1}{|c|}{}            & Entropy of CCCV-CCCT segment of the curve eq \ref{curve_ent} &            \\ \cline{2-2}
\multicolumn{1}{|c|}{}            & Entropy of CCCV-CCCT segment of the curve eq \ref{curve_ent} &            \\ \cline{2-2}
\multicolumn{1}{|c|}{}            & Shannon entropy of CCCV segment of the curve          &  \\ \cline{2-2}
\multicolumn{1}{|c|}{}            & Shannon entropy of CVCC segment of the curve          &  \\ \cline{2-2}
\multicolumn{1}{|c|}{}            & Skewness coefficient of CCCV-CCCT segment of the curve eq \ref{skewness}      &            \\ \cline{2-2}
\multicolumn{1}{|c|}{}            & Skewness coefficient of CVCC-CVCT segment of the curve eq \ref{skewness}      &            \\ \cline{2-2}
\multicolumn{1}{|c|}{}            & Kurtosis coefficient of CCCV-CCCT segment of the curve eq \ref{kurtosis}      &            \\ \cline{2-2}
\multicolumn{1}{|c|}{}            & Kurtosis coefficient of CVCC-CVCT segment of the curve eq \ref{kurtosis}     &            \\ \cline{2-2}
\multicolumn{1}{|c|}{}            & Frechet Distance of CCCV-CCCT segment of the curve eq \ref{frecehet}       &            \\ \cline{2-2}
\multicolumn{1}{|c|}{}            & Frechet Distance of CVCC-CVCT segment of the curve eq \ref{frecehet}       &            \\ \cline{2-2}
\multicolumn{1}{|c|}{}            & Hausdorff Distance of CCCV-CCCT segment of the curve eq \ref{hausdorff}      &            \\ \cline{2-2}
\multicolumn{1}{|c|}{}            & Hausdorff Distance of CVCC-CVCT segment of the curve eq \ref{hausdorff}      &            \\ \hline
\end{tabular}%
}
\caption{Engineered features based on recorded parameters in Table \ref{attributes_table}. Note: CC = consatnt current, CV = constant voltage, CCCV = constant current charge voltage, CVCC = contant voltage charge current, CCCT = constant current charge time, CVCT = constant voltage charge time}
\label{attributes_extended}
\end{table}

From pattern recognition domain, three features are derived, signal mean, kurtosis coefficient and skewness coefficient. Skewness coefficient and kurtosis coefficient are calculated based on the following formulas:

\begin{equation}
    skewness = \frac{\sum_{i=1}^n(x(i)-\bar{x})^3}{(n-1)\sigma_x^3}
    \label{skewness}
\end{equation}

\begin{equation}
    kurtosis = \frac{\sum_{i=1}^n(x(i)-\bar{x})^4}{(n-1)\sigma_x^4}
    \label{kurtosis}
\end{equation}

where $\bar{x}$ and $\sigma_x$ represent the mean and standard deviation of feature $x$.

\begin{figure}[h!]
         \centering
         \includegraphics[width= 0.4\linewidth]{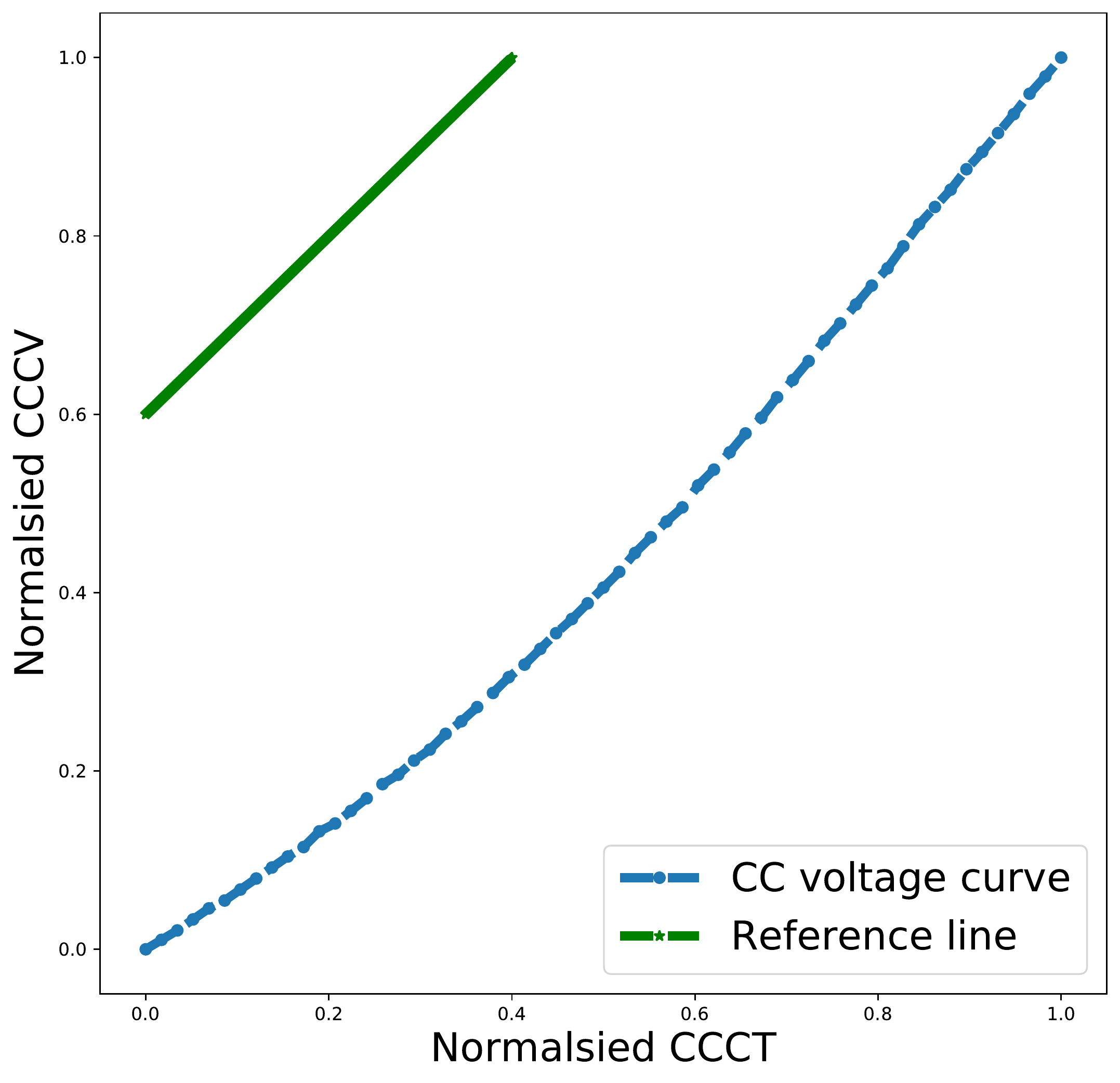}
         \captionof{figure}{Typical constant current (CC) charge curve with associated reference line of equation $y = mx+c$}
         \label{reference_current}
\end{figure}

\begin{figure}[h!]
         \centering
         \includegraphics[width= 0.4\linewidth]{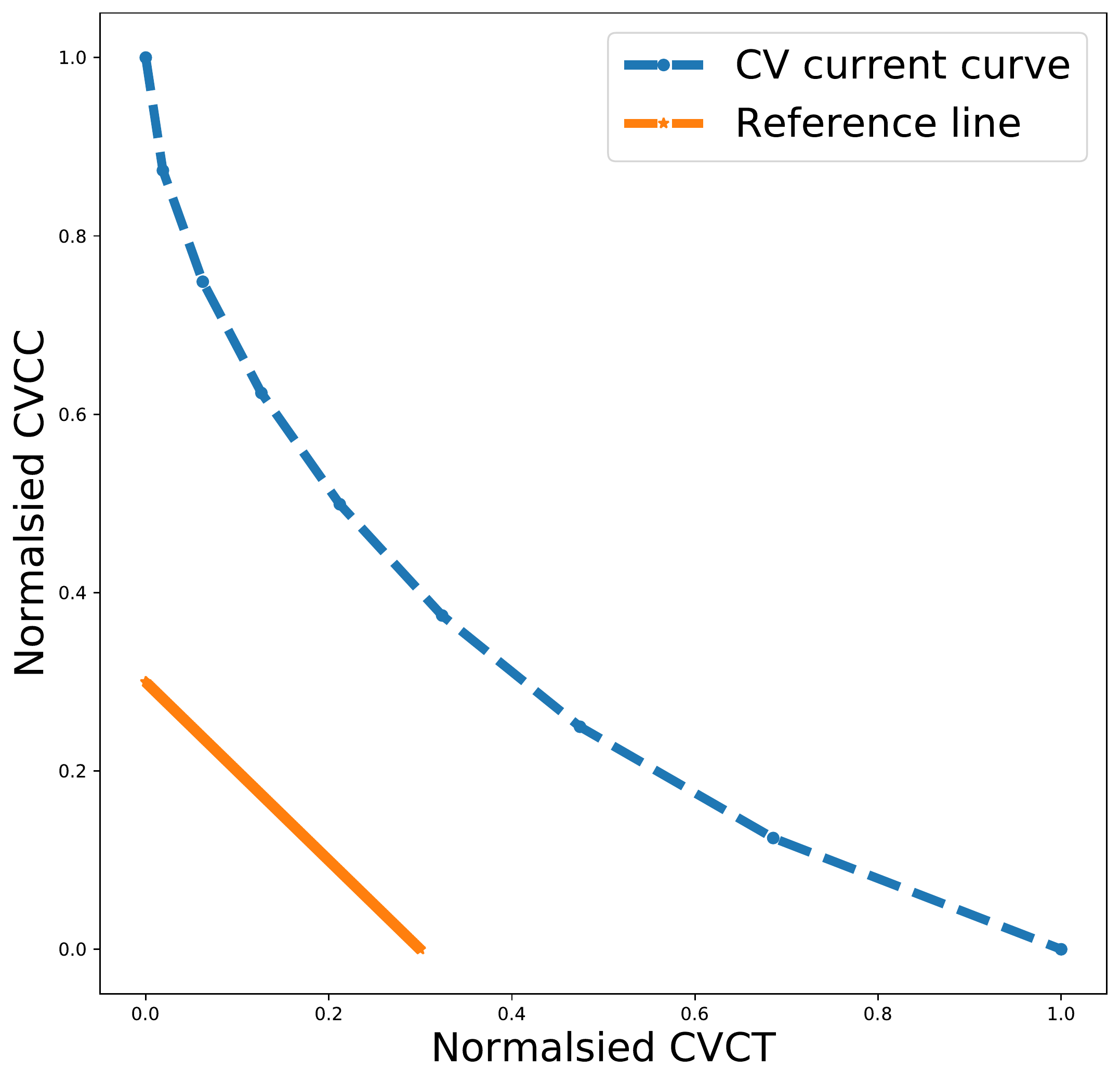}
         \captionof{figure}{Typical constant voltage (CV) charge curve with associated reference line of equation $y = mx+c$}
         \label{reference_voltage}
\end{figure}

In addition to pattern recognition based features, distance measurements from a predetermined reference curve to CVCC - CVCT curve and CCCV - CCCT have also been considered. We choose here as reference a simple line defined by the equation $y = mx+c$ where $y$ represents current or voltage depending on the curve under scrutiny, and $x$ represents time. An illustration of the two curves and their reference lines are shown in figures \ref{reference_current} and \ref{reference_voltage}. Instead of simple Euclidean distance, we employ here two different measurements, namely Directed Hausdorff (DH) and Frechet (FD) distance. Both methods are well established in various domains and thoroughly explained in \cite{HD}, \cite{curve_similarity} and \cite{frechet}. We only consider here Directed Hausdorff distance from charge curve to reference line and not vice-versa. DH distance between two point sets $A(a_1, a_2)$ and $B(b1, b2)$, where $a_1, a_2, b_1, b_2$ are 2D coordinates, is calculated as maximum distance between each point $x$ $\epsilon$ $A$ to its nearest neighbour $y$ $\epsilon$ $B$ and is given by:
\begin{equation}
    H(A, B) = max_{x \epsilon A}\{min_{y \epsilon B}\{||x, y||\}\}
    \label{hausdorff}
\end{equation}

where $||x, y||$ can be any norm, including the Euclidean distance. Note that $H(A, B) \neq H(B, A)$, in other words, DH is not symmetric.

The point set $A$ is represented by one of the two charge curves namely, CCCV-CCCT or CVCC-CVCT, whereas $B$ is represented by a line of 30-40 points as shown in \ref{reference_current} and \ref{reference_voltage}. 

Frechet distance of two curves $A$ , $B$ has been generally described as the minimal length of a leash required to connect a dog to its owner, as they walk along $A$ or $B$, respectively, without backtracking. In contrast to distance notions such as the Hausdorff distance, it takes into account the order of the points along the curve, and thus better captures the similarity as perceived by human observers.\cite{frechet} In mathematical terms, however, the Frechet distance between two curves is defined as:
\begin{equation}
    FD(A, B) = min \{ max ||A(\alpha(t)), B(\beta(t))|| \}
    \label{frecehet}
\end{equation}

where $\alpha(t)$ and $\beta(t)$, range over continuous and increasing functions with $\alpha, \beta, t$ $\epsilon [0, 1]$. Again, $||...||$ can be any norm, including Euclidian distance. A more elaborate mathematical explanation is beyond the scope of the present material, however, a thorough mathematical explanation can be found in \cite{computing_frechet}

The entropy of CVCC-CVCT and CCCV-CCCT curves is also considered as a feature. In information theory, entropy is the average rate at which information is produced by a stochastic source of data \cite{shannon1948_entropy}, whereas in statistical mechanics, entropy is an extensive property of a thermodynamic system. Thermodynamic property of curves has been thoroughly analysed in \cite{dupain_curve_entropy}, \cite{moore_curve_entropy}, \cite{curve_entropy_normal}. Authors in \cite{curve_entropy_normal} provide an algorithmic procedure to compute curve entropy, and it has been adopted here with slight modification as follows. Curve entropy (EC) is defined by:

\begin{equation}
    EC = \frac{\log_2{(\frac{2L}{D})}}{\log_2{(N-1)}}
    \label{curve_ent}
\end{equation}

where $L$ is the length of the plane curve, $D$ is the diameter of the smallest hypersphere covering the curve, and $N-1$ is the number of segments approximating the line. For a thorough mathematical explanation on how all variables have been calculated refer to reference \cite{curve_entropy_normal}.

\begin{figure}[h!]
         \centering
         \includegraphics[width= 0.7\linewidth]{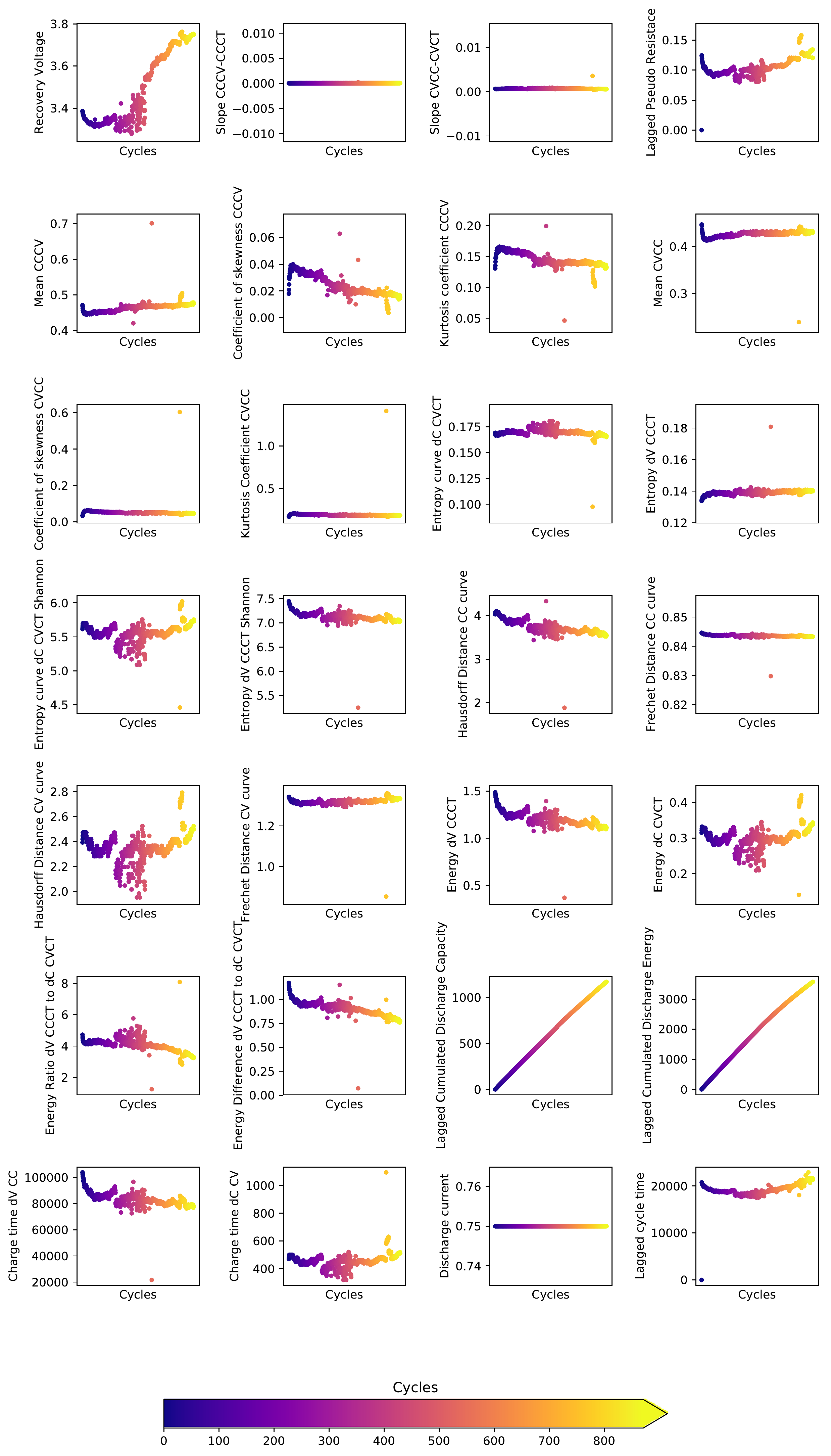}
         \captionof{figure}{Example visualisation of derived features for Group I datasets cell no. 11.}
         \label{no}
\end{figure}

\begin{figure*}[h!]
         \centering
         \includegraphics[width= 0.7\linewidth]{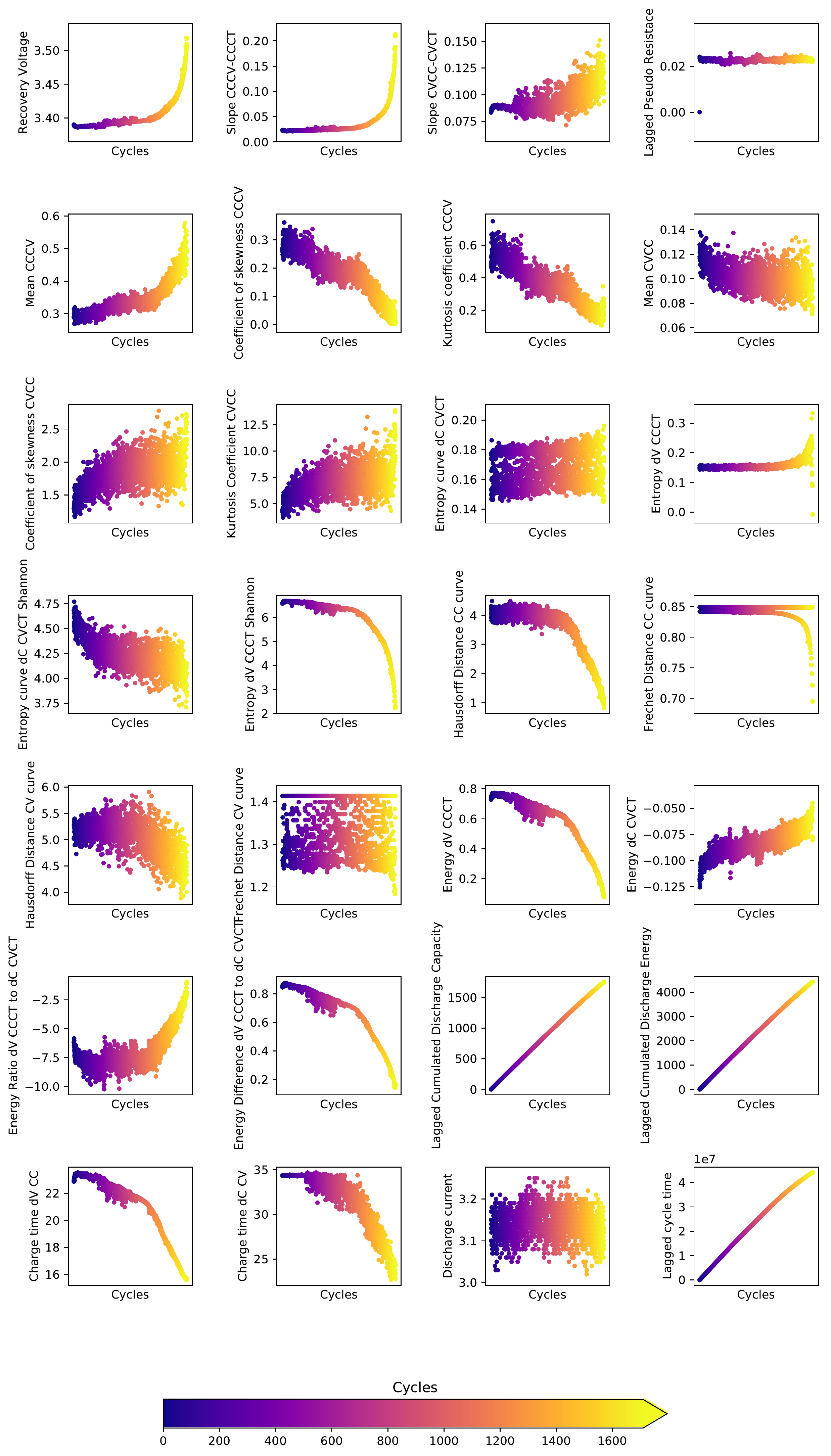}
         \captionof{figure}{Example visualisation of derived features for Group 2 datasets cell no. 1.}
         \label{no_2}
\end{figure*}

\clearpage\pagestyle{plain}


\begin{figure*}[ht!]
  \centering
  \hfil
  \begin{subfigure}[b]{.33\textwidth}
     \centering
     \caption{}
                    \includegraphics[width=\linewidth]{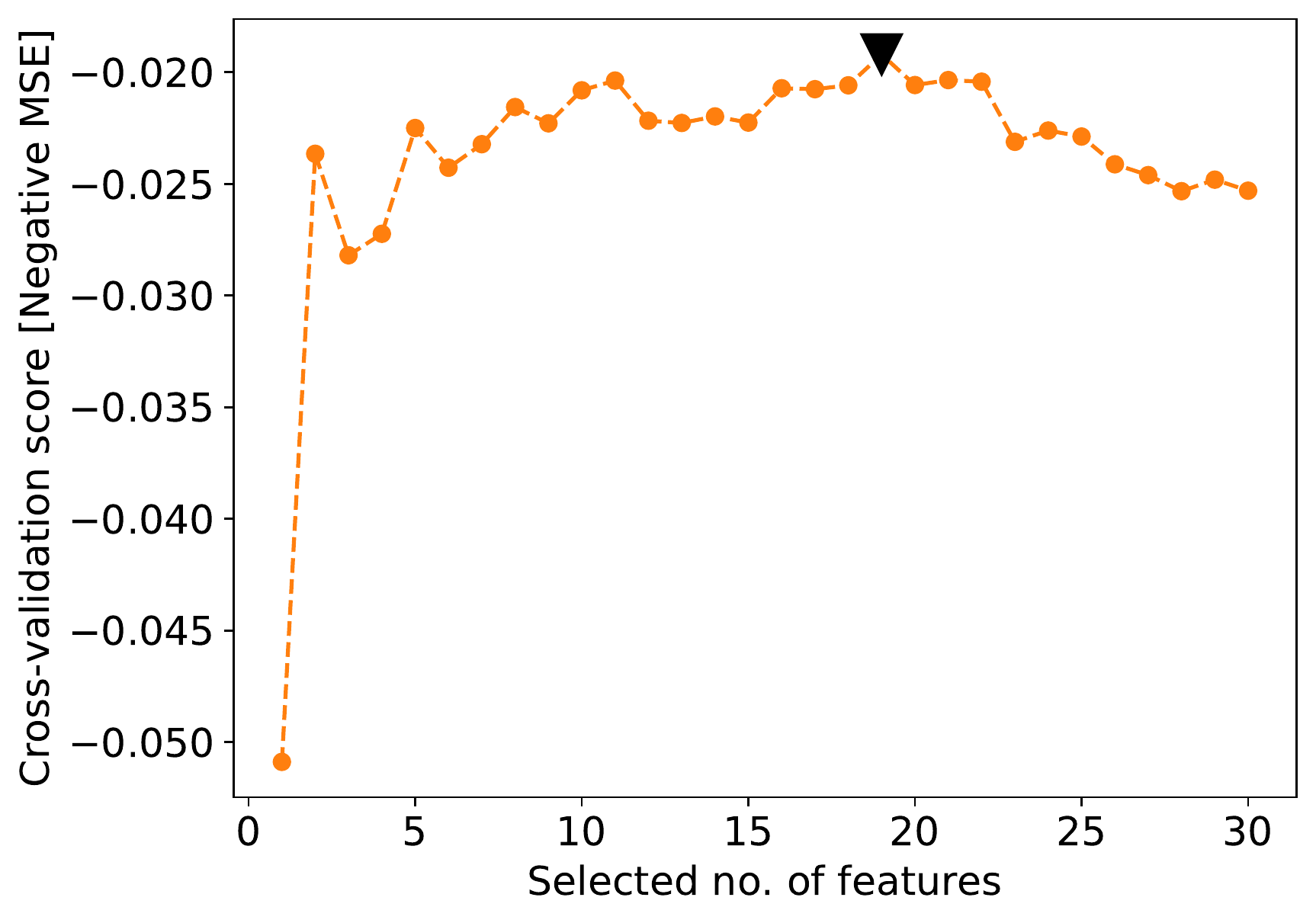}
     \label{fs_group1}
  \end{subfigure}%
  \hfill 
  \begin{subfigure}[b]{.33\textwidth}
     \centering
     \caption{}
                    \includegraphics[width=\linewidth]{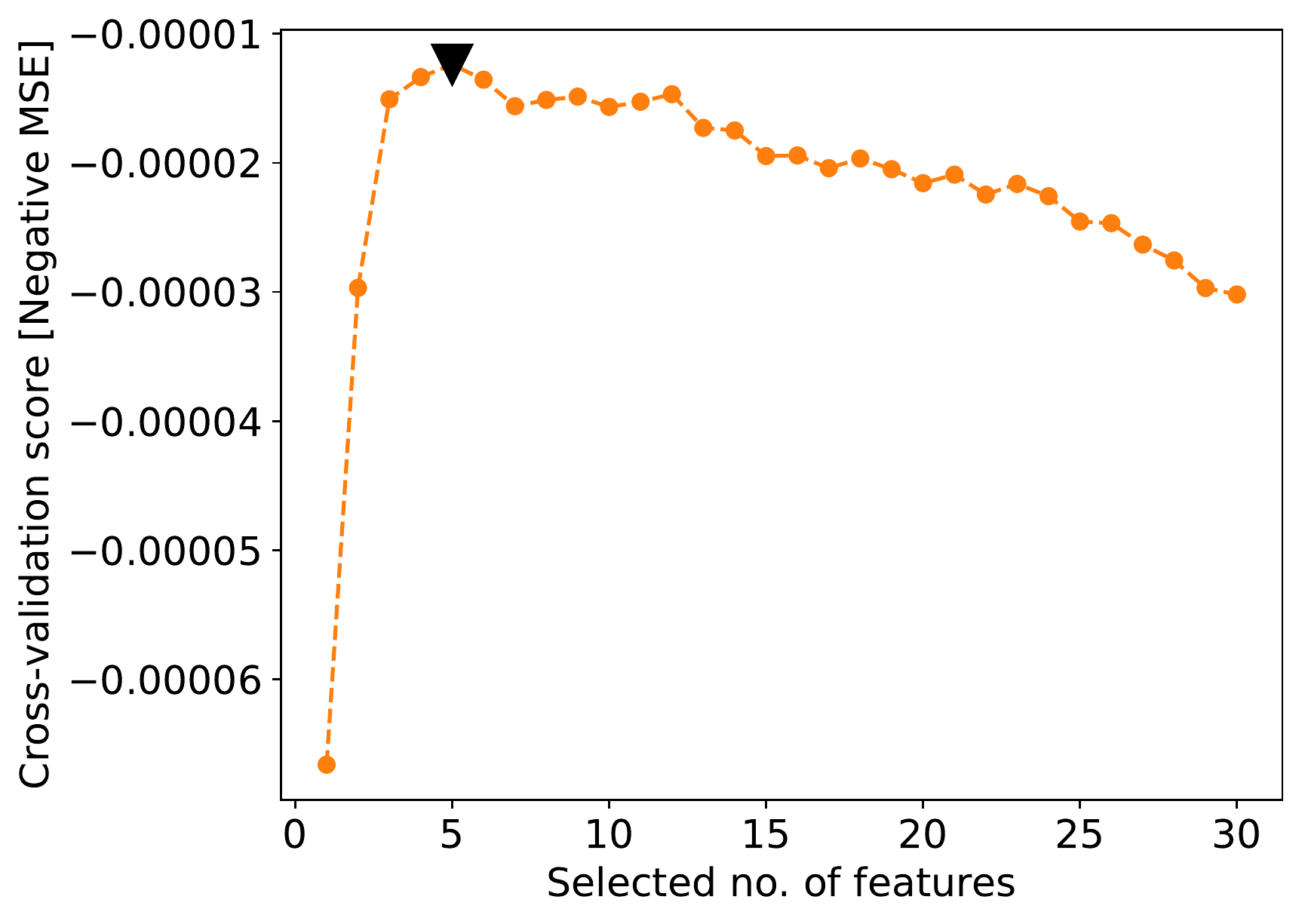}
     \label{fs_group2}
  \end{subfigure}%
  \hfill 
  \begin{subfigure}[b]{.33\textwidth}
     \centering
     \caption{}
                    \includegraphics[width=\linewidth]{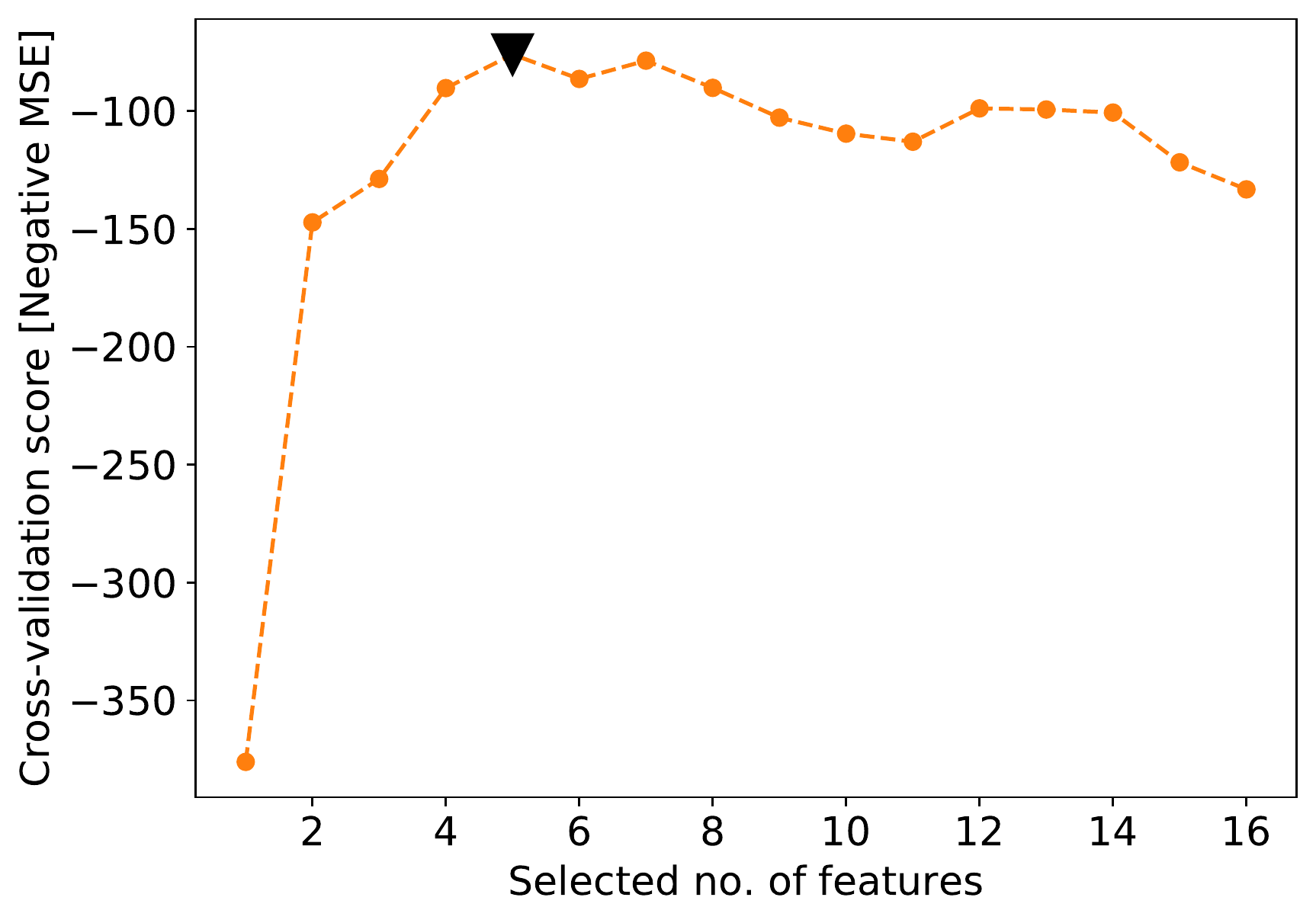}
     \label{fs_group3}
  \end{subfigure}%
  
  \caption{\textbf{Automatic feature selection with RF-RFE-CV - Note: black triangle indicates selected no. of features.} \textbf{\protect\subref{fs_group1}} 18 features selected for Group I. \textbf{\protect\subref{b}} 5 features selected for Group 2, \textbf{\protect\subref{c}} 5 features selected for Group 3.}
  \label{RF_RFE_CV}
 \end{figure*}

\begin{table}[h!]
\centering
\resizebox{\textwidth}{!}{%
\begin{tabular}{|c|c|c|}
\hline
  \textbf{Data type} & \textbf{Feature}             & \textbf{Feature no.} \\ \hline
\multirow{2}{*}{Battery specific data}  & Nominal Capacity {[}Ah{]}           & 1     \\ \cline{2-3} 
            & Charge Current {[}A{]}           & 2     \\ \hline
\multirow{2}{*}{Cumulative (historical) data} & Cumulated Discharge Capacity {[}Ah{]}        & 3     \\ \cline{2-3} 
            & Cumulated Discharge Energy {[}Wh{]}        & 4     \\ \hline
\multirow{2}{*}{1 Cycle Lagged Data}   & Lagged Cycle Time {[}s{]}           & 5     \\ \cline{2-3} 
            & Terminal Voltage @ Start of charge {[}V{]}     & 6     \\ \hline
\multirow{12}{*}{Instantaneous Charge Data} & Charge time of CC segment of charge curve {[}s{]}     & 7     \\ \cline{2-3} 
            & Charge time of CV segment of charge curve {[}s{]}              & 8     \\ \cline{2-3} 
            & Mean current during CC segment of the curve {[}A{]}         & 9     \\ \cline{2-3} 
            & Slope of CCCV-CCCT segment of the curve                     & 10     \\ \cline{2-3} 
            & Slope of CVCC-CVCT segment of the curve                     & 11     \\ \cline{2-3} 
            & Energy during CCCV-CCCT segment of the curve {[}Wh{]}       & 12     \\ \cline{2-3} 
            & Energy during CVCC-CVCT segment of the curve {[}Wh{]}       & 13     \\ \cline{2-3} 
            & \multicolumn{1}{l|}{Energy ratio CCCV-CCCT / CVCC-CVCT segment of the curve}   & 14     \\ \cline{2-3} 
            & \multicolumn{1}{l|}{Energy Difference between curve segments (CCCV-CCCT) - (CVCC-CVCT)} & 15     \\ \cline{2-3} 
            & Entropy of CCCV-CCCT segment of the curve based on \textbackslash{}ref\{\}  & 16     \\ \cline{2-3} 
            & Shannon entropy of CCCV  segment of the curve               & 17     \\ \cline{2-3} 
            & Frechet Distance of CCCV-CCCT segment of the curve          & 18     \\ \hline
\end{tabular}%
}
\caption{Selected features using RF-RFE-CV for Group I. Note: CC = consatnt current, CV = constant voltage, CCCV = constant current charge voltage, CVCC = contant voltage charge current, CCCT = constant current charge time, CVCT = constant voltage charge time}
\label{fs_group1_table}
\end{table}

\begin{table}[ht!]
\centering
\resizebox{.9\textwidth}{!}{%
\begin{tabular}{|c|c|c|}
\hline
\textbf{Data type} & \textbf{Feature} & \textbf{Feature no.} \\ \hline
\multirow{5}{*}{Instantaneous charge data} & Energy during CCCV-CCCT segment of the curve {[}Wh{]} & 1 \\ \cline{2-3}
 & \multicolumn{1}{l|}{Energy Difference between curve segments (CCCV-CCCT) - (CVCC-CVCT)} & 2 \\ \cline{2-3} 
 & Hausdorff Distance of CCCV-CCCT segment of the curve & 4 \\ \cline{2-3} 
 & Shannon entropy of CCCV segment of the curve & 3 \\ \cline{2-3} 
 & Frechet Distance of CCCV-CCCT segment of the curve & 5 \\ \hline
\end{tabular}%
}
\caption{Selected features using RF-RFE-CV for Group 2. Note: CC = consatnt current, CV = constant voltage, CCCV = constant current charge voltage, CVCC = contant voltage charge current, CCCT = constant current charge time, CVCT = constant voltage charge time}
\label{fs_group2_table}
\end{table}

\begin{table}[ht!]
\centering
\resizebox{0.9\textwidth}{!}{%
\begin{tabular}{|c|c|c|}
\hline
\textbf{Data type} & \textbf{Feature} & \textbf{Feature no.} \\ \hline
\multirow{2}{*}{Cumulative (historical) data} & Cumulated Discharge Capacity {[}Ah{]} & 1 \\ \cline{2-3} 
 & Cumulated Discharge Energy {[}Wh{]} & 2 \\ \hline
1 Cycle Lagged Data & Lagged Cycle Time {[}s{]} & 3 \\ \hline
\multirow{2}{*}{Instantaneous Charge Data} & Capacity during CCCV-CCCT segment of the curve {[}Ah{]} & 4 \\ \cline{2-3} 
 & Energy during CCCV-CCCT segment of the curve {[}Wh{]} & 5 \\ \hline
\end{tabular}%
}
\caption{Selected features using RF-RFE-CV for Group 3. Note: CC = consatnt current, CV = constant voltage, CCCV = constant current charge voltage, CVCC = contant voltage charge current, CCCT = constant current charge time, CVCT = constant voltage charge time}
\label{fs_group3_table}
\end{table}

\begin{figure*}[th!]
         \centering
         \includegraphics[width= 0.7\linewidth]{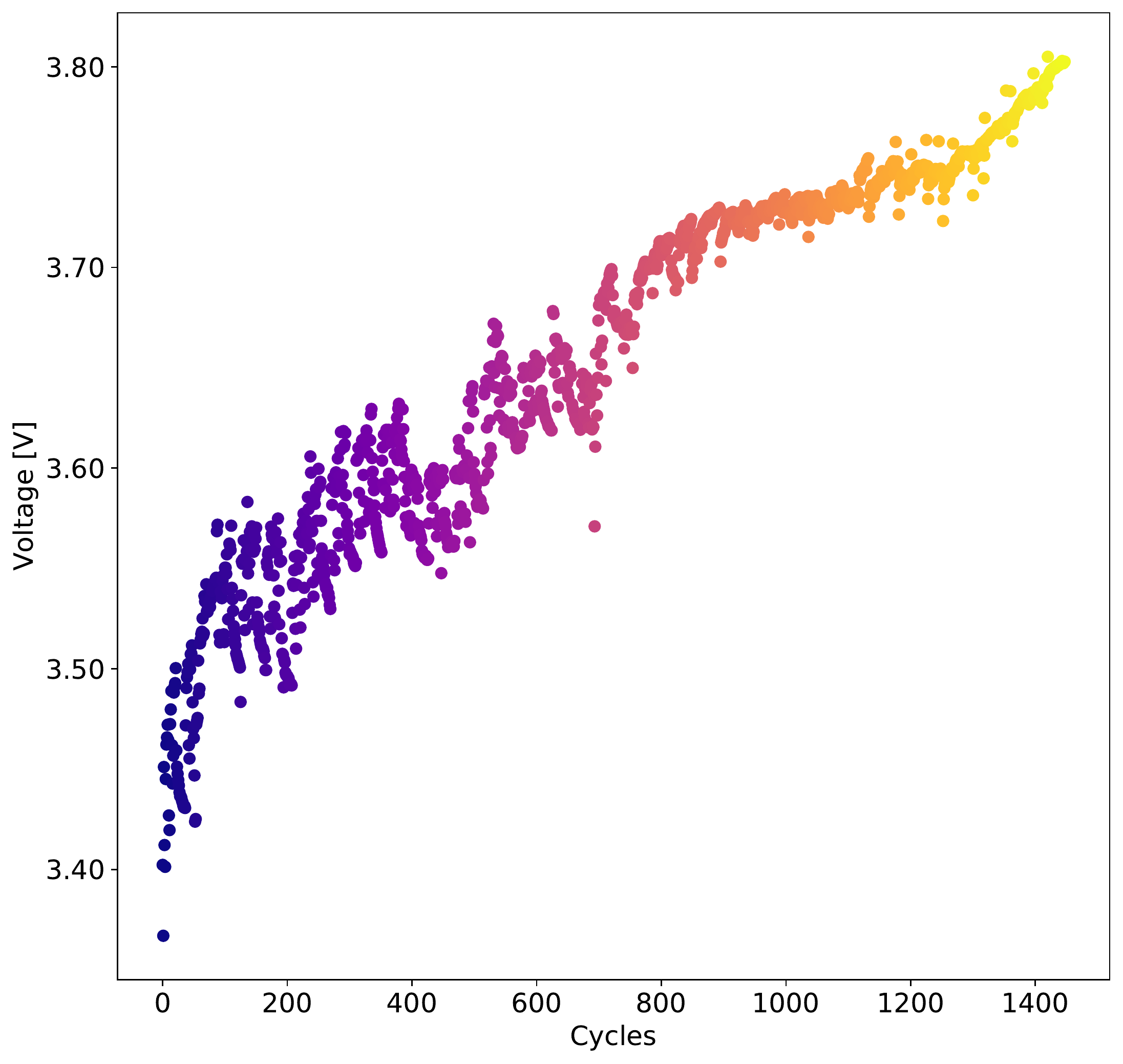}
         \captionof{figure}{ Increase in start of charge voltage between test cycles for a LiCoO$_2$ prismatic battery. The cell underwent a full depth of discharge at a current value of 1 C-rate with constant current - constant voltage charging.}
         \label{voltage_recovery}
\end{figure*}

\subsection*{Supplementary Note 4. Data overview}

Irrespective of dataset, input data consistency is ensured by removing outliers in the training data, possibly introduced due to inherent cell variability and measurement errors. The data preprocessing step involves filtering of the raw data based on erroneous capacity measurements by utilizing Random Sample Consensus (RANSAC) algorithm \cite{Fischler_RANSAC}. Training data that contains a significant percentage of gross errors in capacity from one cycle to another is removed as illustrated in the examples of Supplementary Figure \ref{outlier_removal}. Note, test data has not been processed for outliers to simulate a realistic deployment scenario.

\label{sec:dataset}

\begin{figure*}[h!]
  \centering
  \hfil
  \begin{subfigure}[b]{.25\textwidth}
     \centering
     \caption{}
                    \includegraphics[width=\linewidth]{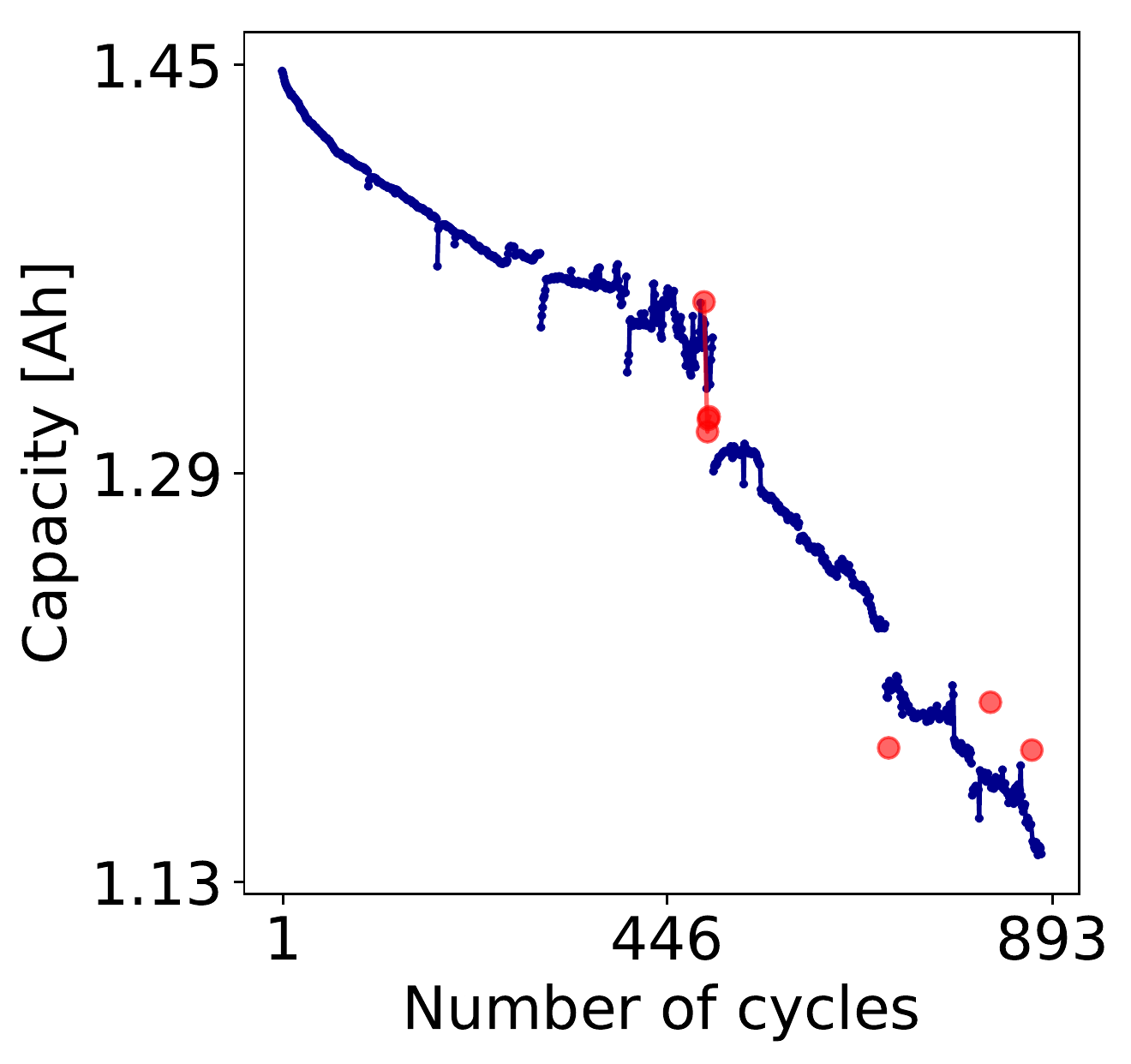}
     \label{a}
  \end{subfigure}%
  \hfill 
  \begin{subfigure}[b]{.235\textwidth}
     \centering
     \caption{}
                    \includegraphics[width=\linewidth]{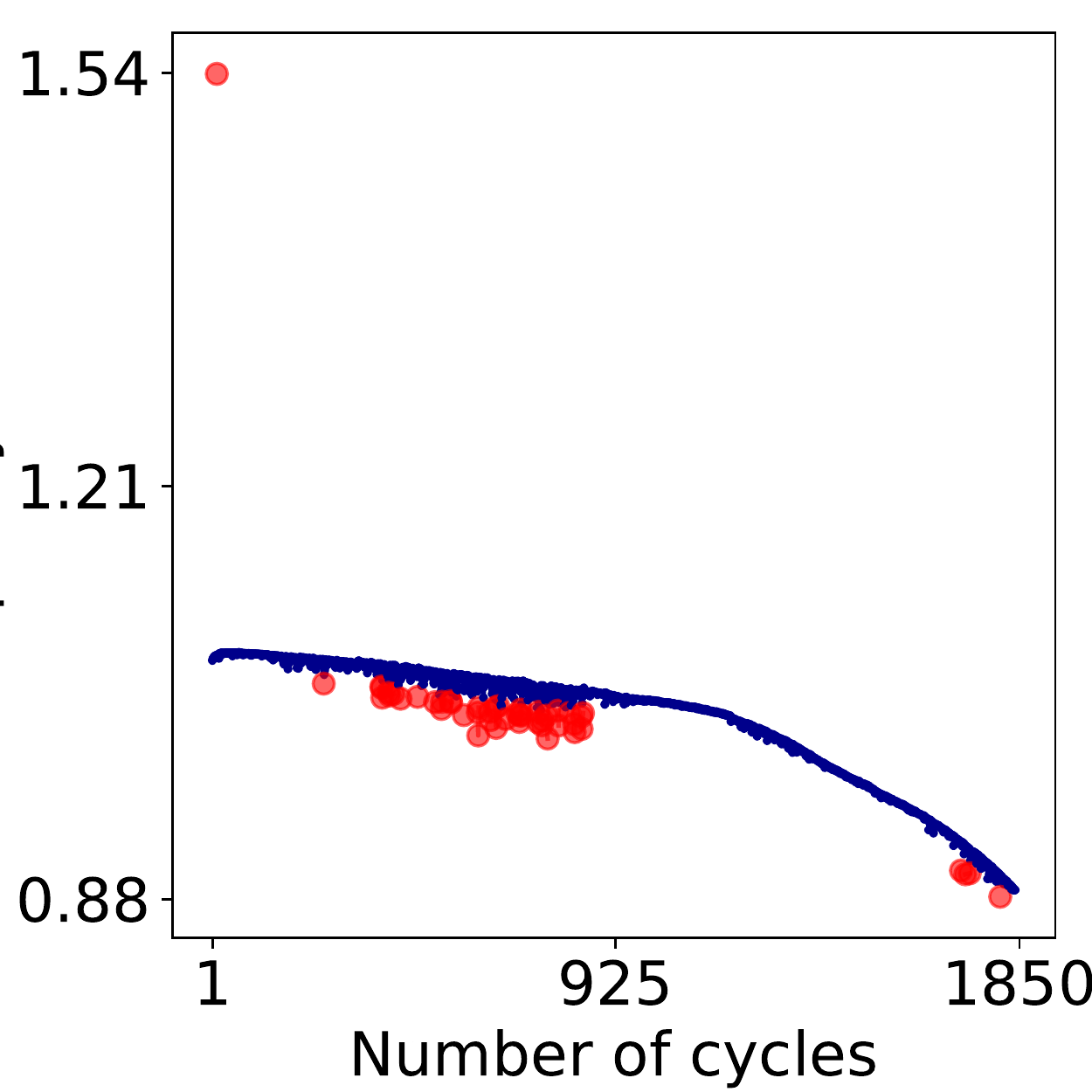}
     \label{b}
  \end{subfigure}%
  \hfill 
  \begin{subfigure}[b]{.25\textwidth}
     \centering
     \caption{}
                    \includegraphics[width=\linewidth]{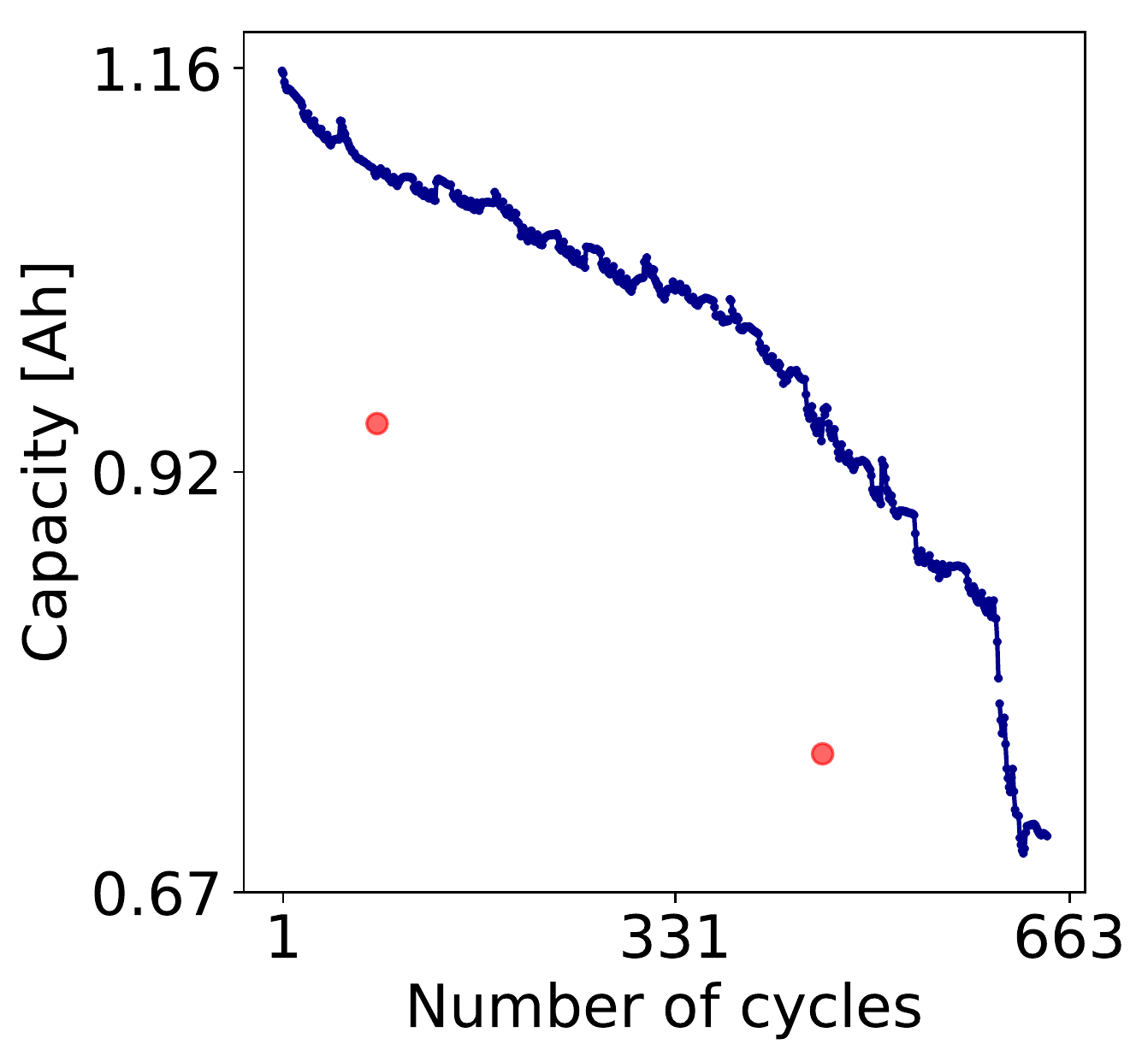}
     \label{c}
  \end{subfigure}%
  \hfill 
  \begin{subfigure}[b]{.25\textwidth}
     \centering
     \caption{}
                    \includegraphics[width=\linewidth]{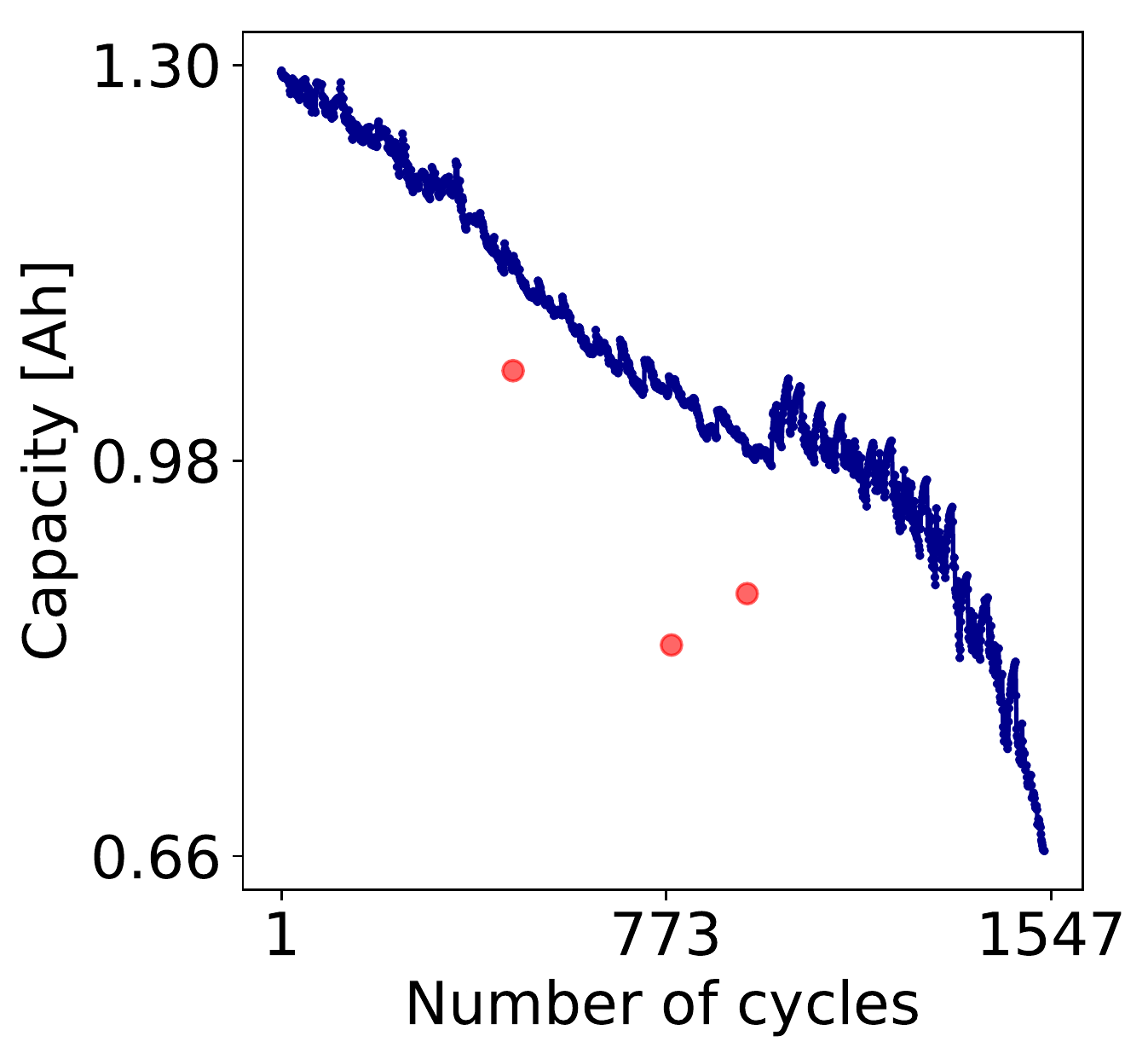}
     \label{d}
  \end{subfigure}

  \caption{\textbf{Training data outlier removal with RANSAC (red denotes outliers, blue denotes inlier).} \textbf{\protect\subref{a}} Cylindrical A123 LFP/graphite training cell 11 Group II. \textbf{\protect\subref{b}} Pouch LCO cell 2 Group I, \textbf{\protect\subref{c}} Prismatic CS2 LCO training cell 34 Group I, \textbf{\protect\subref{d}} Prismatic CX2 LCO training cell 34 Group I.}
  \label{outlier_removal}
 \end{figure*}

 \subsubsection*{CALCE dataset}
    
 Data sourced from CALCE battery group consists of three batteries. For ease of reference, we preserve the original dataset names as per their website \url{https://web.calce.umd.edu/batteries/data.htm}. All cells in the dataset underwent the same charging profile, the standard CC-CV. The CC phase of charging profile includes a 0.5 C-rate charging current until the voltage reached the cut-off threshold value of 4.2V. The CV top-up phase sustained the previously reached 4.2V until the current dropped to a value of 0.05 C-rate, at which point the charging is complete. Except for batteries in CALCE PL dataset, which were discharged at 1 C-rate, the other two datasets have been discharged at both 0.5 C-rate and 1 C-rate until the battery voltage reached the pre-defined discharge cut-off voltage of 2.7V. A schematic of the charge profile together with a detailed summary of discharge conditions for each battery can be found in Figure \ref{CC_CV} and in Supplementary Table \ref{calce_data}, respectively.

\subsubsection*{NASA dataset}
NASA data can be retrieved from the public NASA Ames Prognostics Centre of Excellence website \url{https://ti.arc.nasa.gov/tech/dash/groups/pcoe/prognostic-data-repository/} and includes two datasets. The first repository, the battery dataset denoted here by NASA5, includes a mixture of constant discharge current and squared wave-based discharge current experiments at different temperatures. The second repository, the randomised battery usage dataset, denoted here by NASA11, includes batteries that are continuously cycled with randomly generated current profiles. The randomised nature of the load profiles is an ideal representation of practical battery usage. Both NASA5 and NASA11 dataset follow the traditional CC-CV charge protocol. CC charging mode was carried at 1.5A until the battery voltage reached 4.2V and then continued in a CV fashion until the charge current dropped to 0.02A, at which point the battery was deemed fully charged. In terms of discharge, NASA5 discharge was carried out at a constant current level of 2A or square wave loading profile of 4A until the battery voltage fell to 2.7V, 2.5V or 2.2V. Whereas, NASA11 undergone a randomised discharge profile of varying duration ranging from 5 minutes to 3 hours as well as varying discharge current values ranging from 0A to 5A. All cells underwent a periodic characterisation test whereby a 2A CC and 0.02A CV current cut-off charge protocol and a 2A constant current discharge was applied. The characterisation test data was used in BHUMP to evaluate battery health as a function of capacity, as opposed to cyclic data. Details of charging and discharging profiles per battery are found in Supplementary Table \ref{nasa_data}. 
 
 \subsubsection*{TRI dataset}
The work supported by Toyota Research Institute in partnership with MIT and Stanford generated a lifecycle battery dataset consisting of 124 cells, available at \url{https://data.matr.io/1/projects/5c48dd2bc625d700019f3204}. The dataset was used in \cite{severson_satandford} where more details on battery type, manufacturer and testing equipment can be found. All cells in the dataset were cycled with a total of 72 different fast charging-polices but identically discharged with a current 4 C-rate between 3.6V and 2.0V. The charging protocol included a two-step fast charge between 0\% to 80\% SOC. The fast charge section is followed by a CC protocol, i.e. a uniform charge current value of 1 C-rate to 3.6V until the voltage reaches the cut-off value of 3.6V, immediately followed by a CV top-up phase until current dropped to 0.02 C-rate. The raw data from each cycle is used as input to BHUMP pipeline. Details regarding the charge profile as well as the cycling regimes for each battery can be found in \cite{severson_satandford}, whilst Figure \ref{features_tri} illustrates the charging protocol and Supplementary Table \ref{data_tri_split} indicates which cells have been used for training and testing of the algorithms.

\subsubsection*{Oxford dataset}
The Oxford Battery Degradation Dataset and can be accessed at \url{https://ora.ox.ac.uk/objects/uuid:03ba4b01-cfed-46d3-9b1a-7d4a7bdf6fac}. A comprehensive explanation of the testing method, equipment and battery specific characteristic is found in \cite{oxford_birkl_thesis}. The data consists of ageing experiments by repeatedly cycling the cells via a CC charge profile coupled with the ARTEMIS urban drive cycle discharge profile. The CC charge protocol uses a 2 C-rate current to a voltage of 4.2V. The discharge profile voltage range is 4.2V to 2.7V. After every 100 cycles of repeated charge-discharge using the protocol mentioned above, a characterisation test (incorporating a full constant current charge-discharge at C/18.5 (40 mA), repeated every 100 drive cycles.) is carried out. The characterisation test data is used in this work for battery health degradation estimation purposes. Supplementary Table \ref{data_oxford_split} indicates which cells have been used for training/testing the algorithms.

\subsection*{Supplementary Note 5. Data partitioning}

\subsubsection*{Group I}

Out of the 47 cells in Group I, we use 23 cells for training (out of this 10 are used for feature selection), 5 cells for calibration and remaining 19 for evaluating the algorithm performance (the cells used during training-testing can be found in Supplementary Tables \ref{calce_data}, \ref{nasa_data}, \ref{nasa_data_2}). Note that the calibration dataset is neither used in training nor testing to prevent overfitting.

\subsubsection*{Group II}

Group II dataset is randomly split into 63 cells for training (out of which 37 cells are used for feature selection), 10 for calibration and the remainder 51 cells for testing, refer to Supplementary Table \ref{data_tri_split} for cell partition in each dataset.

\subsubsection*{Group III}

 Group III dataset is split into 3 cells for training (cells 1 to 3), one cell for calibration (cell no. 4), and the remainder of 4 cells for testing (see Supplementary Table \ref{data_oxford_split} for details).

\begin{table}[h!]
\centering
\resizebox{.55\textwidth}{!}{%
\begin{tabular}{|l|l|l|}
\hline
\multicolumn{1}{|c|}{\textbf{Cell name}} & \multicolumn{1}{c|}{\textbf{Discharge condition}} & \multicolumn{1}{c|}{\textbf{Dataset}} \\ \hline
CS2 - 33 & 0.5 C-rate & Test \\ \hline
CS2 - 34 & 0.5 C-rate & Train \\ \hline
CS2 - 35 & 1 C-rate & Train \& Feature Selection \\ \hline
CS2 - 36 & 1 C-rate & Train \& Feature Selection \\ \hline
CS2 - 37 & 1 C-rate & Calibration \\ \hline
CS2 - 38 & 1 C-rate & Test \\ \hline
CX2 - 33 & 0.5 C-rate & Test \\ \hline
CX2 - 34 & 0.5 C-rate & Train \\ \hline
CX2 - 35 & 0.5 C-rate & Train \& Feature Selection \\ \hline
CX2 - 36 & 0.5 C-rate & Calibration \\ \hline
CX2 - 37 & 0.5 C-rate & Train \& Feature Selection \\ \hline
CX2 - 38 & 0.5 C-rate & Test \\ \hline
PL - 11 & 0.5 C-rate & Train \\ \hline
PL - 13 & 0.5 C-rate & Test \\ \hline
\end{tabular}%
}
\caption{Group I: CALCE battery data discharge conditions and train, calibration and test split. For complete details on test conditions access \url{https://web.calce.umd.edu/batteries/data.htm}. }
\label{calce_data}
\end{table}

\begin{table}[h!]
\centering
\resizebox{.55\textwidth}{!}{%
\begin{tabular}{|l|l|l|}
\hline
\multicolumn{1}{|c|}{\textbf{Cell name}} & \multicolumn{1}{c|}{\textbf{Discharge condition}} & \multicolumn{1}{c|}{\textbf{Dataset}} \\ \hline
B0005 & 2A & Train \& Feature Selection \\ \hline
B0006 & 2A & Test \\ \hline
B0007 & 2A & Train \\ \hline
B0018 & 2A & Test \\ \hline
B0025 & Square wave @ 4A & Test \\ \hline
B0026 & Square wave @ 4A & Train \& Feature Selection \\ \hline
B0027 & Square wave @ 4A & Train \\ \hline
B0028 & Square wave @ 4A & Calibration \\ \hline
\end{tabular}%
}
\caption{Group I: NASA 5 battery data discharge conditions and train, calibration and test split. For complete details on test conditions access \url{https://ti.arc.nasa.gov/tech/dash/groups/pcoe/prognostic-data-repository/}.}
\label{nasa_data}
\end{table}

\begin{table}[h!]
\centering
\resizebox{.55\textwidth}{!}{%
\begin{tabular}{|l|l|l|}
\hline
\multicolumn{1}{|c|}{\textbf{Cell name}} & \multicolumn{1}{c|}{\textbf{Discharge condition}} & \multicolumn{1}{c|}{\textbf{Dataset}} \\ \hline
RW1 & Random Sequence & Train \& Feature Selection \\ \hline
RW2 & Random Sequence & Train \\ \hline
RW3 & Random Sequence & Train \\ \hline
RW4 & Random Sequence & Train \\ \hline
RW5 & Random Sequence & Test \\ \hline
RW6 & Random Sequence & Test \\ \hline
RW7 & Random Sequence & Test \\ \hline
RW8 & Random Sequence & Test \\ \hline
RW9 & Random Sequence & Train \& Feature Selection \\ \hline
RW10 & Random Sequence & Train \\ \hline
RW11 & Random Sequence & Calibration \\ \hline
RW12 & Random Sequence & Test \\ \hline
RW13 & Random Sequence & Train \\ \hline
RW14 & Random Sequence & Train \\ \hline
RW15 & Random Sequence & Test \\ \hline
RW16 & Random Sequence & Test \\ \hline
RW20 & Random Sequence & Train \& Feature Selection \\ \hline
RW21 & Random Sequence & Train \& Feature Selection \\ \hline
RW22 & Random Sequence & Train \\ \hline
RW23 & Random Sequence & Test \\ \hline
RW24 & Random Sequence & Test \\ \hline
RW25 & Random Sequence & Train \& Feature Selection \\ \hline
RW26 & Random Sequence & Train \\ \hline
RW27 & Random Sequence & Test \\ \hline
RW28 & Random Sequence & Calibration \\ \hline

\end{tabular}%
}
\caption{Group I: NASA 11 battery data discharge conditions and train, calibration and test split. Note: batteries are discharged to 3.2V using a randomized sequence of discharging loads between 0.5A and 4A. For complete details on test conditions access \url{https://ti.arc.nasa.gov/tech/dash/groups/pcoe/prognostic-data-repository/}. }
\label{nasa_data_2}
\end{table}

\begin{table}[h!]
\centering
\resizebox{\textwidth}{!}{%
\begin{tabular}{|l|l|c|}
\hline
\multicolumn{1}{|c|}{\textbf{Dataset}} & \multicolumn{1}{c|}{\textbf{Cell number}} & \textbf{Number of cells} \\ \hline
Feature Selection & \begin{tabular}[c]{@{}l@{}}2,   6,   8,  14,  18,  19,  26,  28,  32,  35,  37,  45,  51, 53,  55,  58,  60,  61,  65,  69,  72, \\ 76,  79,  83,  90,  91, 92, 103, 107, 109, 110, 113, 115, 116, 119, 120, 124\end{tabular} & 37 \\ \hline
Training & \begin{tabular}[c]{@{}l@{}}2,   3,   6,   8,   9,  13,  14,  16,  18,  19,  20,  21,  23, 25,  26,  28,  32,  35,  37,  42,  45,  46, \\ 50,  51,  53,  55, 56,  58,  60,  61,  63,  64,  65,  66,  69,  72,  73,  76,  79, 83,  84,  86, \\ 88,  90,  91,  92,  94,  95,  98, 100, 103, 105, 106, 107, 109, 110, 113, 115, 116, \\ 118, 119, 120, 124\end{tabular} & 63 \\ \hline
Calibration & 7,  12,  22,  48,  54,  59,  68,  77,  82, 108 & 10 \\ \hline
Testing & \begin{tabular}[c]{@{}l@{}}1,   4,   5,  10,  11,  15,  17,  24,  27,  29,  30,  31,  33, 34,  36,  38,  39,  40,  41,  43,  44, \\ 47,  49,  52,  57,  62, 67,  70,  71,  74,  75,  78,  80,  81,  85,  87,  89,  93,  96, 97,  99, \\ 101, 102, 104, 111, 112, 114, 117, 121, 122, 123\end{tabular} & 51 \\ \hline
\end{tabular}%
}
\caption{Group II: TRI dataset splitting for: feature selection, training, calibration and testing. For complete details on test conditions access \url{https://data.matr.io/1/projects/5c48dd2bc625d700019f3204}.}
\label{data_tri_split}
\end{table}

\begin{table}[h!]
\centering
\resizebox{.53\textwidth}{!}{%
\begin{tabular}{|l|l|c|}
\hline
\multicolumn{1}{|c|}{\textbf{Dataset}} & \multicolumn{1}{c|}{\textbf{Cell number}} & \textbf{Total number of cells} \\ \hline
Feature Selection & 1, 3 & 2 \\ \hline
Training & 1, 2, 3 & 3 \\ \hline
Calibration & 4 & 1 \\ \hline
Testing & 5, 6, 7, 8 & 4 \\ \hline
\end{tabular}%
}
\caption{Group III: Oxford dataset splitting for: feature selection, training, calibration and testing. For complete details on test conditions access \url{https://ora.ox.ac.uk/objects/uuid:03ba4b01-cfed-46d3-9b1a-7d4a7bdf6fac}.}
\label{data_oxford_split}
\end{table}

\clearpage

\begin{figure*}[ht!]
  \centering
  \hfil
  \begin{subfigure}[b]{.33\textwidth}
     \centering
     \caption{}
					\includegraphics[width=\linewidth]{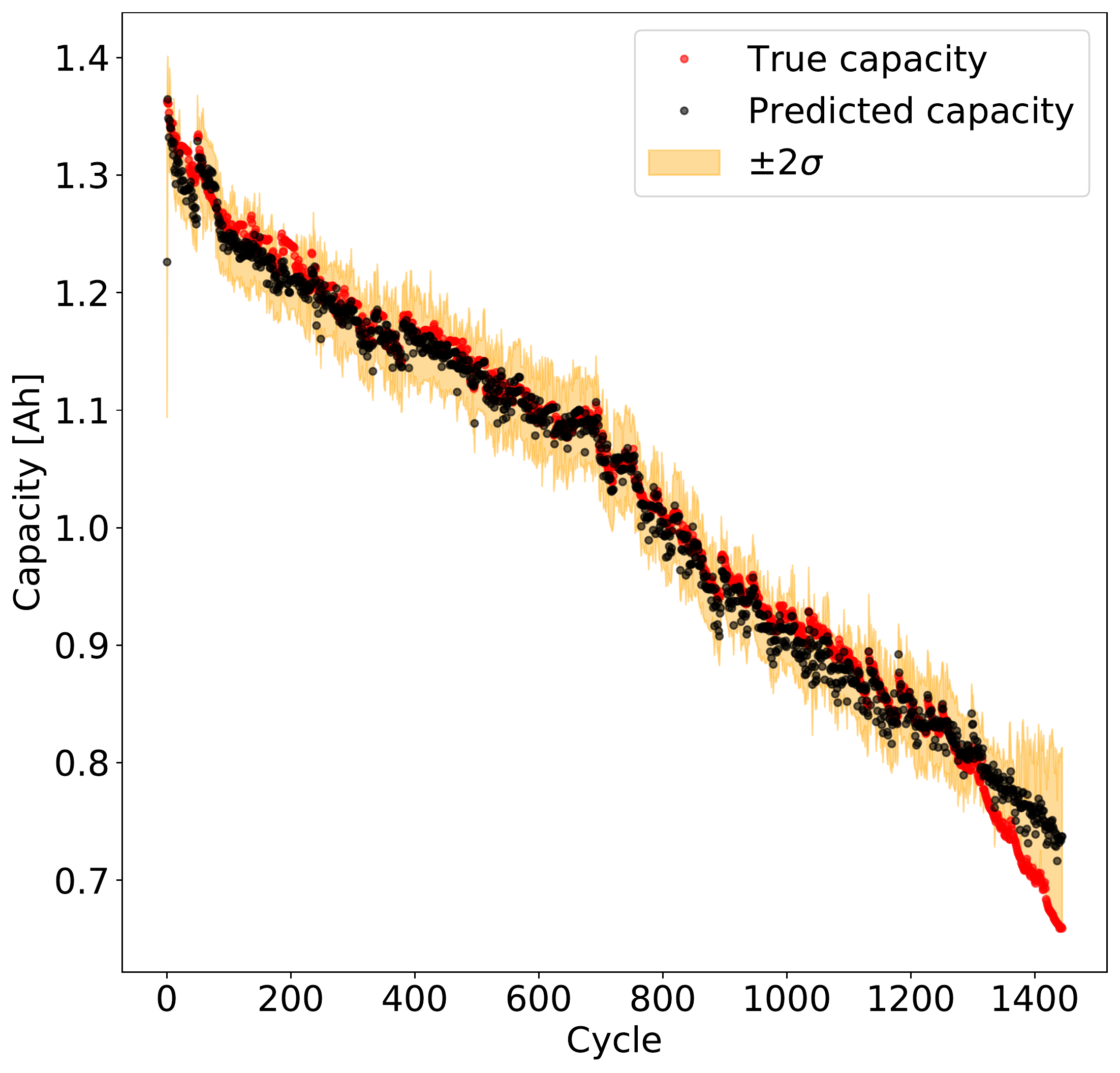}
     \label{pred_group1_BRR}
  \end{subfigure}%
  \hfill 
  \begin{subfigure}[b]{.33\textwidth}
     \centering
     \caption{}
					\includegraphics[width=\linewidth]{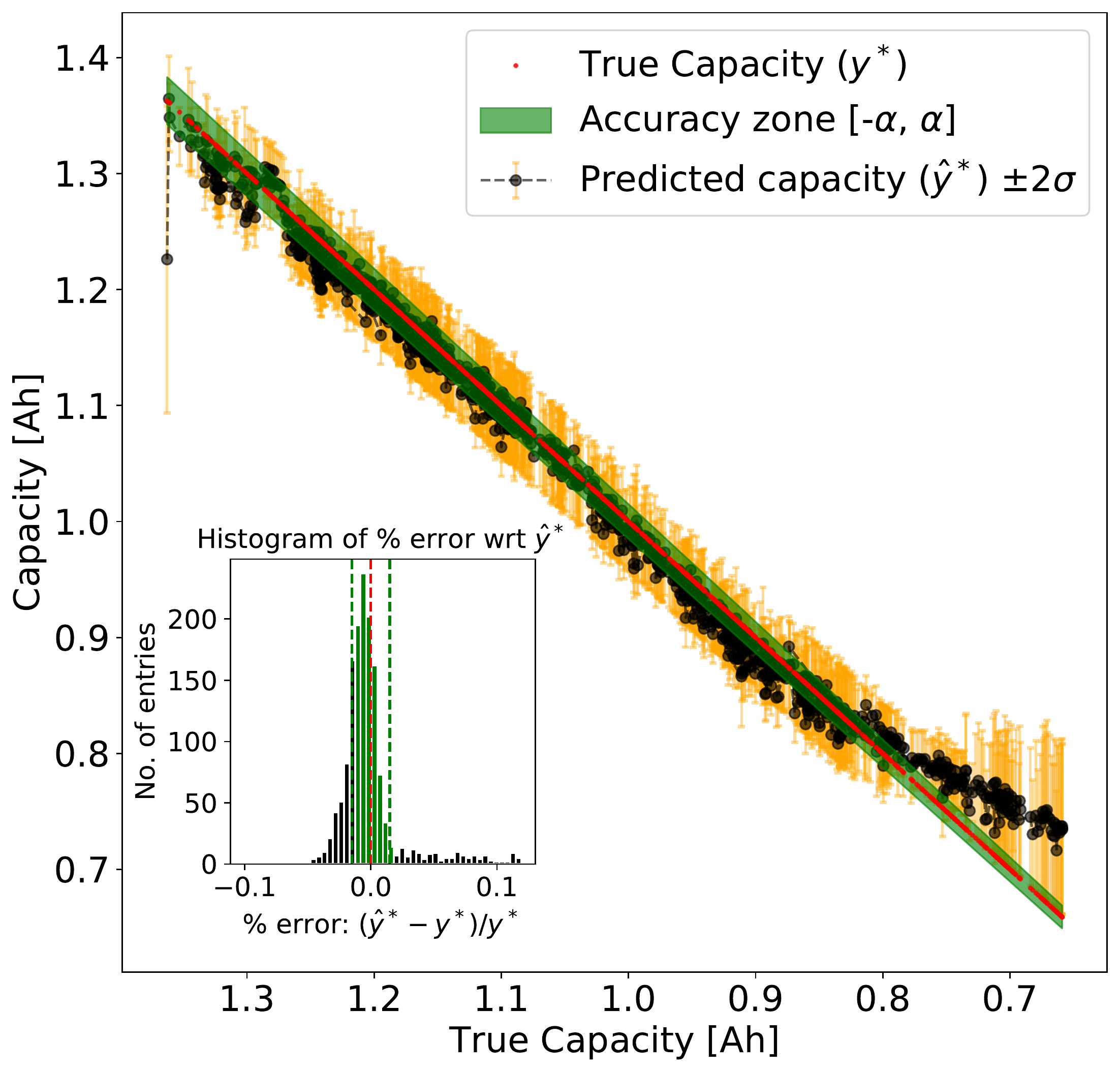}
     \label{hist_group1_BRR}
  \end{subfigure}%
  \hfill 
  \begin{subfigure}[b]{.33\textwidth}
     \centering
     \caption{}
					\includegraphics[width=\linewidth]{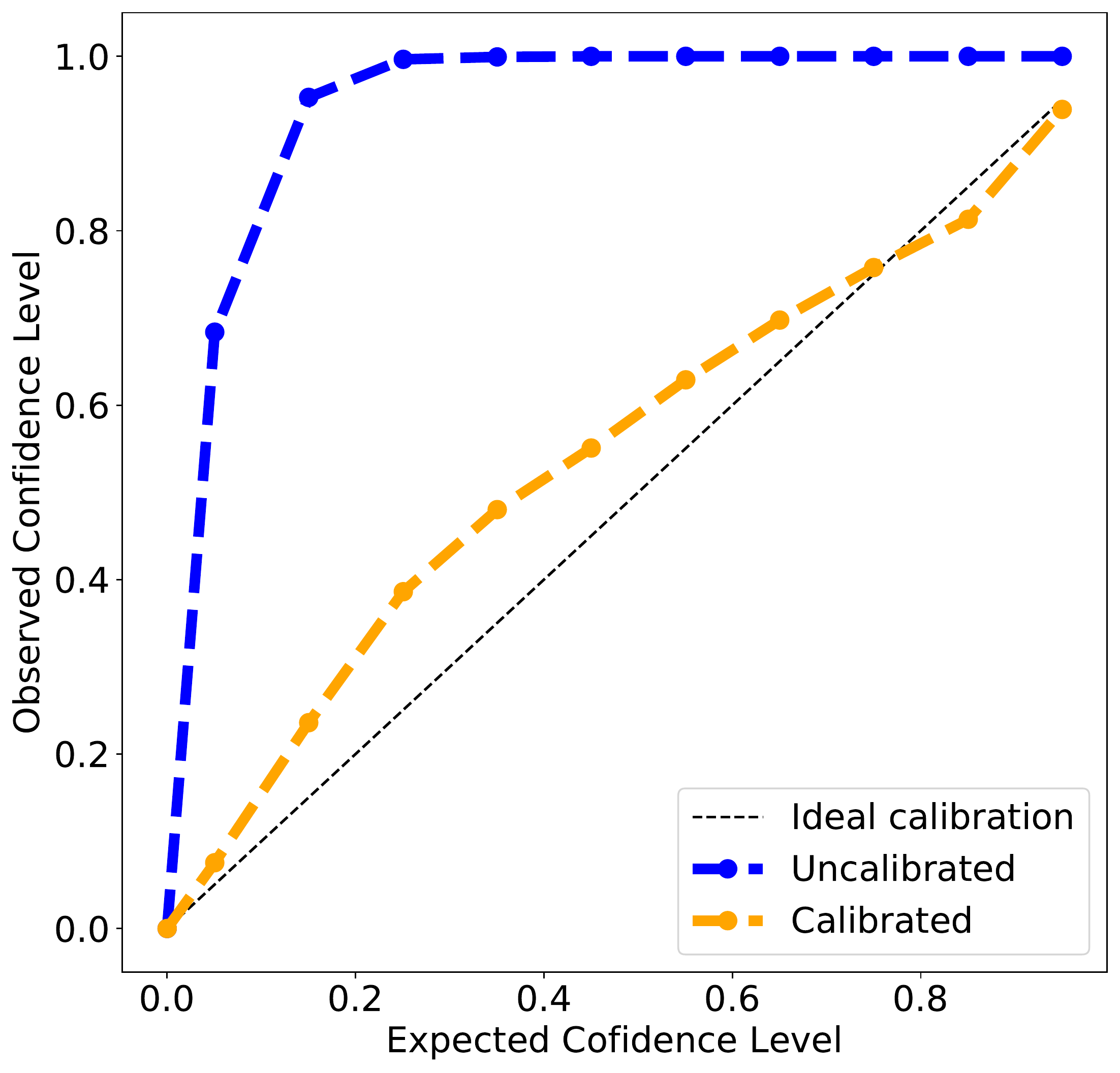}
     \label{calibration_group1_BRR}
  \end{subfigure}%
  
  \caption{\textbf{Prediction results with BRR Group I cell no. 38.} \textbf{\protect\subref{pred_group1_BRR}} Prediction as a function of cycle numbers, \textbf{\protect\subref{hist_group1_BRR}} Actual vs. predicted capacity, \textbf{\protect\subref{calibration_group1_BRR}} Calibration results.}
  \label{predictions_group1_BRR}

\medskip
  \centering
  \hfil
  \begin{subfigure}[b]{.33\textwidth}
     \centering
     \caption{}
					\includegraphics[width=\linewidth]{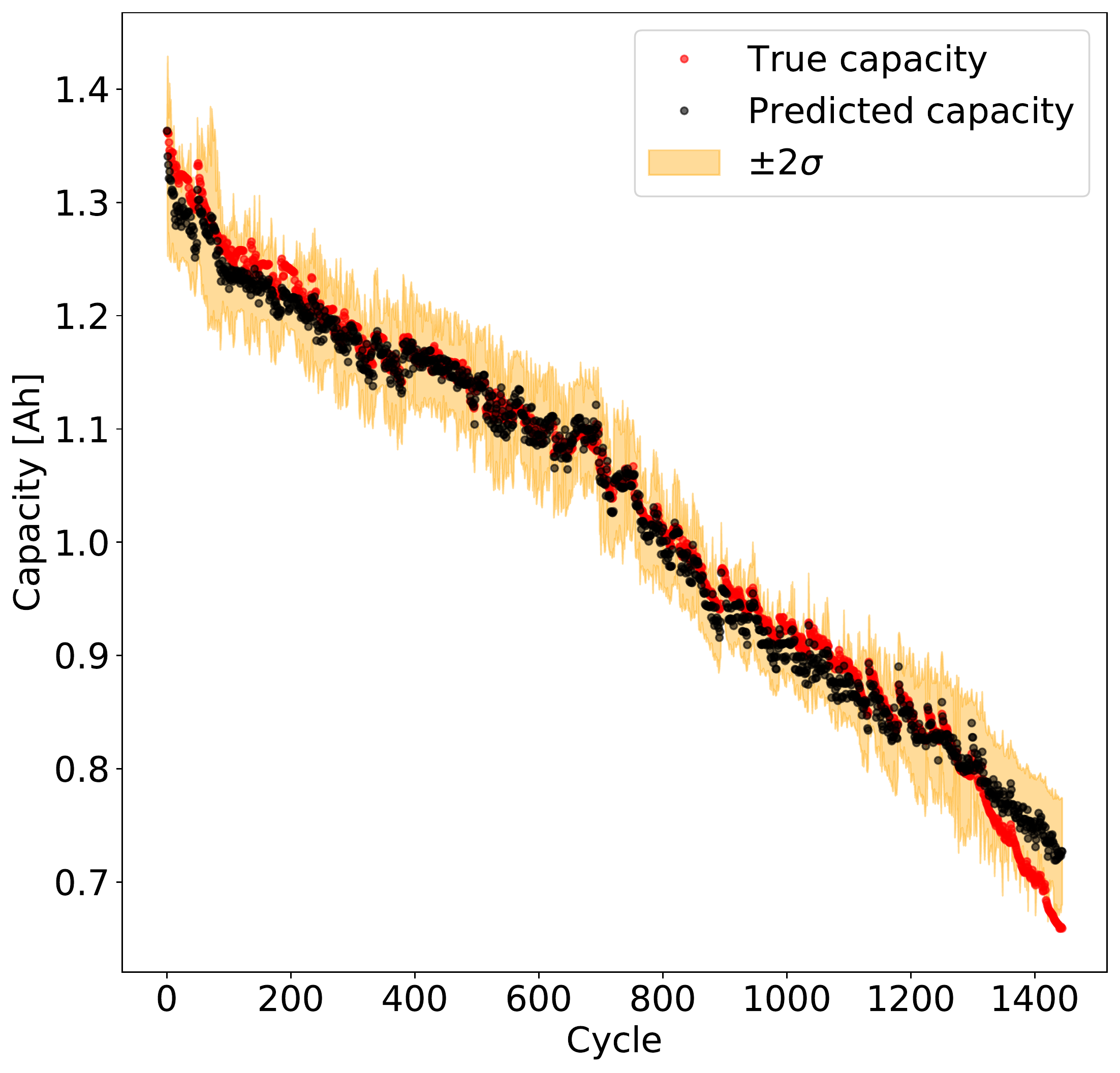}
     \label{pred_group1_GPR}
  \end{subfigure}%
  \hfill 
  \begin{subfigure}[b]{.33\textwidth}
     \centering
     \caption{}
					\includegraphics[width=\linewidth]{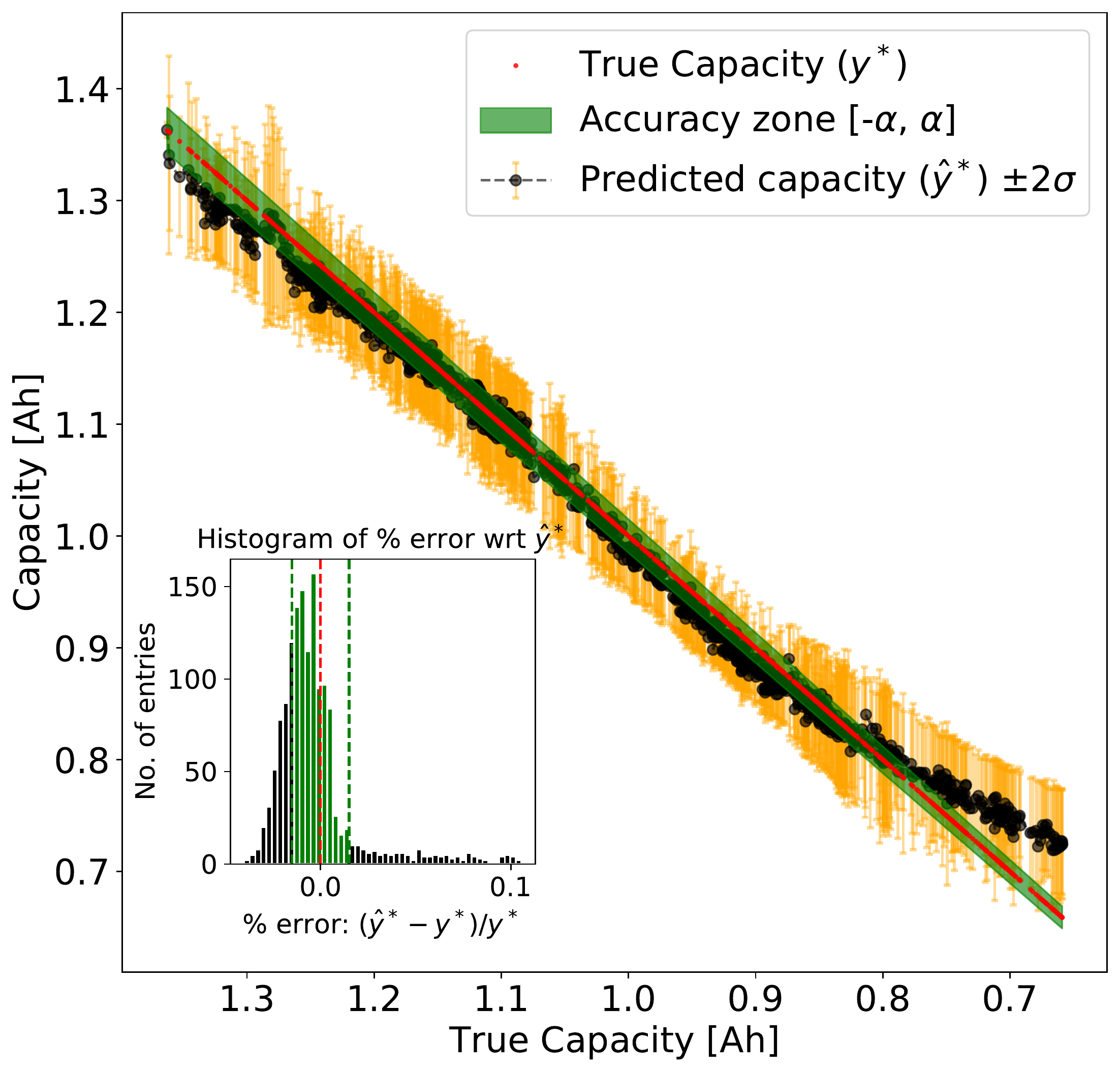}
     \label{hist_group1_GPR}
  \end{subfigure}%
  \hfill 
  \begin{subfigure}[b]{.33\textwidth}
     \centering
     \caption{}
					\includegraphics[width=\linewidth]{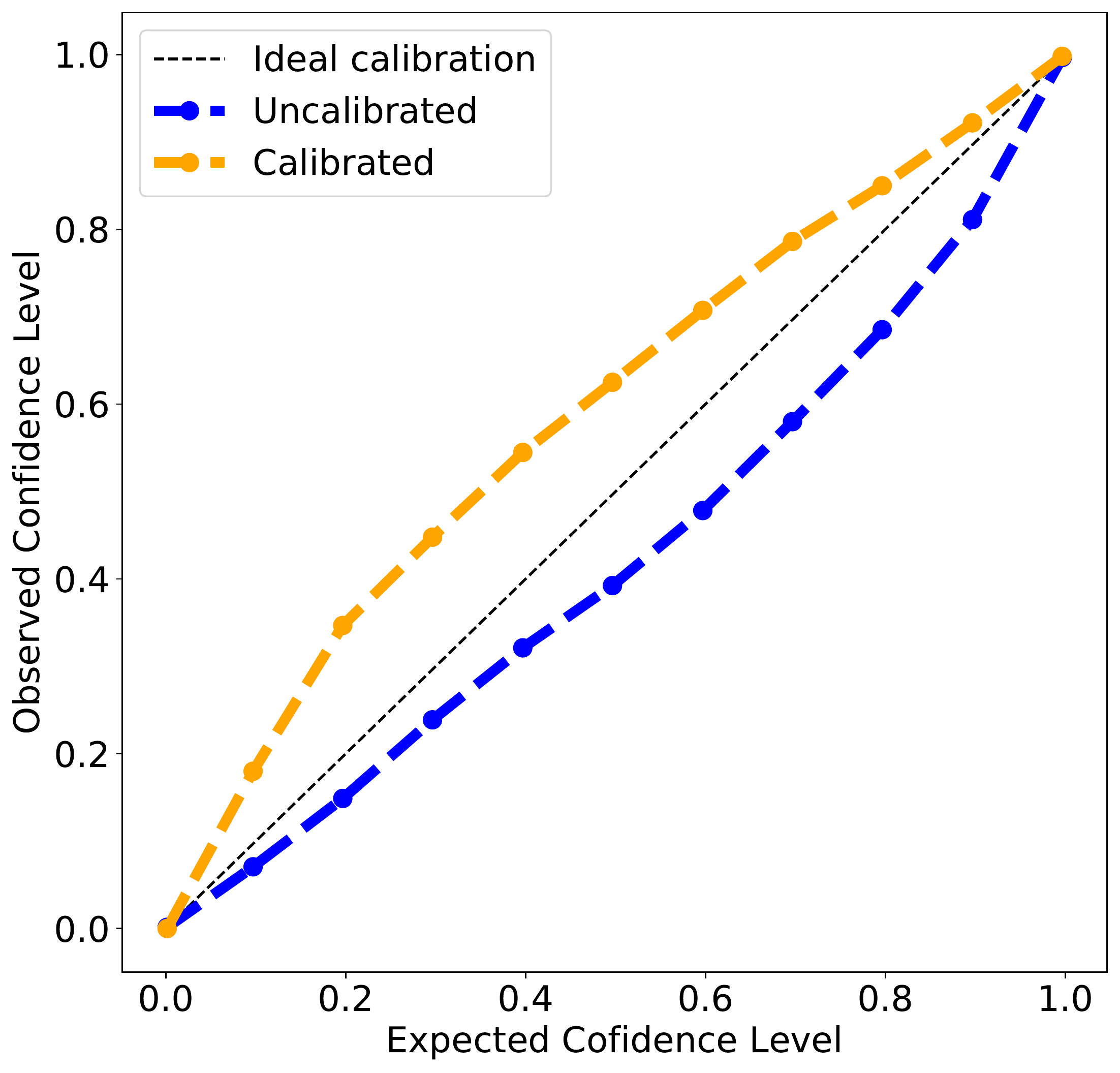}
     \label{calibration_group1_GPR}
  \end{subfigure}%
  
  \caption{\textbf{Prediction results with GPR Group I cell no. 38.} \textbf{\protect\subref{pred_group1_GPR}} GPR prediction as a function of cycle numbers, \textbf{\protect\subref{hist_group1_GPR}} GPR actual vs. predicted capacity, \textbf{\protect\subref{calibration_group1_GPR}} GPR calibration results.}
  \label{predictions_group1_GPR}

\medskip
  \centering
  \hfil
  \begin{subfigure}[b]{.33\textwidth}
     \centering
     \caption{}
					\includegraphics[width=\linewidth]{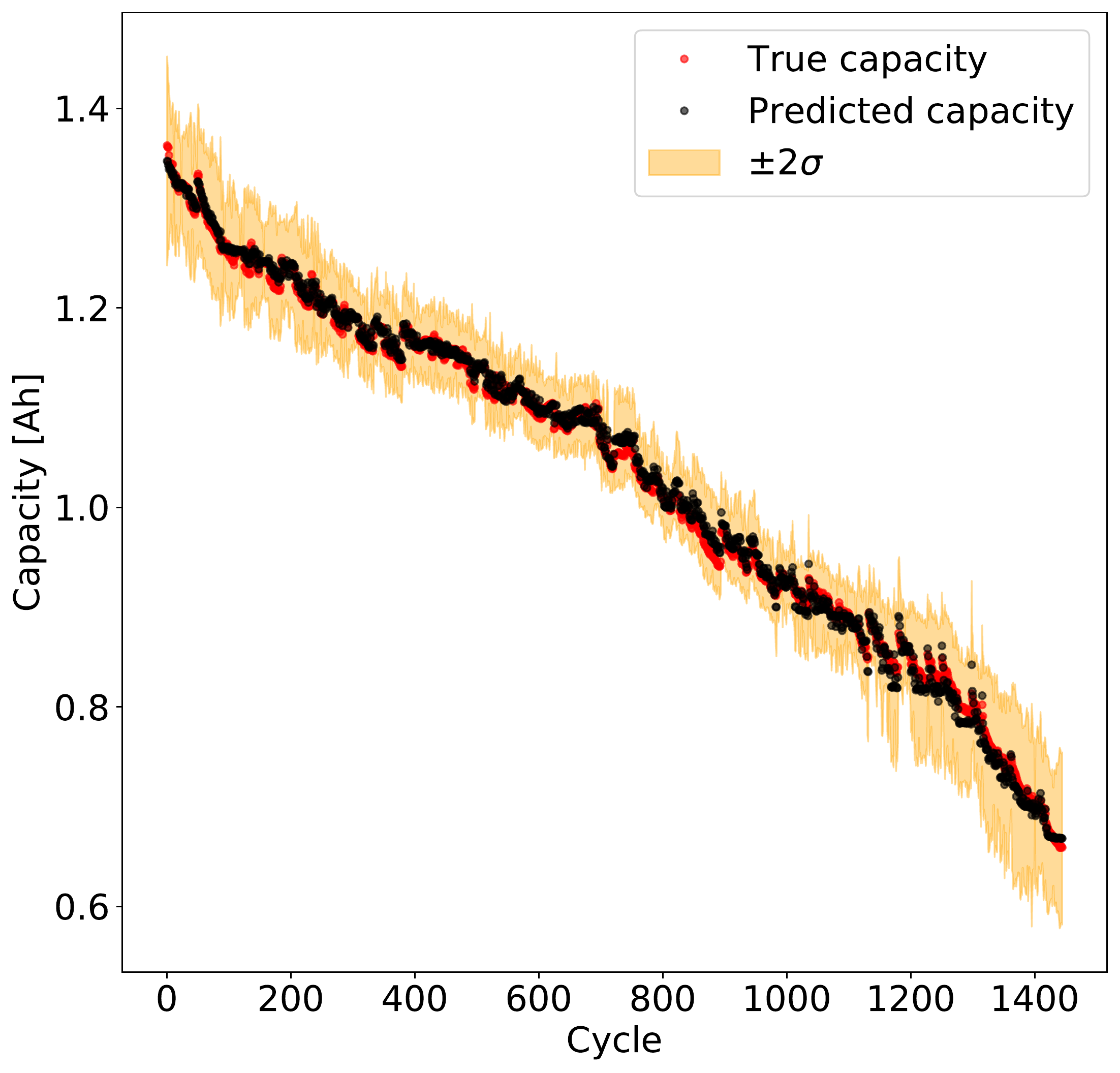}
     \label{pred_group1_RF}
  \end{subfigure}%
  \hfill 
  \begin{subfigure}[b]{.33\textwidth}
     \centering
     \caption{}
					\includegraphics[width=\linewidth]{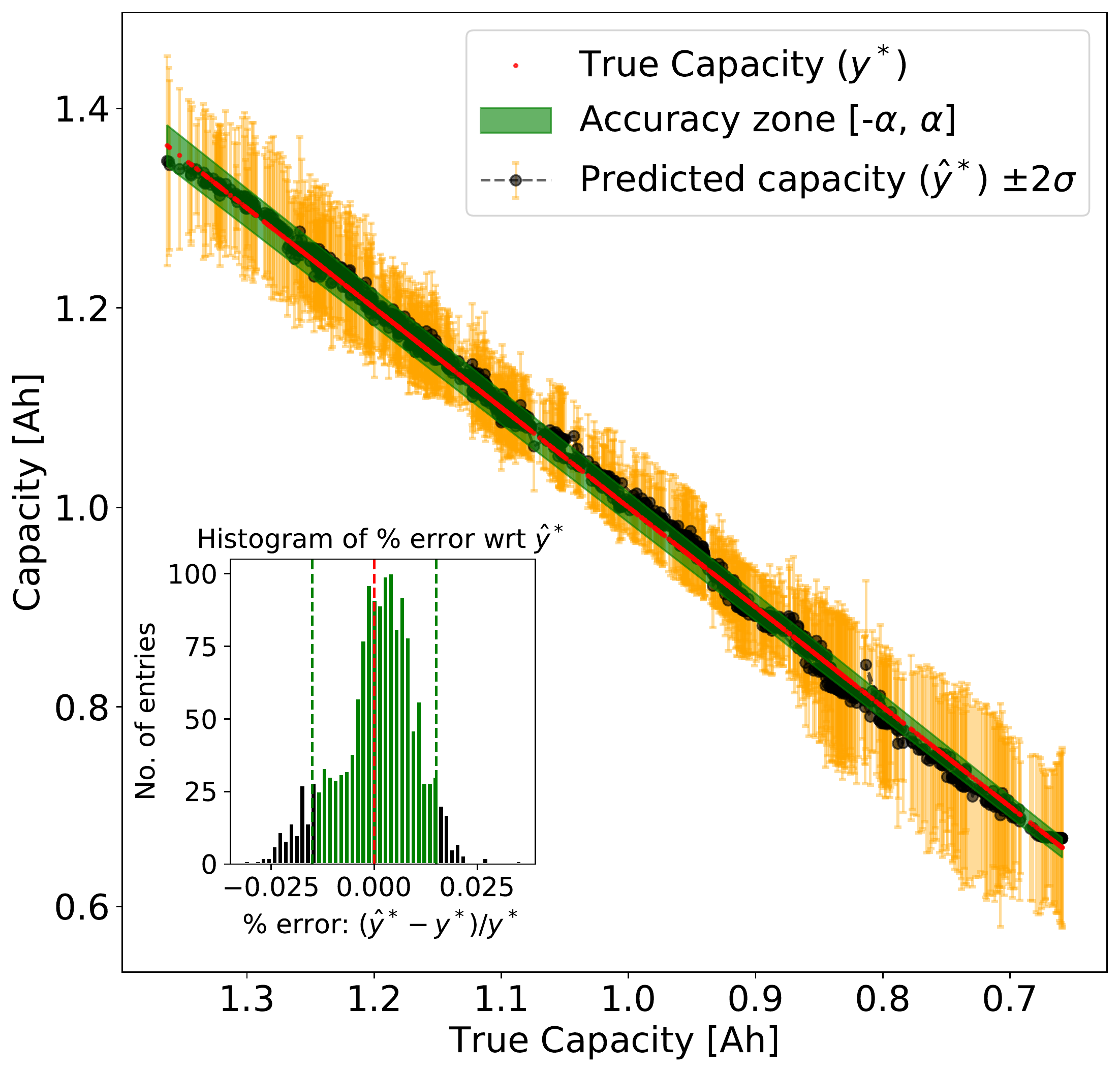}
     \label{hist_group1_RF}
  \end{subfigure}%
  \hfill 
  \begin{subfigure}[b]{.33\textwidth}
     \centering
     \caption{}
					\includegraphics[width=\linewidth]{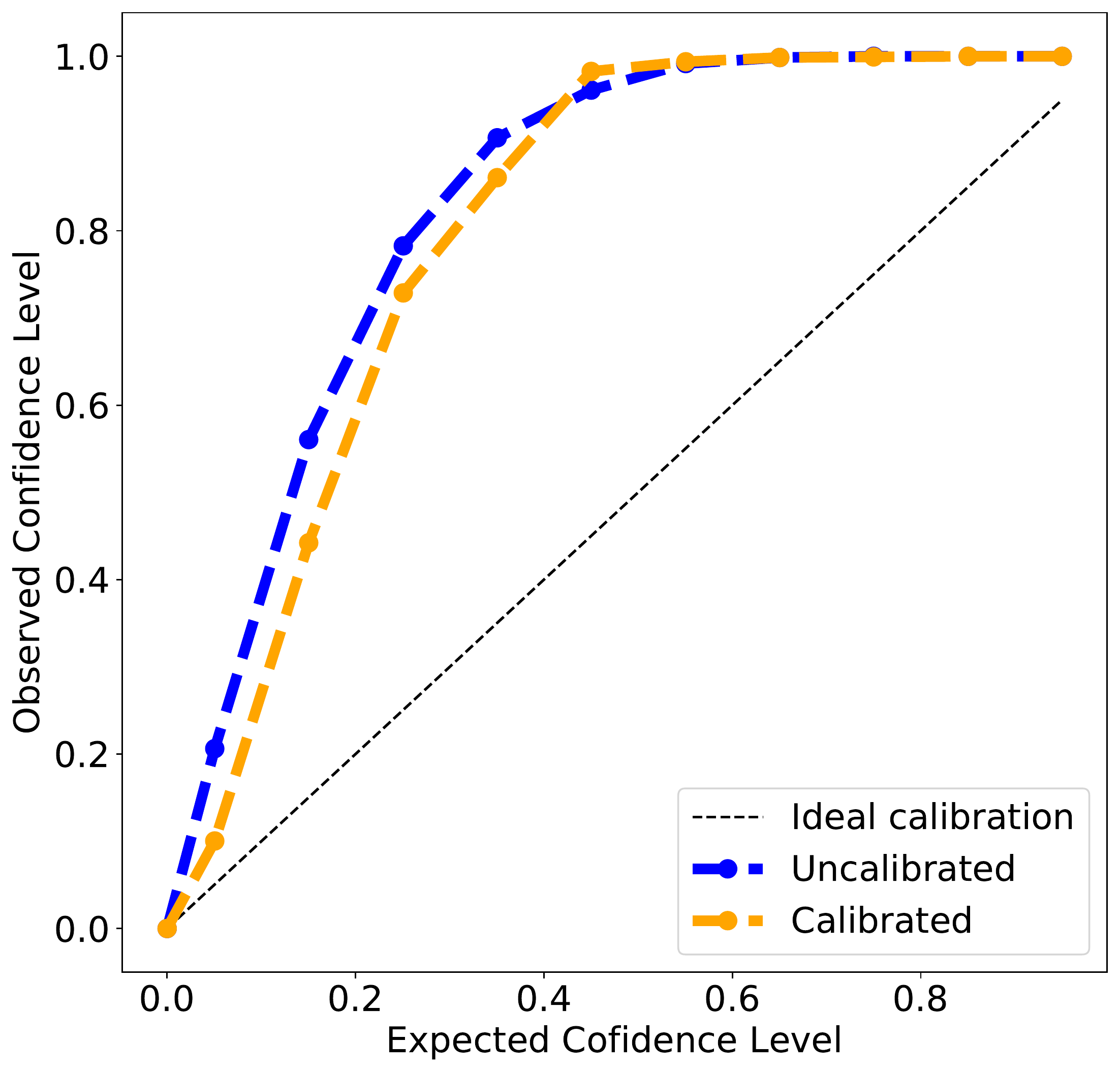}
     \label{calibration_group1_RF}
  \end{subfigure}%
  
  \caption{\textbf{Prediction results with RF Group I cell no. 38.} \textbf{\protect\subref{pred_group1_RF}} RF prediction as a function of cycle numbers, \textbf{\protect\subref{hist_group1_RF}} RF actual vs. predicted capacity, \textbf{\protect\subref{calibration_group1_RF}} RF calibration results.}
  \label{predictions_group1_RF}
 \end{figure*}

\clearpage

\begin{figure*}[ht!]
  \centering
  \hfil
  \begin{subfigure}[b]{.33\textwidth}
     \centering
     \caption{}
					\includegraphics[width=\linewidth]{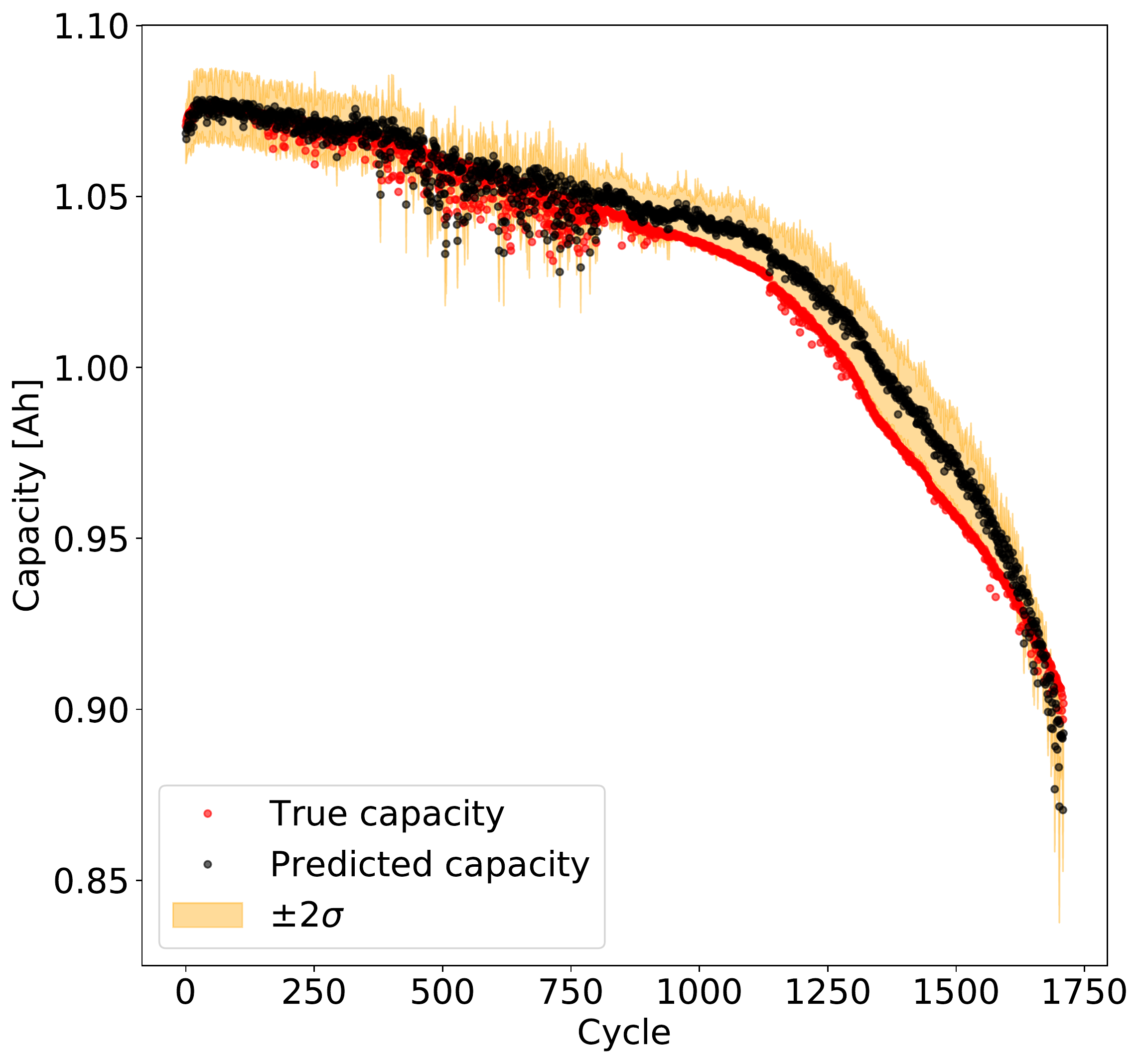}
     \label{pred_group2_BRR}
  \end{subfigure}%
  \hfill 
  \begin{subfigure}[b]{.33\textwidth}
     \centering
     \caption{}
					\includegraphics[width=\linewidth]{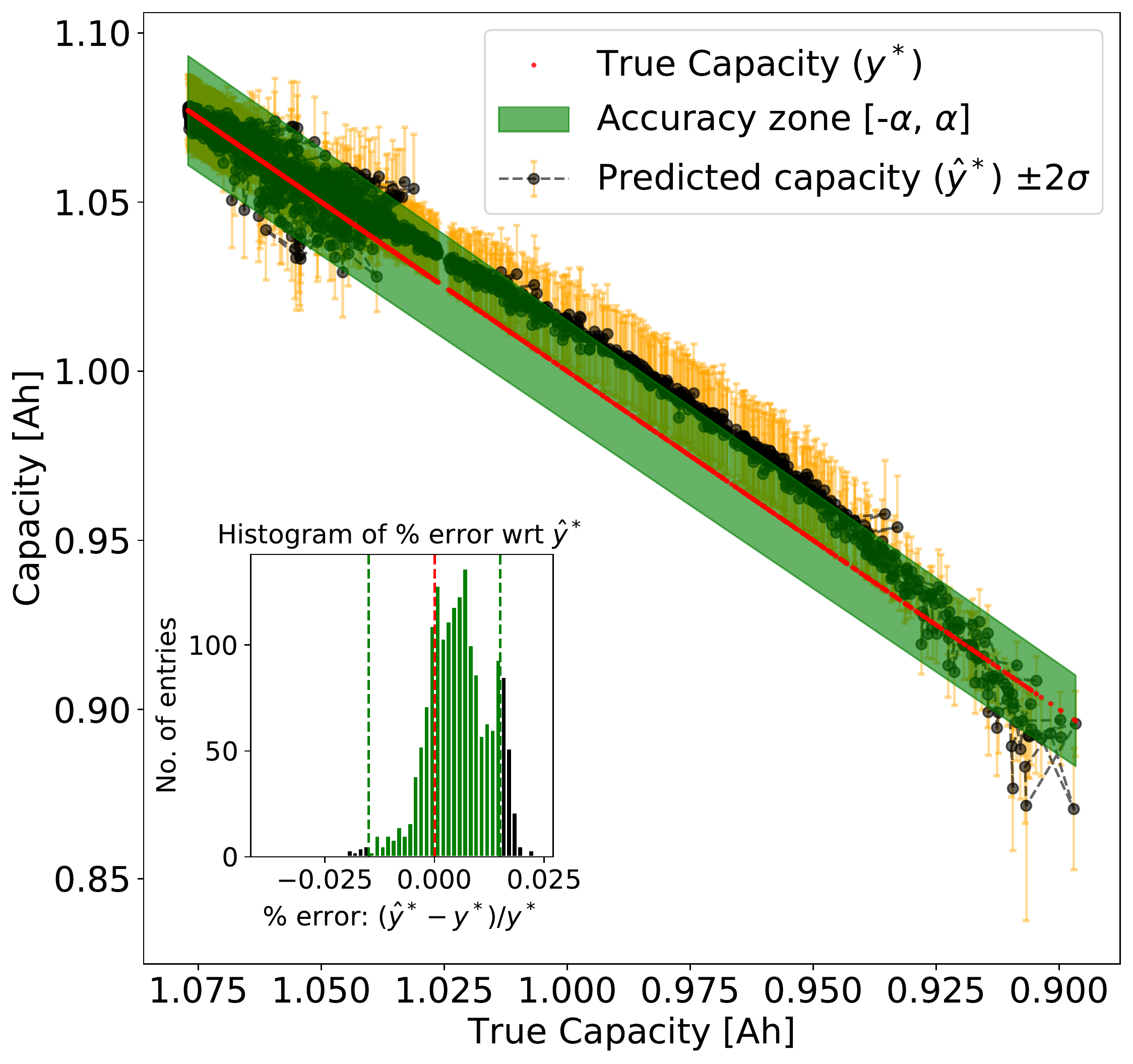}
     \label{hist_group2_BRR}
  \end{subfigure}%
  \hfill 
  \begin{subfigure}[b]{.33\textwidth}
     \centering
     \caption{}
					\includegraphics[width=\linewidth]{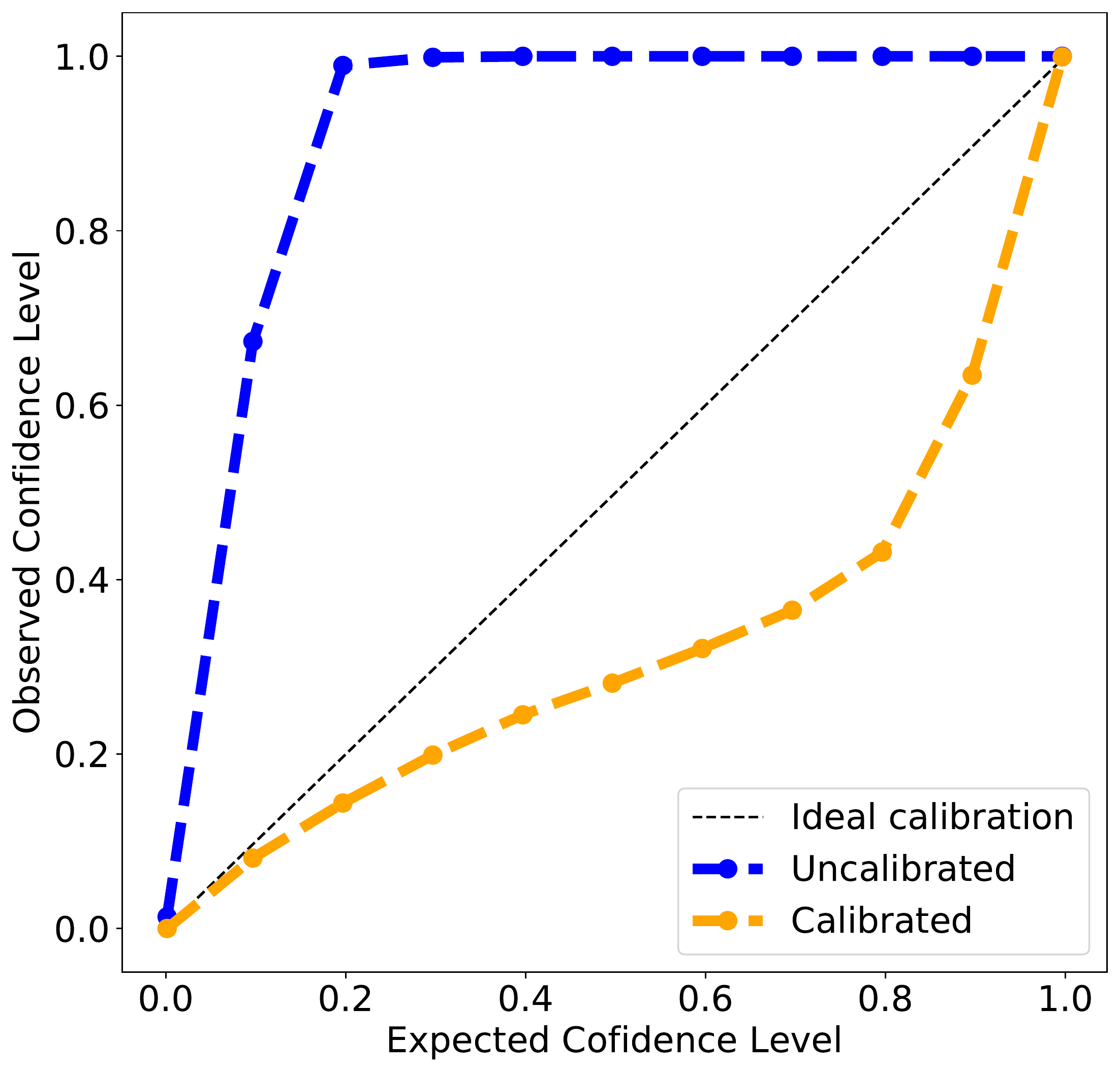}
     \label{calibration_group2_BRR}
  \end{subfigure}%
  
  \caption{\textbf{Prediction results with BRR Group II cell no. 1.} \textbf{\protect\subref{pred_group2_BRR}} BRR prediction as a function of cycle numbers, \textbf{\protect\subref{hist_group2_BRR}} BRR actual vs. predicted capacity, \textbf{\protect\subref{calibration_group2_BRR}} BRR calibration results.}
  \label{predictions_group2_BRR}
 \medskip
  \centering
  \hfil
  \begin{subfigure}[b]{.33\textwidth}
     \centering
     \caption{}
					\includegraphics[width=\linewidth]{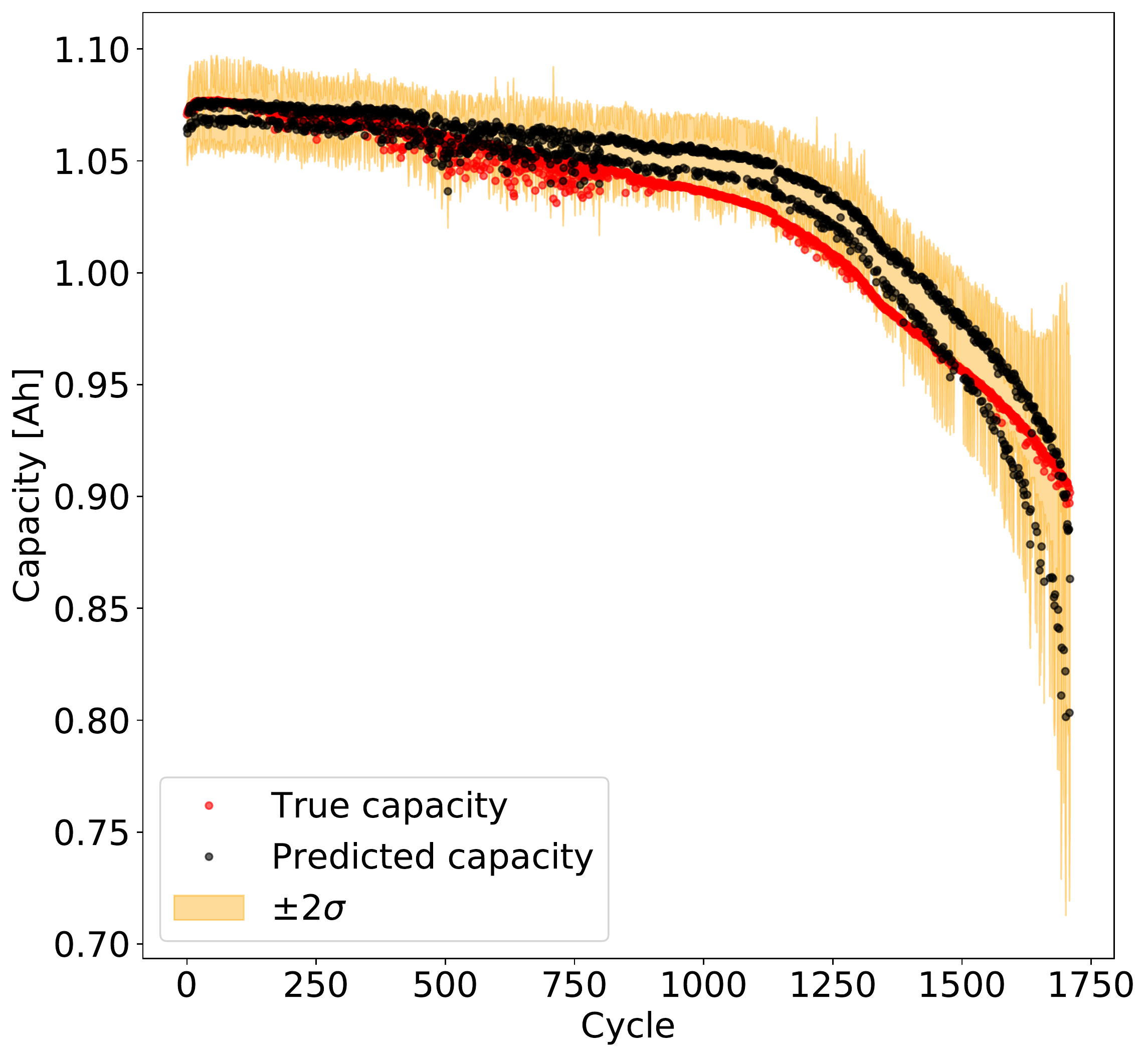}
     \label{pred_group2_GPR}
  \end{subfigure}%
  \hfill 
  \begin{subfigure}[b]{.33\textwidth}
     \centering
     \caption{}
					\includegraphics[width=\linewidth]{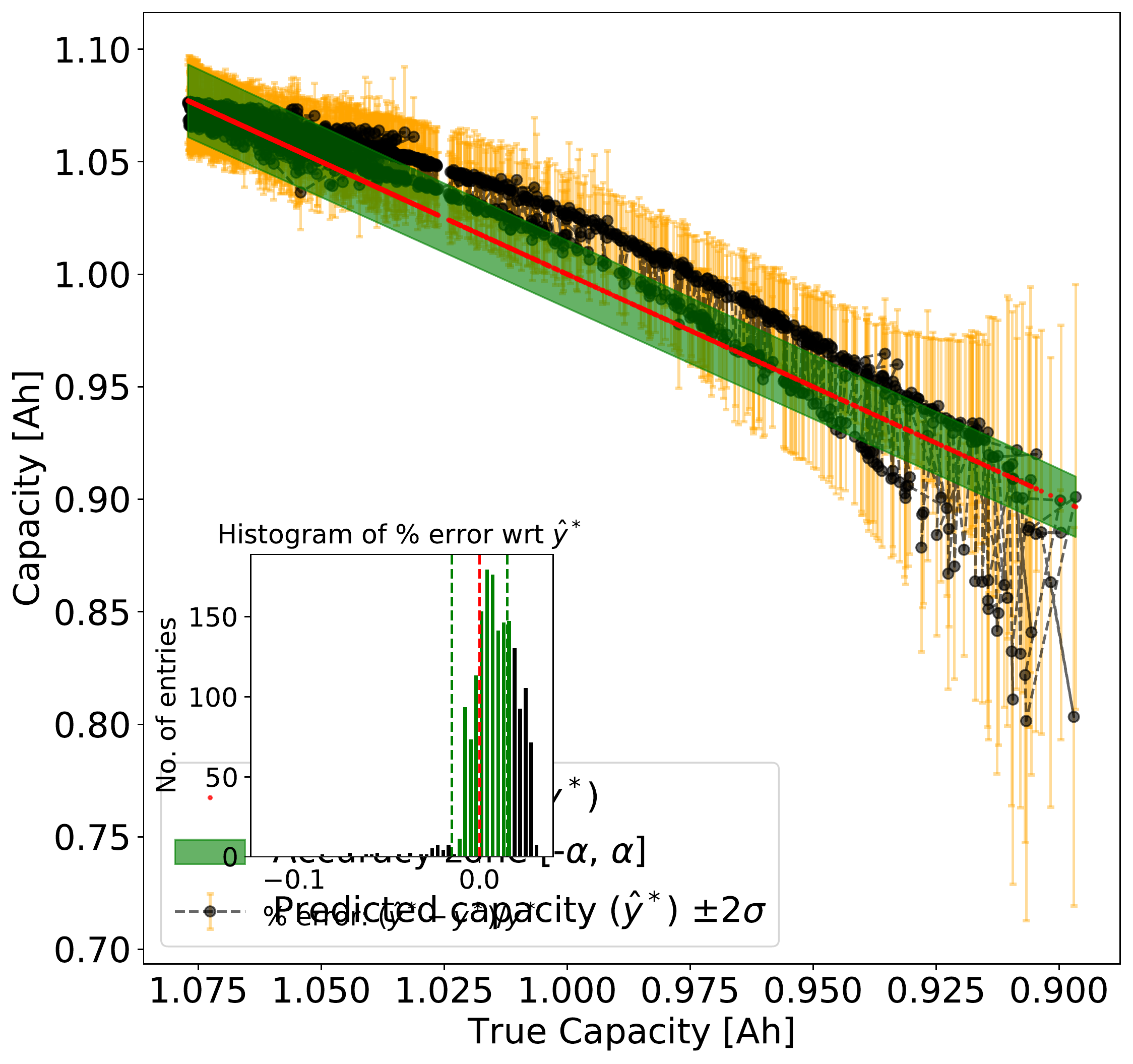}
     \label{hist_group2_GPR}
  \end{subfigure}%
  \hfill 
  \begin{subfigure}[b]{.33\textwidth}
     \centering
     \caption{}
					\includegraphics[width=\linewidth]{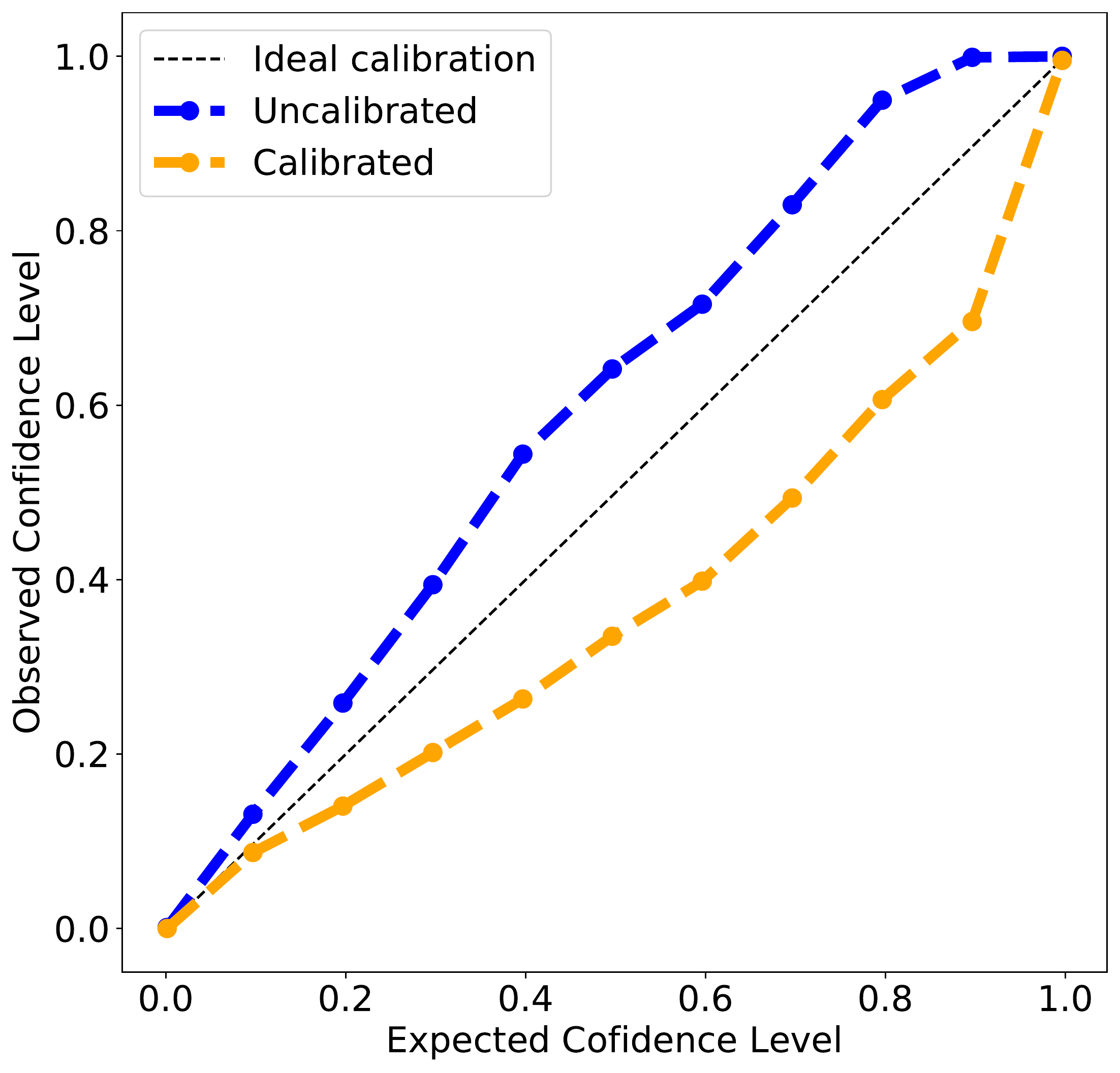}
     \label{calibration_group2_GPR}
  \end{subfigure}%
  
  \caption{\textbf{Prediction results with GPR Group II cell no. 1.} \textbf{\protect\subref{pred_group2_GPR}} GPR prediction as a function of cycle numbers, \textbf{\protect\subref{hist_group2_GPR}} GPR actual vs. predicted capacity, \textbf{\protect\subref{calibration_group2_GPR}} GPR calibration results.}
  \label{predictions_group2_GPR}
\medskip
  \centering
  \hfil
  \begin{subfigure}[b]{.33\textwidth}
     \centering
     \caption{}
					\includegraphics[width=\linewidth]{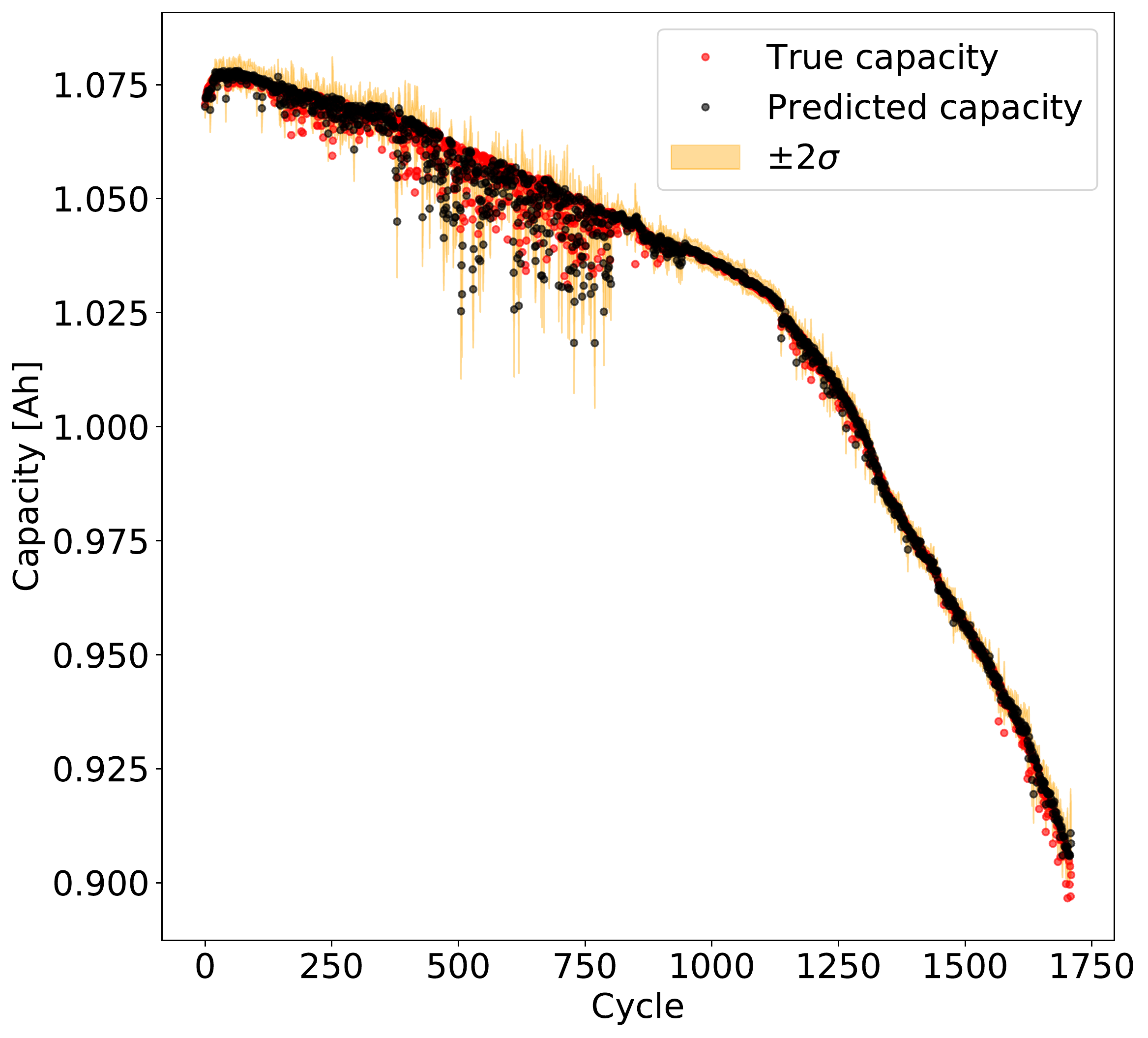}
     \label{pred_group2_RF}
  \end{subfigure}%
  \hfill 
  \begin{subfigure}[b]{.33\textwidth}
     \centering
     \caption{}
					\includegraphics[width=\linewidth]{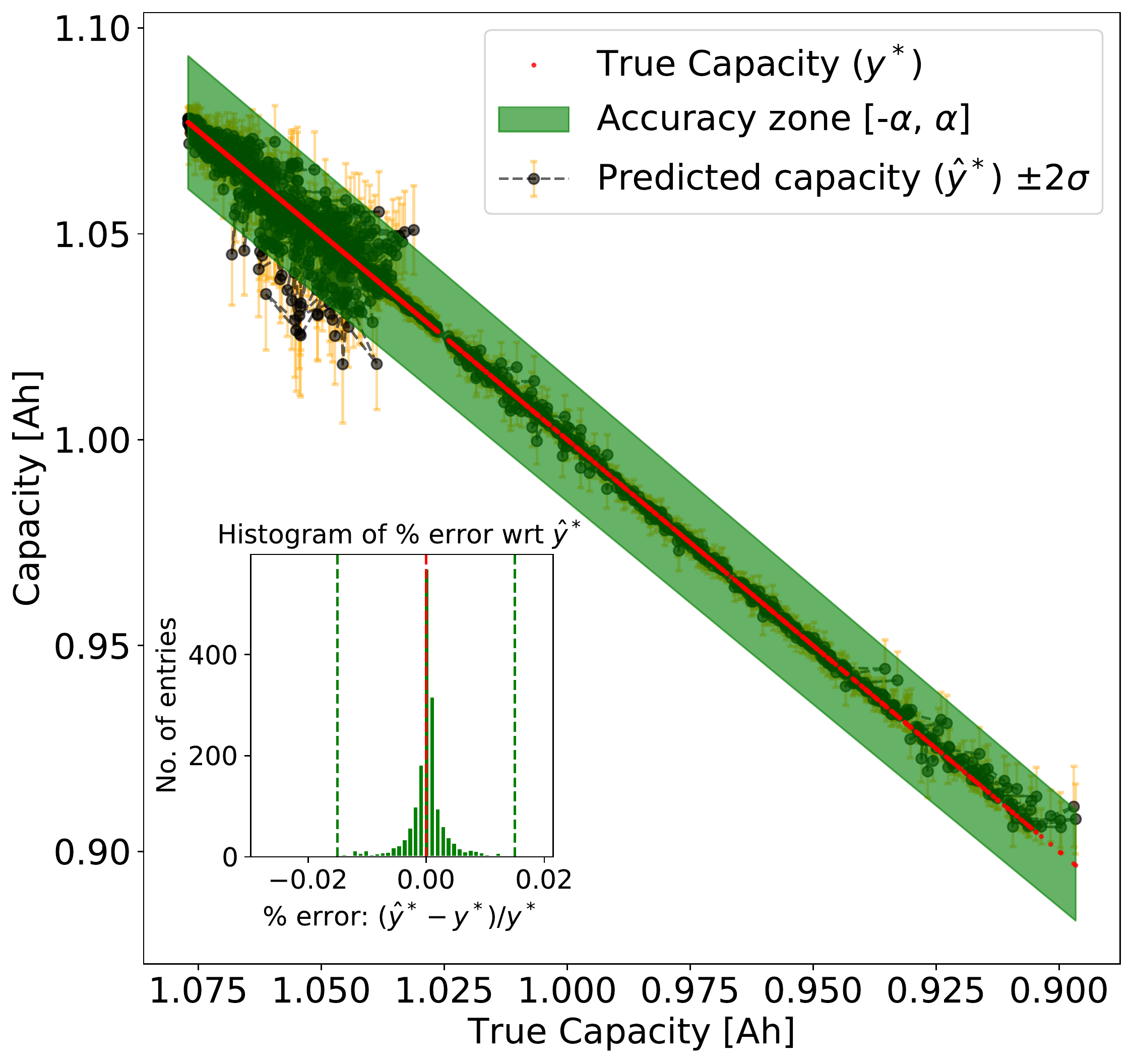}
     \label{hist_group2_RF}
  \end{subfigure}%
  \hfill 
  \begin{subfigure}[b]{.33\textwidth}
     \centering
     \caption{}
					\includegraphics[width=\linewidth]{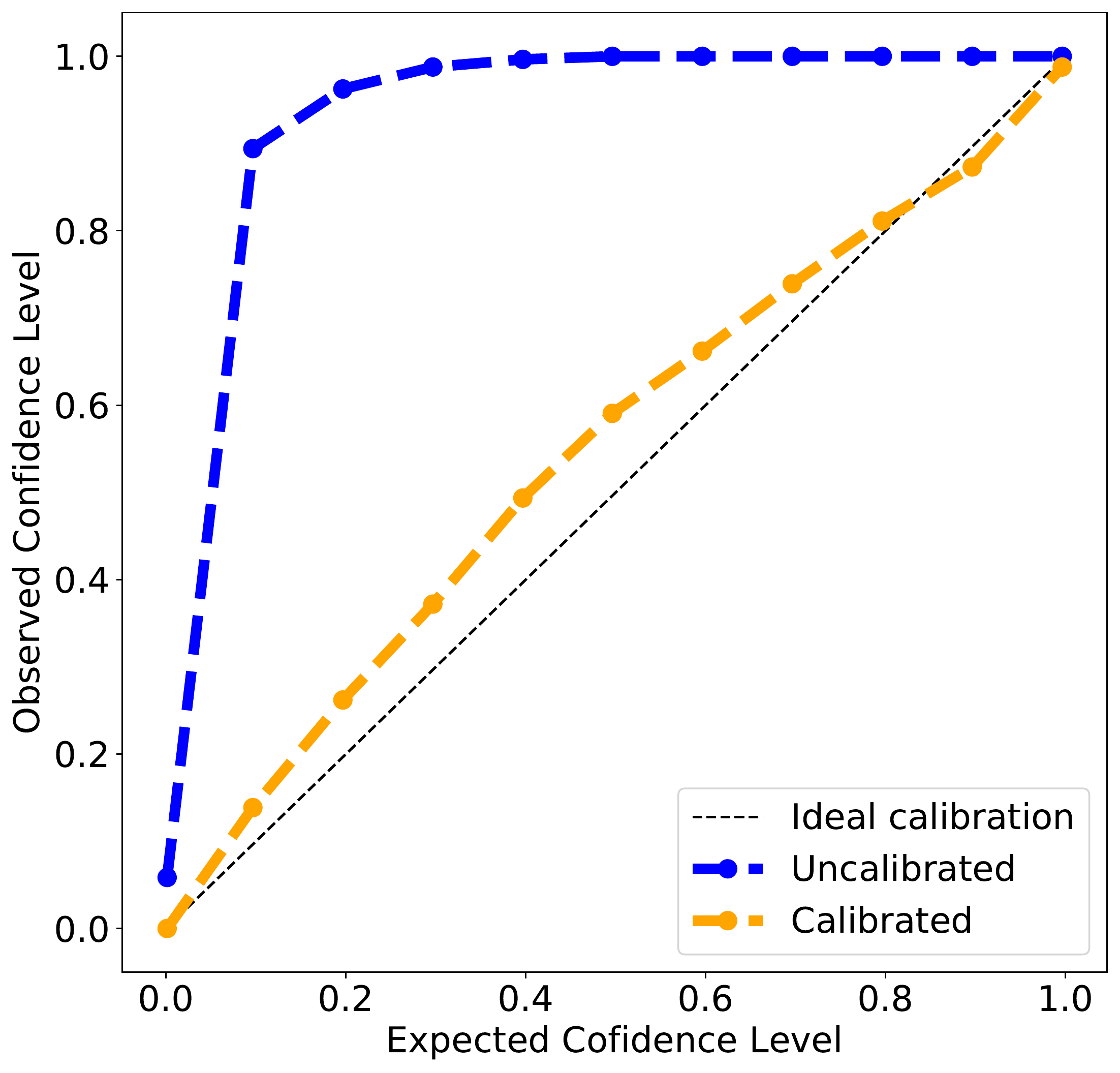}
     \label{calibration_group2_RF}
  \end{subfigure}%
  
  \caption{\textbf{Prediction results with RF Group II cell no. 1.} \textbf{\protect\subref{pred_group2_RF}} RF prediction as a function of cycle numbers, \textbf{\protect\subref{hist_group2_RF}} RF actual vs. predicted capacity, \textbf{\protect\subref{calibration_group2_RF}} RF calibration results.}
  \label{predictions_group2_RF}
 \end{figure*}
 
\clearpage
 
\begin{figure*}[ht!]
  \centering
  \hfil
  \begin{subfigure}[b]{.33\textwidth}
     \centering
     \caption{}
					\includegraphics[width=\linewidth]{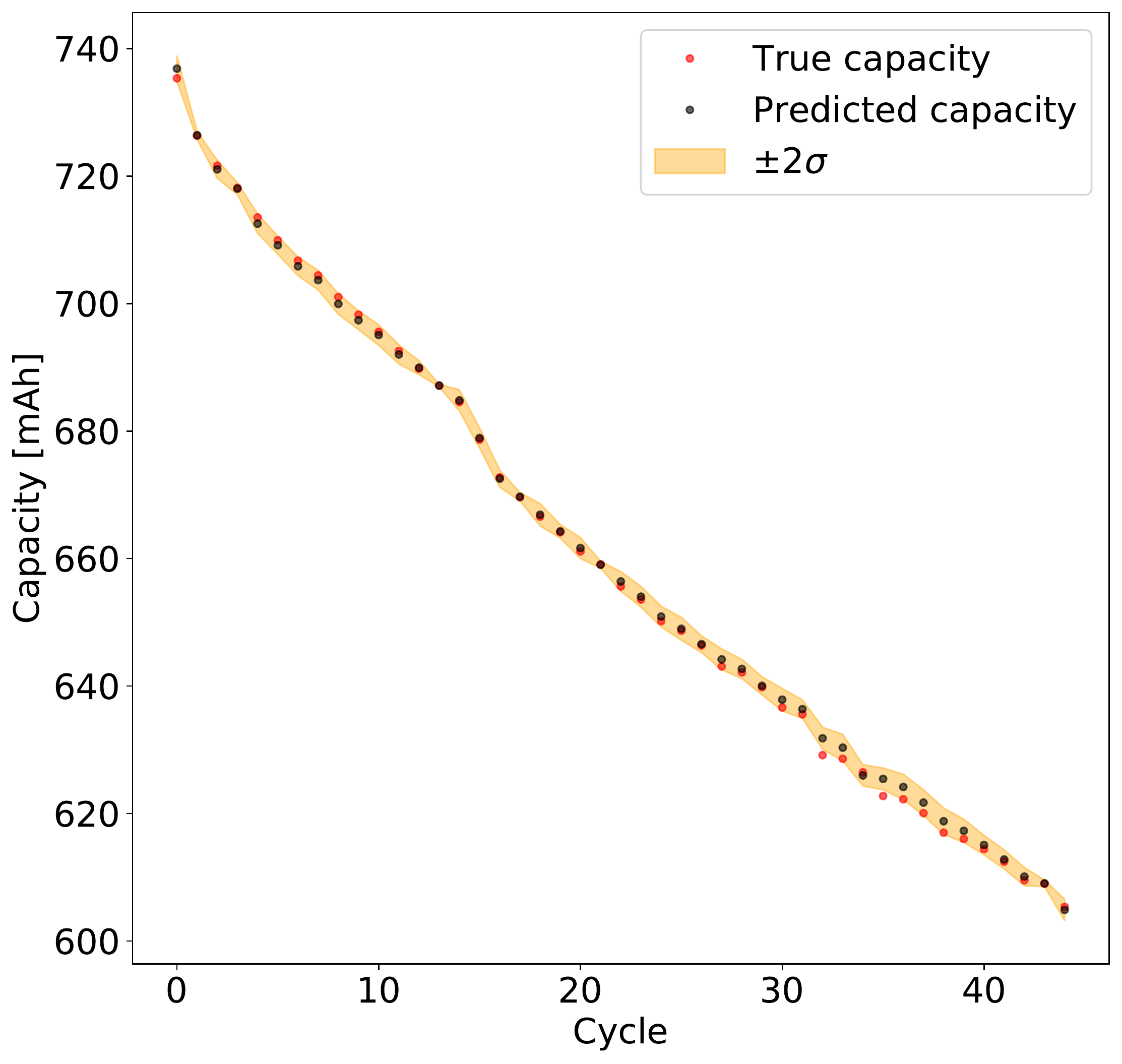}
     \label{pred_group3_BRR}
  \end{subfigure}%
  \hfill 
  \begin{subfigure}[b]{.33\textwidth}
     \centering
     \caption{}
					\includegraphics[width=\linewidth]{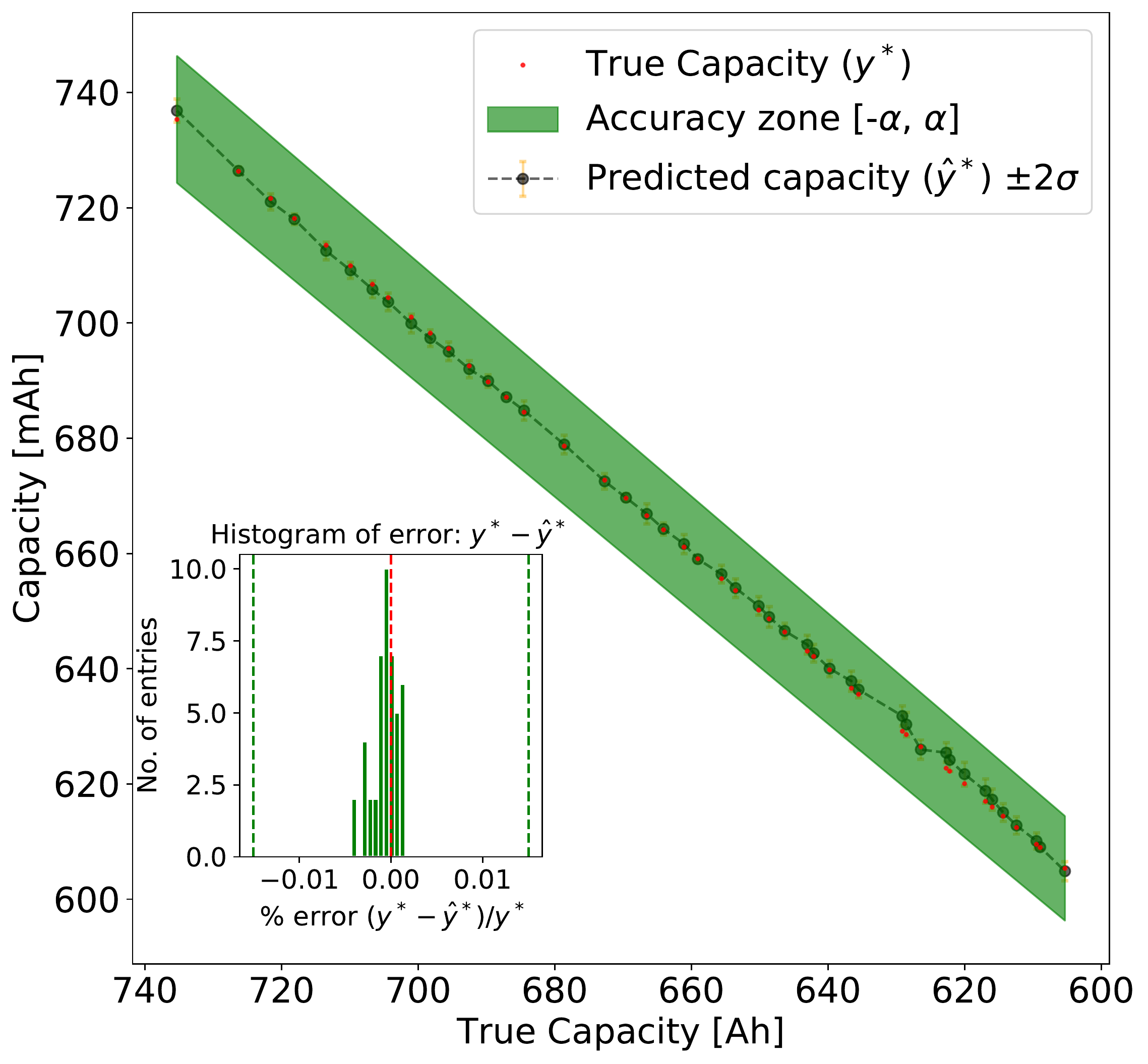}
     \label{hist_group3_BRR}
  \end{subfigure}%
  \hfill 
  \begin{subfigure}[b]{.33\textwidth}
     \centering
     \caption{}
					\includegraphics[width=\linewidth]{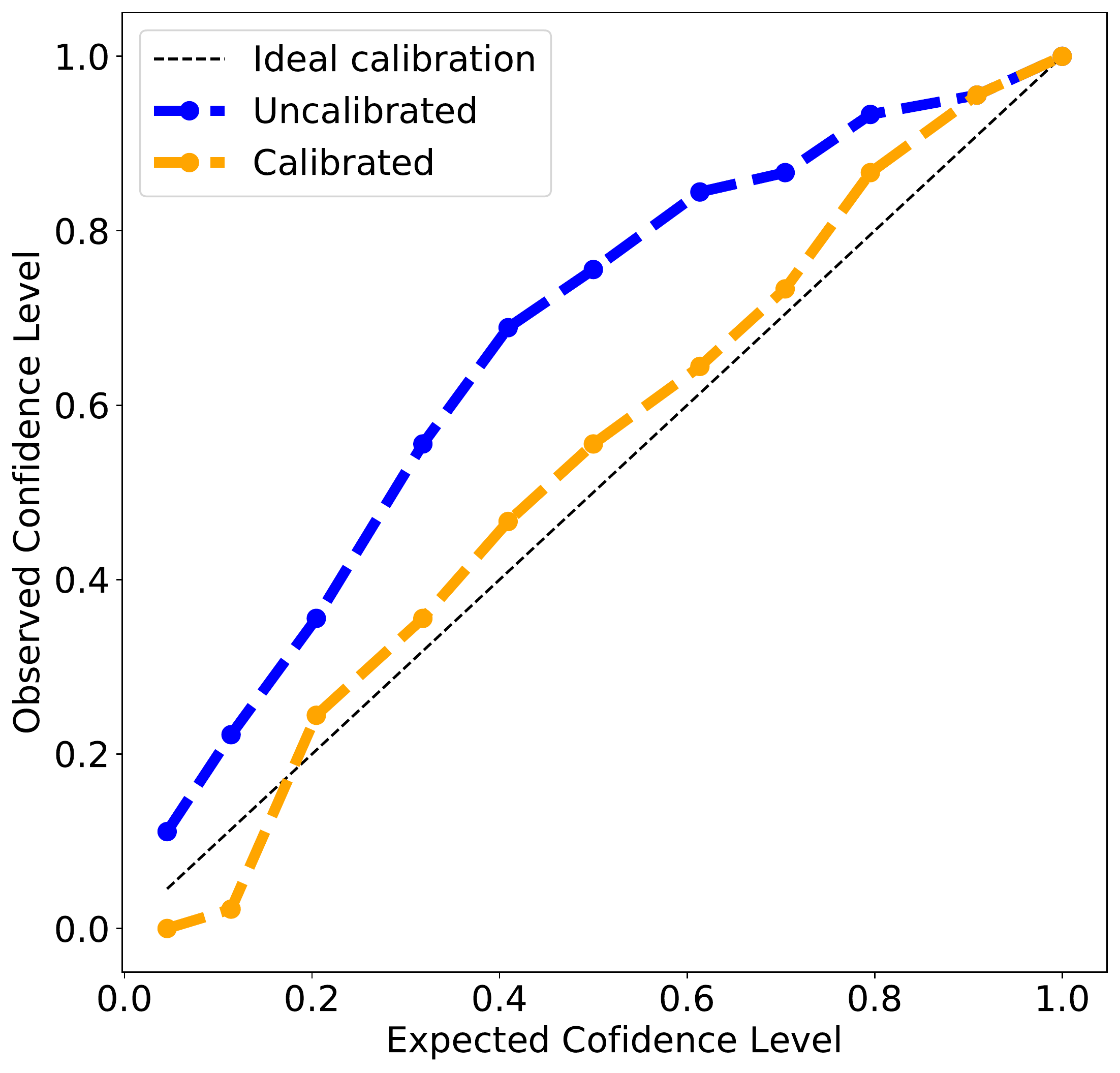}
     \label{calibration_group3_BRR}
  \end{subfigure}%
  
  \caption{\textbf{Prediction results with BRR Group III cell no. 5.} \textbf{\protect\subref{pred_group3_BRR}} BRR prediction as a function of cycle numbers, \textbf{\protect\subref{hist_group3_BRR}} BRR actual vs. predicted capacity, \textbf{\protect\subref{calibration_group3_BRR}} BRR calibration results.}
  \label{predictions_group3_BRR}
\medskip
  \centering
  \hfil
  \begin{subfigure}[b]{.33\textwidth}
     \centering
     \caption{}
					\includegraphics[width=\linewidth]{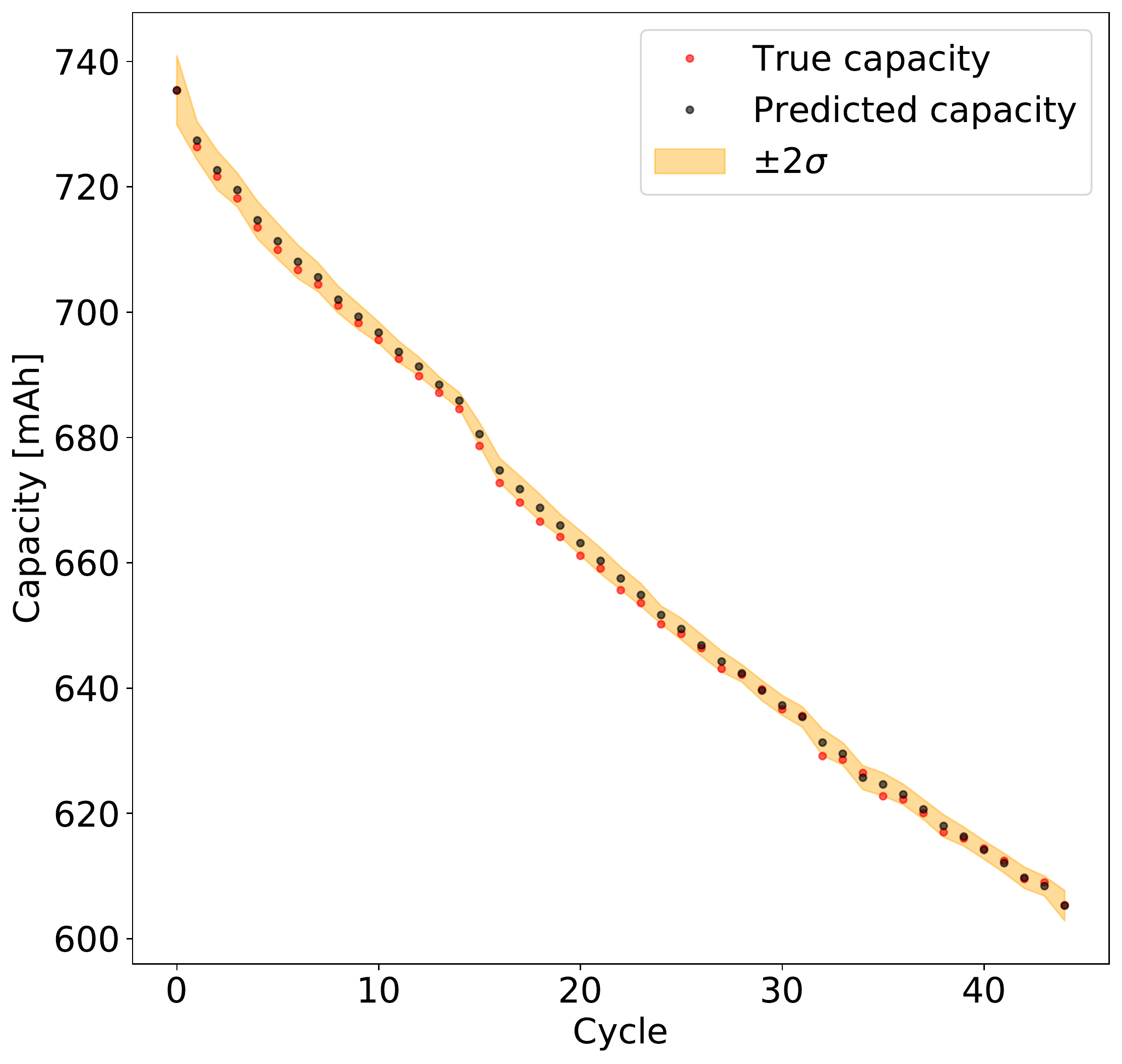}
     \label{pred_group3_GPR}
  \end{subfigure}%
  \hfill 
  \begin{subfigure}[b]{.33\textwidth}
     \centering
     \caption{}
					\includegraphics[width=\linewidth]{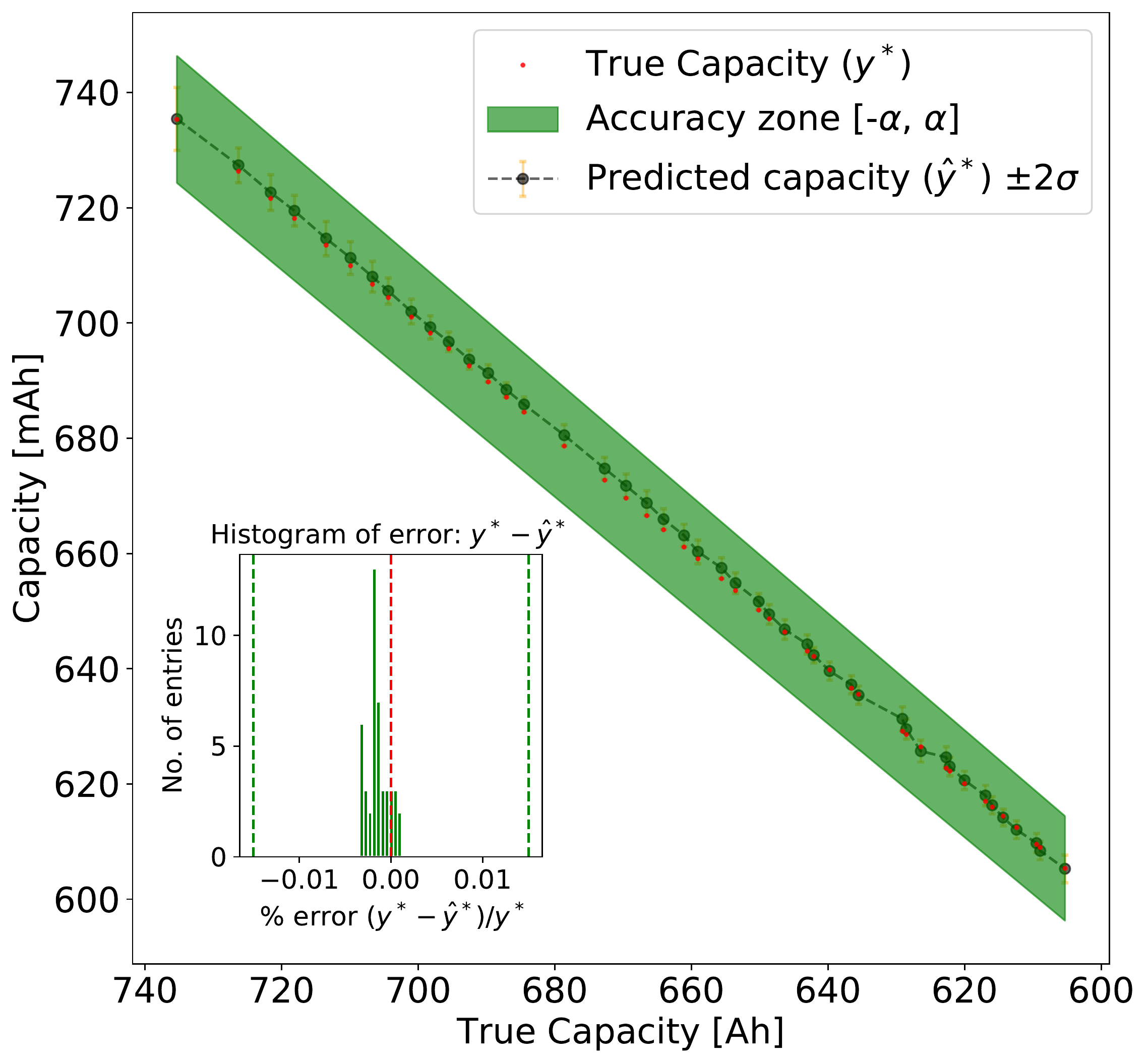}
     \label{hist_group3_GPR}
  \end{subfigure}%
  \hfill 
  \begin{subfigure}[b]{.33\textwidth}
     \centering
     \caption{}
					\includegraphics[width=\linewidth]{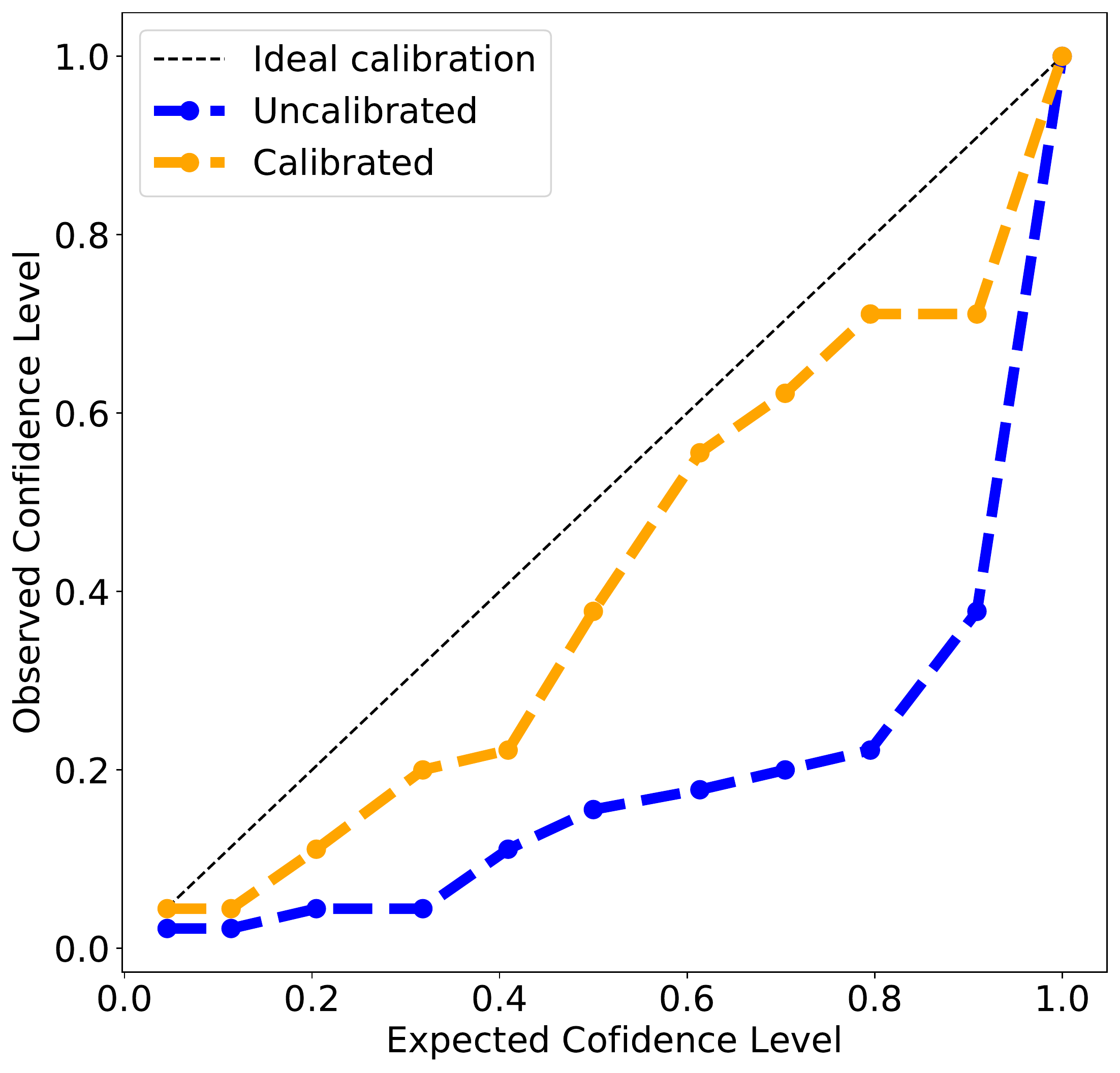}
     \label{calibration_group3_GPR}
  \end{subfigure}%
  
  \caption{\textbf{Prediction results with GPR Group III cell no. 5.} \textbf{\protect\subref{pred_group3_GPR}} GPR prediction as a function of cycle numbers, \textbf{\protect\subref{hist_group3_GPR}} GPR actual vs. predicted capacity, \textbf{\protect\subref{calibration_group3_GPR}} GPR calibration results.}
  \label{predictions_group3_GPR}
\medskip
  \centering
  \hfil
  \begin{subfigure}[b]{.33\textwidth}
     \centering
     \caption{}
					\includegraphics[width=\linewidth]{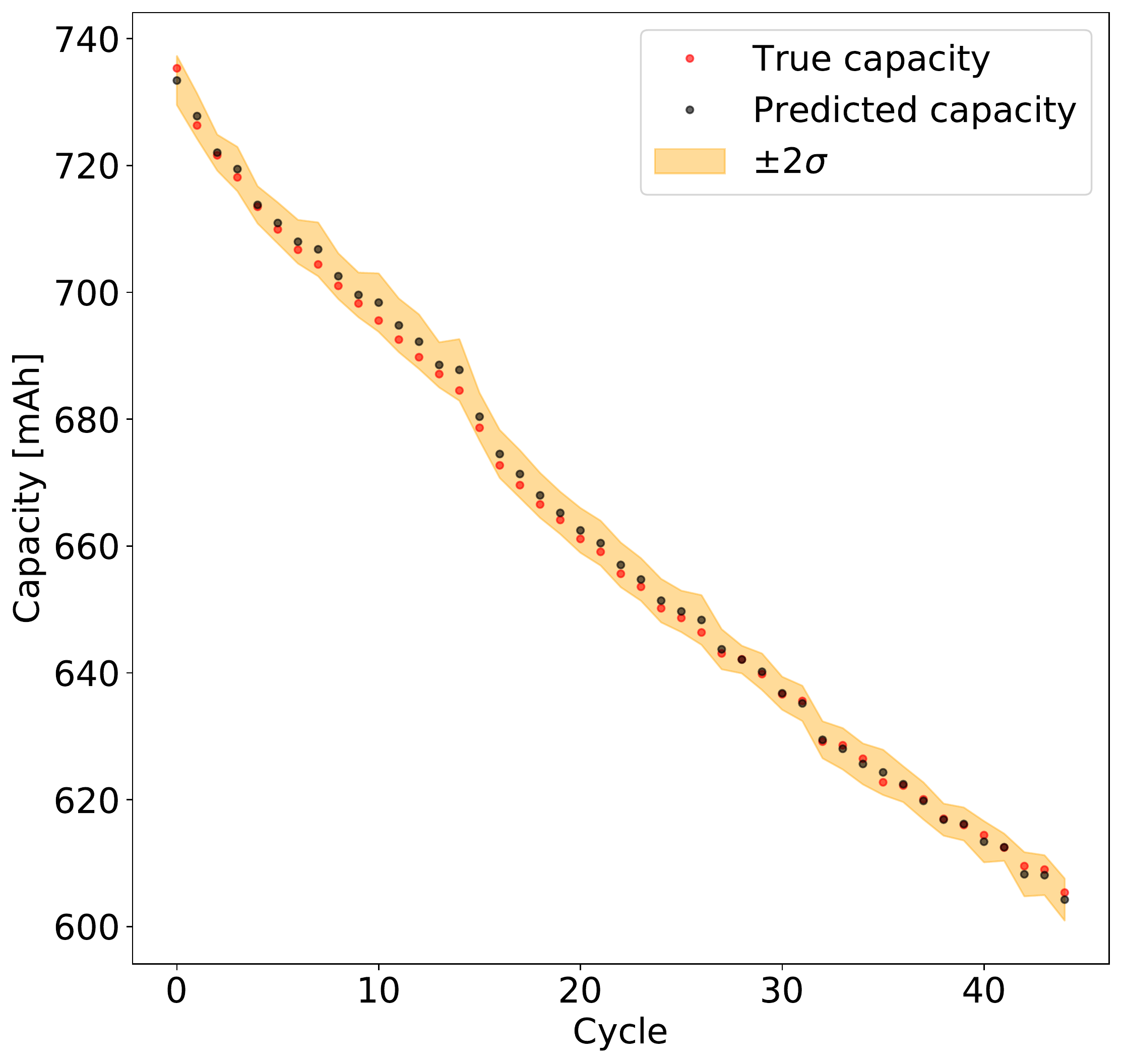}
     \label{pred_group3_RF}
  \end{subfigure}%
  \hfill 
  \begin{subfigure}[b]{.33\textwidth}
     \centering
     \caption{}
					\includegraphics[width=\linewidth]{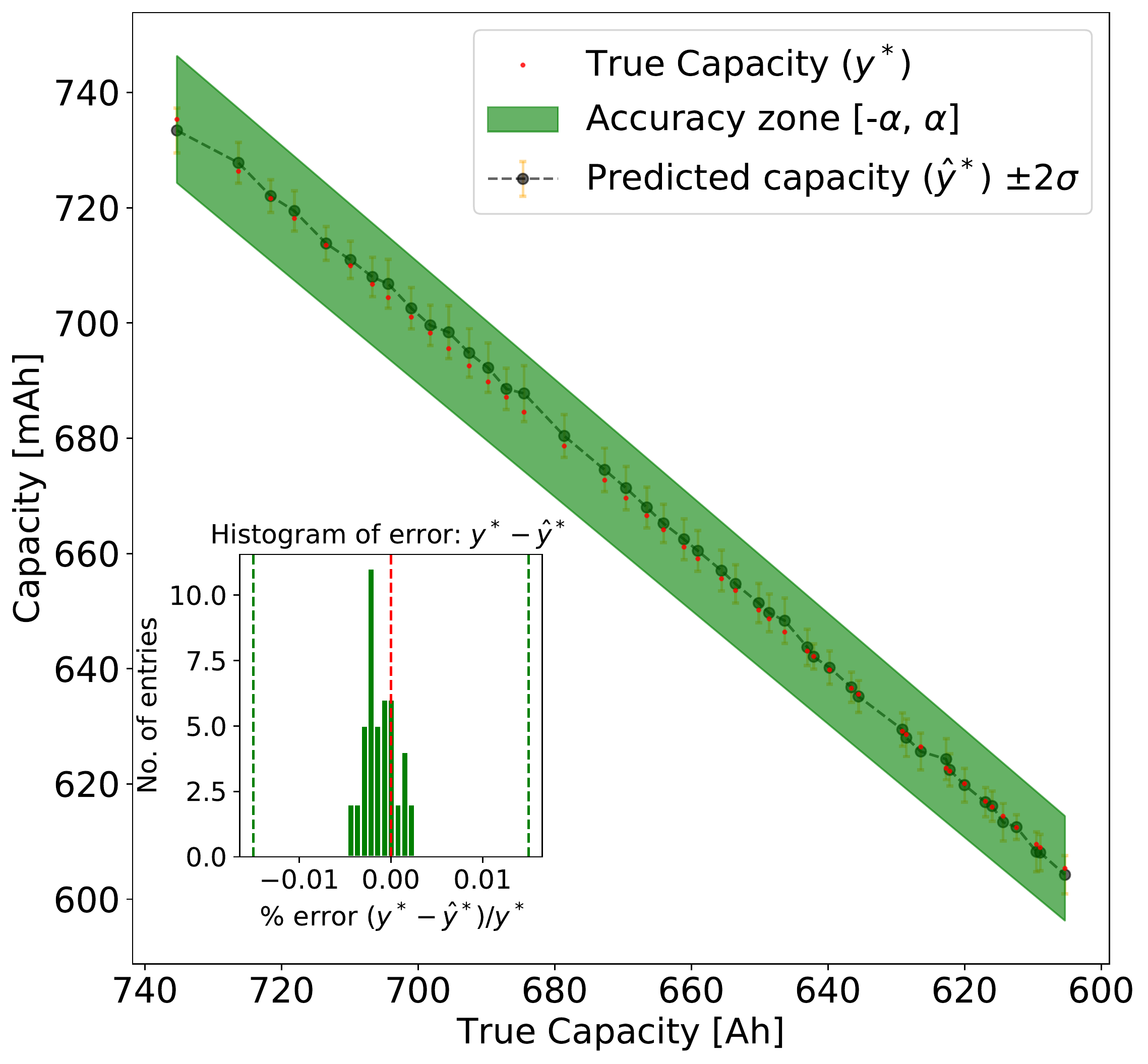}
     \label{hist_group3_RF}
  \end{subfigure}%
  \hfill 
  \begin{subfigure}[b]{.33\textwidth}
     \centering
     \caption{}
					\includegraphics[width=\linewidth]{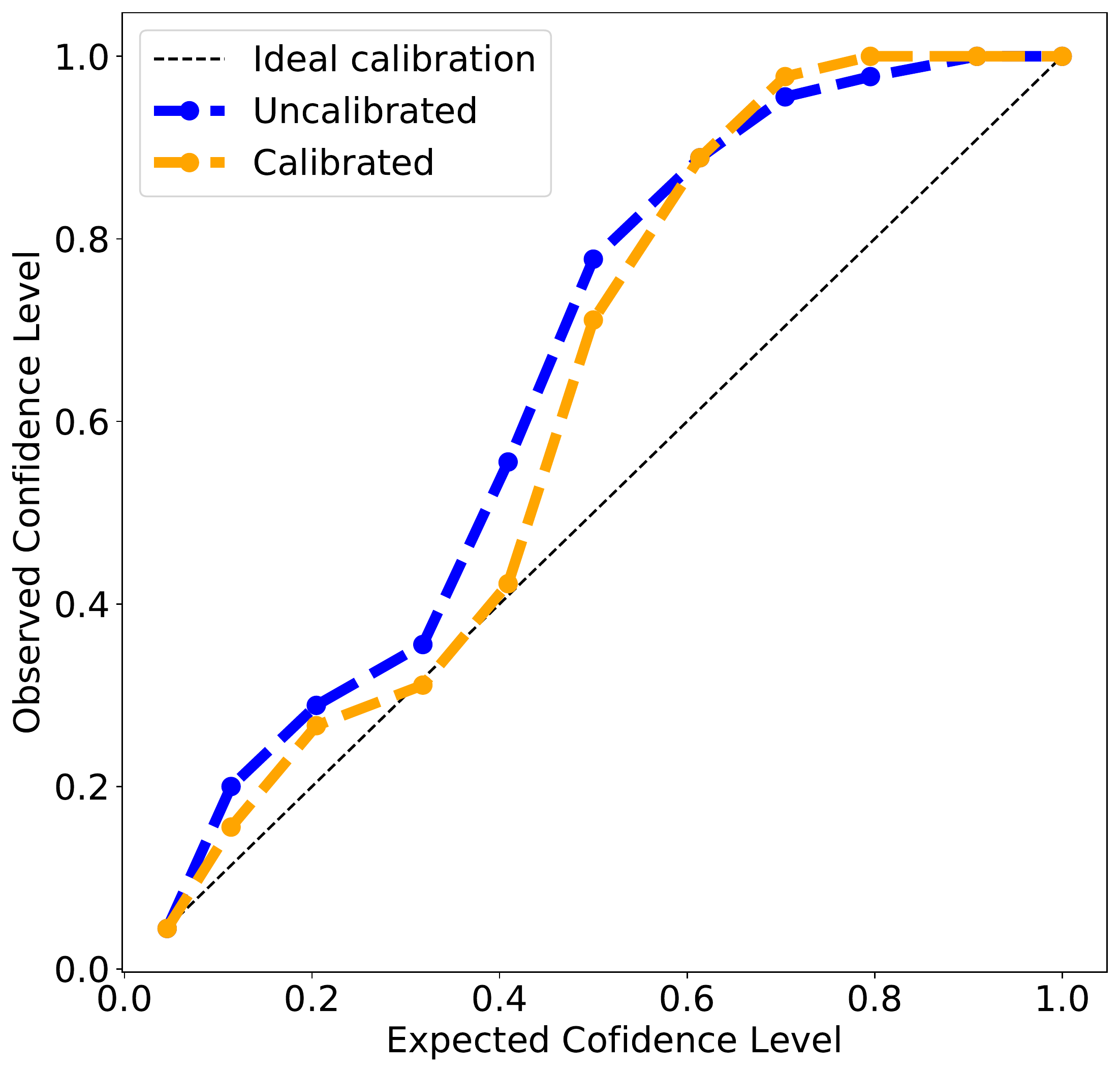}
     \label{calibration_group3_RF}
  \end{subfigure}%
  
  \caption{\textbf{Prediction results with RF Group III cell no. 5.} \textbf{\protect\subref{pred_group3_RF}} RF prediction as a function of cycle numbers, \textbf{\protect\subref{hist_group3_RF}} RF actual vs. predicted capacity, \textbf{\protect\subref{calibration_group3_RF}} RF calibration results.}
  \label{predictions_group3_RF}
 \end{figure*}

\end{document}